\documentclass[authoryear,a4paper]{elsarticle}

\usepackage{verbatim}

\usepackage{geometry}

\usepackage{color,soul}

\usepackage[hidelinks]{hyperref}
\hypersetup{
  colorlinks   = true,
  urlcolor     = blue,
  linkcolor    = blue,
  citecolor    = blue
}

\usepackage{graphicx}
\graphicspath{{./images/}}
\DeclareGraphicsExtensions{.pdf,.png,.jpg}
\usepackage{placeins}
\usepackage{caption}

\usepackage{subcaption}

\usepackage{booktabs}
\usepackage[table,xcdraw]{xcolor}
\usepackage{graphicx}
\usepackage{adjustbox}
\usepackage{multirow}
\usepackage{makecell}

\usepackage{amssymb}
\usepackage{gensymb}
\usepackage[mathscr]{euscript}
\usepackage{bm}
\usepackage{mathtools}
\usepackage{nccmath}
\usepackage{amsmath}
\usepackage{braket}

\usepackage{algorithm}
\usepackage{algpseudocode}

\usepackage{lineno, hyperref}

\usepackage{lipsum}

\journal{Remote Sensing of Environment}

\begin{document}

\begin{frontmatter}
    \title{Dynamic landslide susceptibility mapping over recent three decades to uncover variations in landslide causes in subtropical urban mountainous areas}

    \author[cuhk,cuhk1,cuhk2]{Peifeng Ma}
    \author[cuhk,swjtu]{Li Chen\corref{cor1}}
    \author[cuhk]{Chang Yu}
	\author[swjtu]{Qing Zhu}
    \author[swjtu]{Yulin Ding}
    
    \cortext[cor1]{Corresponding Author: lichen001@cuhk.edu.hk}

    \address[cuhk]{Institute of Space and Earth Information Science, The Chinese University of Hong Kong, Hong Kong, China}
    \address[cuhk1]{Department of Geography and Resource Management, The Chinese University of Hong Kong, Hong Kong, China}
    \address[cuhk2]{Shenzhen Research Institute, The Chinese University of Hong Kong, Shenzhen, China}
    \address[swjtu]{Faculty of Geosciences and Environmental Engineering, Southwest Jiaotong University, Chengdu, China}

    \begin{abstract}
        Landslide susceptibility assessment (LSA) is of paramount importance in mitigating landslide risks. Recently, there has been a surge in the utilization of data-driven methods for predicting landslide susceptibility due to the growing availability of aerial and satellite data. Nonetheless, the rapid oscillations within the landslide-inducing environment (LIE), primarily due to significant changes in external triggers such as rainfall, pose difficulties for contemporary data-driven LSA methodologies to accommodate LIEs over diverse timespans. This study presents dynamic landslide susceptibility mapping that simply employs multiple predictive models for annual LSA. In practice, this will inevitably encounter small sample problems due to the limited number of landslide samples in certain years. Another concern arises owing to the majority of the existing LSA approaches train black-box models to fit distinct datasets, yet often failing in generalization and providing comprehensive explanations concerning the interactions between input features and predictions. Accordingly, we proposed to meta-learn representations with fast adaptation ability using a few samples and gradient updates; and apply SHAP for each model interpretation and landslide feature permutation. Additionally, we applied MT-InSAR for LSA result enhancement and validation. The chosen study area is Lantau Island, Hong Kong, where we conducted a comprehensive dynamic LSA spanning from 1992 to 2019. The proposed methods outperform other methods even adopting a fast adaptation strategy. The model interpretation results demonstrate that the primary factors responsible for triggering landslides in Lantau Island are terrain slope and extreme rainfall. The results also indicate that the variation in landslide causes can be primarily attributed to extreme rainfall events, which result from global climate change, and the implementation of the Landslip Prevention and Mitigation Programme (LPMitP) by the Hong Kong government.
    \end{abstract}

    \begin{keyword}
        Dynamic landslide susceptibility mapping \sep MT-InSAR \sep landslide feature permutation \sep climate change
    \end{keyword}
\end{frontmatter}

\captionsetup[figure]{labelfont={bf},labelformat={default},labelsep=period,name={Fig.}}

\section{Introduction}
\label{s:introduction}
A landslide is the downward movement of a large amount of rock or soil along a slope, stimulated by tectonic, hydrologic, climatic, or human activities under the influence of gravity \citep{hungr2014varnes}. Over the past few decades, landslides in Hong Kong have resulted in significant loss of life and property. Especially in Lantau Island, the occurrence of landslides is frequent. Identifying and assessing landslide risks in subtropical urban mountainous areas involves a complex analysis of multiple factors, which is further complicated by the dense vegetation coverage. Therefore, government organizations and institutions in Hong Kong have made extensive efforts to conduct risk assessments and implement disaster mitigation measures. Landslide susceptibility (LS), known as evaluating the likelihood of a certain site given specific geographical conditions, is crucial for potential landslide hazard identification and prevention \citep{REICHENBACH201860,FRATTINI201062,BARRA2022113294}.

In recent years, advancements of remote sensing technology and deep learning theory have promoted data-driven approaches for landslide susceptibility assessment (LSA) \citep{merghadi2020machine,STEGER2021145935,RAY20102624,zeng2022graph}. The expanding availability of abundant and high-quality thematic information fulfills the data requirements for machine/deep learning, while the end-to-end nonlinear mapping capabilities of data-driven approaches enable automatic and efficient LSA. Most of the applied methods are basically supervised regression, including logistical regression \citep{goyes2021landslide}, support vector machine (SVM) \citep{ali2022ensemble}, tree ensembles such as boosting \citep{SAHA2021142928} and bagging \citep{9157927}, Multilayer perceptron (MLP) \citep{karakas2022comprehensive}, probabilistic graphical model (PGM) such as deep Boltzmann Machines (deepBM) and deep belief network (DBN) \citep{9130821}, and convolutional neural network (CNN) \citep{SAMEEN2020104249}. In respond to specific scenarios, unsupervised learning techniques were applied to transform raw data into new representations with designed properties such as hidden factor disentanglement \citep{bengio2013representation}, sparse representation \citep{rubinstein2010dictionaries}, transferability \citep{zhanggeo}, resilience to noise and feature deletion \citep{vincent_DAE}, and robustness \citep{CHEN2022102807}. Despite the success, the temporal variations in the landslide inducing environment (LIE), caused by significant changes in external triggering factors, such as rainfall, pose challenges for current LSA methods in adapting to all LIEs across different time periods \citep{lewkowicz2019extremes}.

 In the past few decades, under the effect of global climate change, surface temperature and sea level are rising, and rainstorms are becoming more frequent, resulting in the growth of triggered landslides that are prone to extreme events \citep{cramer2018climate}. In 2010, a Landslip Prevention and Mitigation Programme (LPMitP) was initiated by the Geotechnical Engineering Office (GEO), Hong Kong, with the aim of systematically managing the risks associated with both man-made and natural slopes \citep{LPMitP}. Thereafter, landslide events have been significantly reduced. The innate causes of a landslide occurring are complicated, and these causes vary from year to year \citep{dille2022acceleration}. As a result, it is difficult to expect a data-driven model to adapt to multi-temporal LSA tasks. Most of the current LSA approaches neglect the timeliness of landslide samples and integrate them all for supervised regression. This study presents to separately predict landslide susceptibility maps (LSMs) for different periods (dynamic LSA) and summarizes the dynamic changes of these maps to identify the evolution process of dominant landslide inducing factors (LIFs) with the impact of climate change and local slope management policy. In practice, the dynamic landslide susceptibility mapping for exploring variations in landslide causes still faces certain limitations.

\textbf{Dynamic LSA with high time resolution leads to small sample problem.} To conduct dynamic LSA, referred to in this article as yearly LS mapping, landslide samples are divided according to the time of occurrence. Over a large temporal span for the dynamic LSA, the small sample problem occurs due to the costly access to landslide surveying, especially when the time resolution is high. Inadequate samples make the model training suffer from \textit{overfitting}, thus failing to be generalized to the prediction of each mapping unit \citep{CHEN20230213}. Few LSA studies have discussed the feasibility of their models in scenarios with few samples. For this issue, we examined the possibility of learning representations that respond sensitively to changes in certain aspects related to general and crucial landslide causes. This way, the LSA model for each period could be few-shot adapted \citep{pmlr-v119-teshima20a}.

\textbf{A trained black-box model can fit a dataset well, but is not interpretable.} Although the predictive model can identify 'where' the high landslide susceptibility sites are located, it cannot indicate ‘which’ input landslide features are important and ‘how’ these factors interact and influence the landslide susceptibility prediction \citep{David-Explainable,cheng2021method}. Additionally, it’s difficult to infer innate variation of landslide causes by simply comparing LSM series that generated by the dynamic LSA. Inadequate knowledge of the LIFs is not conductive to excavate landslide evolution law and carry out scientific and effective landslide prevention measures. 

\textbf{Additionally, we explored the application of MT-InSAR for the enhancement and validation of LSM.} LSMs inferred from data-driven methods essentially indicate the correlation between the likelihood of landslide occurrence and landslide features. Ground deformation directly reveals slope dynamics \citep{EZQUERRO2023113668}, providing valuable support for enhancing and validating LSM. The InSAR technique is superior for measuring large-scale ground deformation with millimeter level accuracy \citep{OSMANOGLU201690,WU2023113545,KIM2022113231}. Multi-temporal InSAR (MT-InSAR) achieves precise deformation accuracy by using multiple SAR images. In urban areas, PSInSAR identifies high-coherence features like man-made structures, while DSInSAR locates widely distributed scatters in suburban or mountainous regions using methods like SqueeSAR \citep{ferretti2011new} and SBAS \citep{DITRAGLIA201895}. Hence, it is necessary to identify ways for applying InSAR techniques to enhance and validate landslide susceptibility prediction \citep{CIAMPALINI2016302}.

To address these problems, this article introduces a novel dynamic LSA method, applied to the case of Lantau Island spanning the years 1992 to 2019. Specifically, to account for the variations in landslide causes on Lantau Island, we decomposed the entire LSA task into multiple subtasks through temporal division. In this case, some subtasks faced small sample scenarios. Accordingly, we proposed to meta-learn representations capable of transferring commonalities of landslide causes among different LSA subtasks, enabling fast adaptation of the predictive model for these subtasks. To interpret the model predictions and characterize the dominant LIFs for various periods, we utilized Shapley Additive exPlanations (SHAP) for feature permutation. Finally, we apply InSAR derived deformation map to enhance the initial LSM using an empirical matrix \citep{zhou2022enhanced}. The code and data are publicly available at \url{https://github.com/CLi-de/D_LSM}.

\begin{figure}[t]
	\centering
	\includegraphics[width=\textwidth]{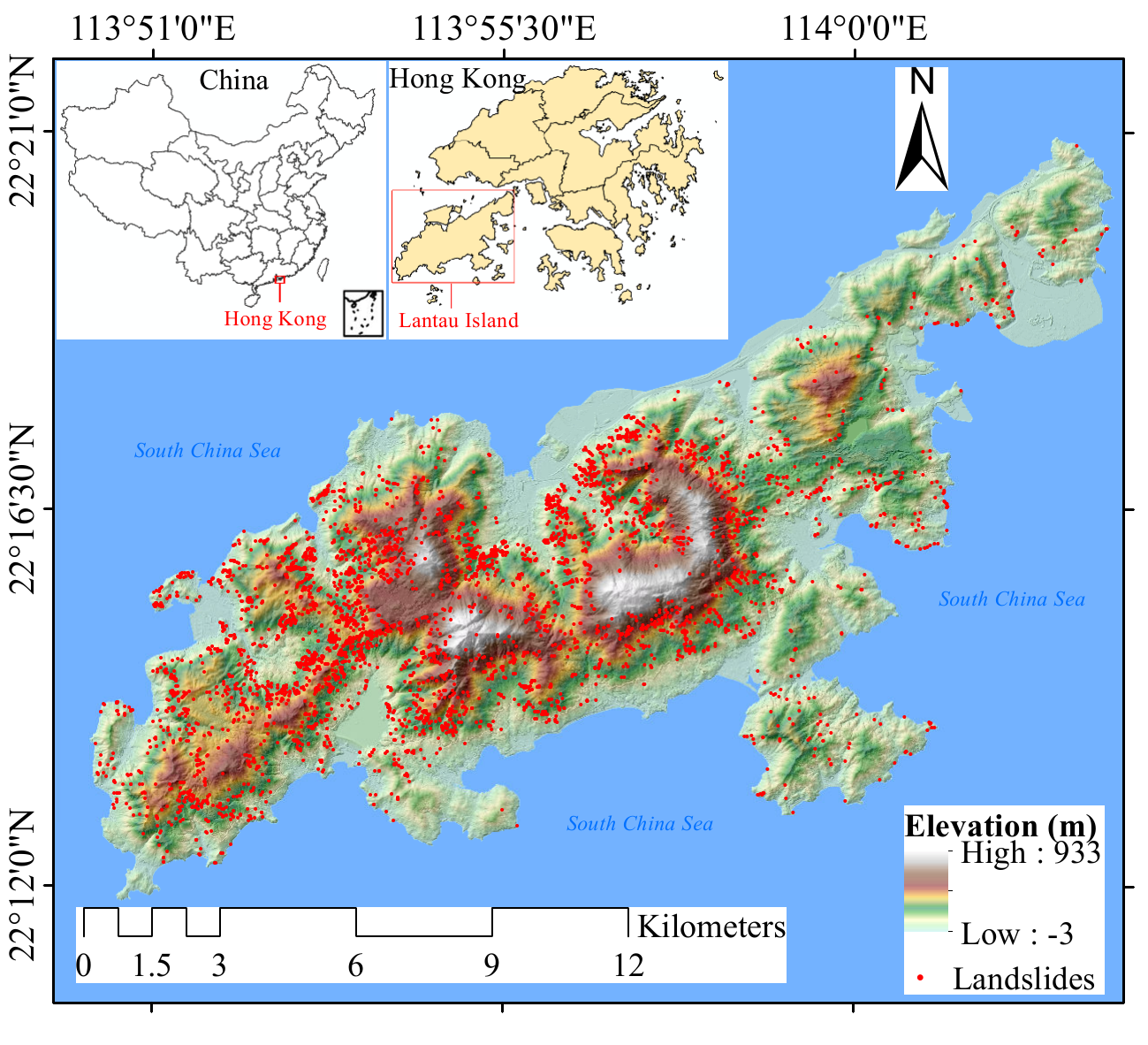}
	\caption{Location of study area and distribution of landslides from 1992 to 2019.}
	\label{fig:location}
\end{figure}

\section{Study area and dataset}
\label{s:Study area and dataset}

Lantau Island is the largest outlying island located in the southwest of Hong Kong, China, with an area of 147.16 $km^2$. As seen in Fig. \ref{fig:location}, the island features mainly mountainous terrain, with steep hills and slope gradients that commonly fall between 25$^{\circ}$ and 40$^{\circ}$. The island's natural foot slopes are typically covered by woody forests while the mid-slopes are dominated by bushes and grass. The bedrock of Lantau Island comprises Mesozoic volcanic rocks found in the west and younger intrusive igneous rocks located in the northwest, both of which have undergone significant in-situ weathering. Lantau Island was selected as the experimental area due to its high occurrence of natural landslides. The Hong Kong government and academic organizations have extensively gathered valuable thematic information on the terrain, geology, hydrological environment, and other aspects to provide effective support for high-resolution spatiotemporal dynamic landslide susceptibility mapping.

The historical landslides are available at \url{https://www.geomap.cedd.gov.hk/GEOOpenData/eng/ENTLI.aspx}, maintained by the Civil Engineering and Development Department (CEDD), Hong Kong, in the Enhanced Natural Terrain Landslide Inventory (ENTLI). There are over 4000 recorded landslides in Lantau Island ranging from 1992 to 2019 that were used for positive sample generation. Accordingly, we manually selected non-landslide samples on \textit{Arcmap} based on some certain priors, such as the tendency to select the negative samples in areas with low slope or frequency of landslide. The considered LIFs include elevation, slope, curvature, aspect, lithology, land use, normalized difference vegetation index (NDVI), stream power index (SPI), topographic wetness index (TWI), annual rainfall (AR), annual extreme rainfall days (AERD), distance to faults, drainage, catchment, and road lines. These LIFs involve a wide range of fields, including topography, geology, hydrology, land cover, human activity, and climate. The related multi-source data are processed on ArcGIS platform to obtain landslide thematic data (as shown in Fig. \ref{fig:Thematic maps of Lantau Island}) for later sample vector sampling. Table \ref{tab:data source} includes comprehensive details of these data. We collected SAR data from various spaceborne sensors, including 31 ascending ENVISAT ASAR images (4/2003-9/2010), 21 ascending ALOS-PALSAR-1 images (6/2007-1/2011), and 209 ascending Sentinel-1A images (6/2015-12/2019).

\begin{table*}[h]
    \caption{Data source and details.}
    \centering
    \footnotesize
    \renewcommand\arraystretch{1.25}  
    \setlength\tabcolsep{2.5pt}  
    \begin{tabular}{ccccc}
        \hline
        Type                            & Data           & Source                                       & \makecell[c]{Spatial \\ resolution}    & \makecell[c]{Temporal \\ coverage}  \\ \hline
        \multirow{4}{*}{Topography}     & Elevation      & Land Department, Hong Kong                   & 5m         & -           \\ \cline{2-5} 
                                        & Slope          & \multirow{3}{*}{Derived from elevation}      & 5m         & -           \\ \cline{2-2}  \cline{4-5} 
                                        & Curvature      &                                              & 5m         & -           \\ \cline{2-2}  \cline{4-5} 
                                        & Aspect         &                                              & 5m         & -           \\ \hline
        \multirow{2}{*}{Geology}        & Lithology      & \multirow{2}{*}{\makecell[c]{Geotechnical Engineering Office, \\ CEDD, Hong Kong }}               & Polygon     & -  \\ \cline{2-2}  \cline{4-5}
                                        & Faults         &                                                                                                   & Polyline    & -  \\ \hline
        \multirow{4}{*}{Hydrology}      & SPI            & \multirow{2}{*}{Derived from elevation}      & 5m     & -  \\ \cline{2-2}  \cline{4-5}
                                        & TWI            &                                              & 5m     & -  \\ \cline{2-5}
                                        & Drainage       & \makecell[c]{Resource and Environmental Science \\ and Data Center, CAS, China}           & Polyline    & -  \\ \cline{2-5}
                                        & Catchment      & \makecell[c]{Water Supplies Department, \\ Hong Kong}                                     & Polygon     & -  \\ \hline
        \multirow{2}{*}{Land cover}     & Land use       & Planning Department, Hong Kong                                                            & Polygon     & -  \\ \cline{2-5}
                                        & NDVI           & Geospatial Data Cloud, CAS, China                                                         & 30m         & -  \\ \hline
        Human activity                  & Road lines     & OpenStreetMap                                                                             & Polyline    & -  \\ \hline
        \multirow{2}{*}{Climate}        & AR             & Hong Kong Observatory                        & -          & 01/1992 - 12/2019 \\ \cline{2-5}
                                        & AERD           & Hong Kong Observatory                        & -          & 01/1992 - 12/2019 \\ \hline
        \multirow{3}{*}{InSAR data}     & ASAR           & Airbus Defence and Space, UK                 & -          & 04/2003 - 09/2010 \\ \cline{2-5}
                                        & ALOS           & JAXA, Japan                                  & -          & 06/2007 - 01/2011 \\ \cline{2-5}
                                        & Sentinel-1A    & European Space Agency                        & -          & 06/2015 - 12/2019 \\ \hline  
    \end{tabular}
    \label{tab:data source}
\end{table*}

\begin{figure}[tbhp]
    \centering
    \begin{subfigure}{0.495\textwidth}
        \includegraphics[width=\textwidth]{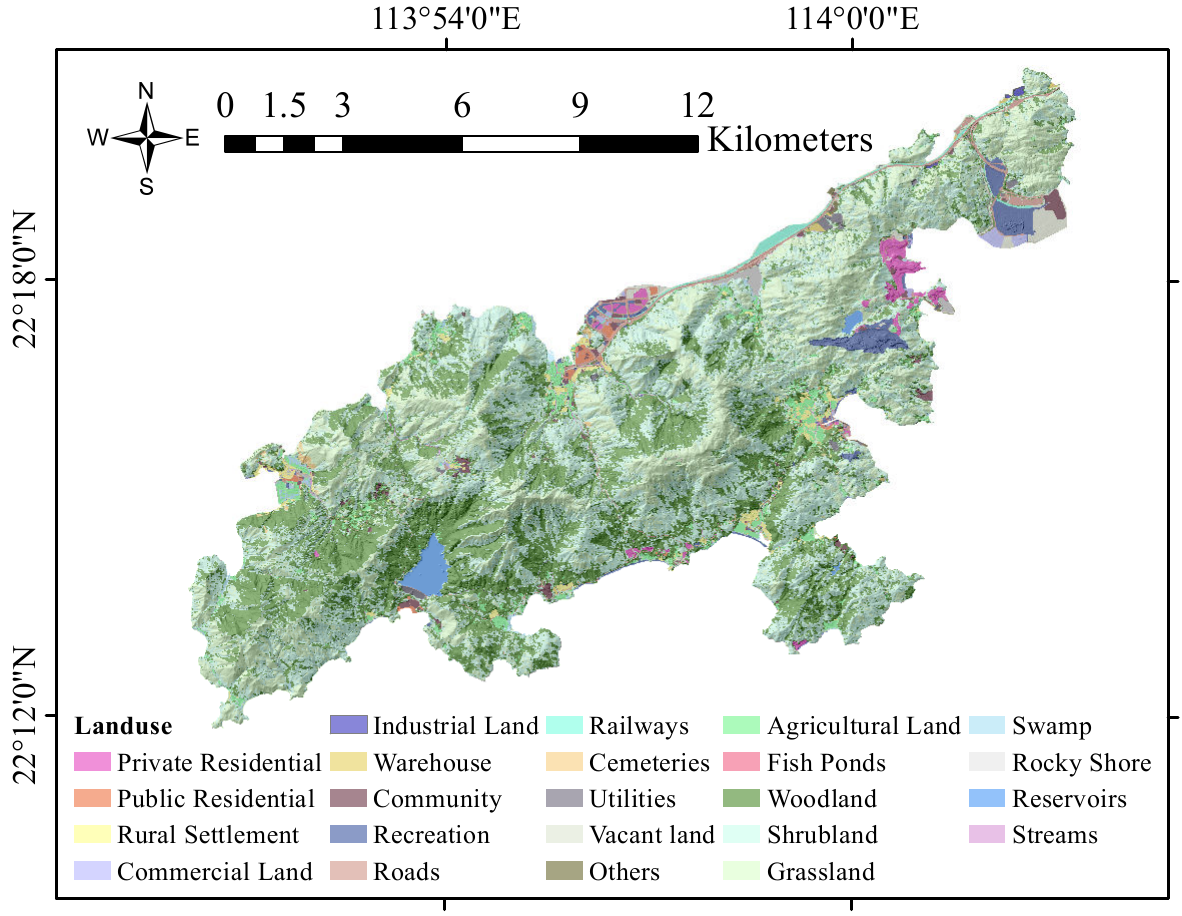}
        \caption{land use}
    \end{subfigure}
    \begin{subfigure}{0.495\textwidth}
        \includegraphics[width=\textwidth]{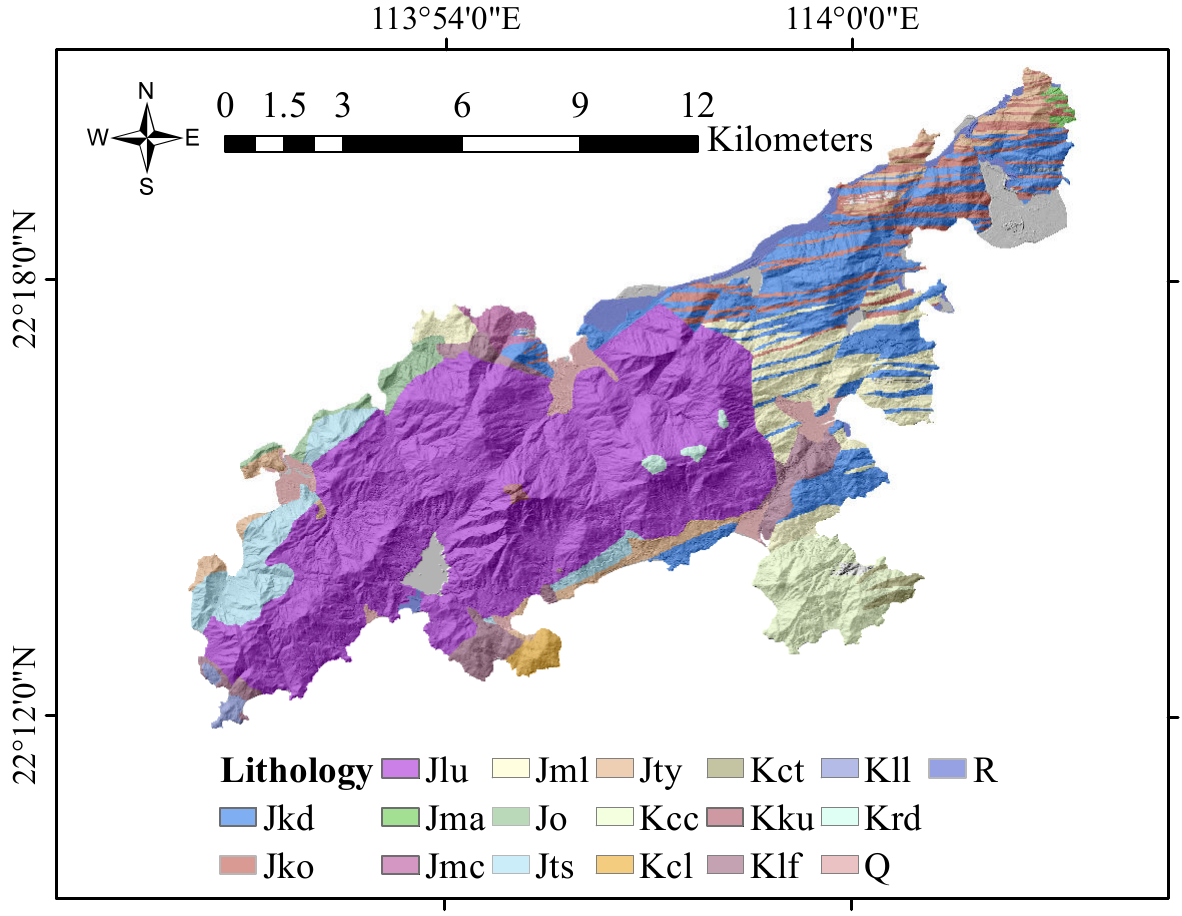}
        \caption{lithology}
    \end{subfigure}

	\begin{subfigure}{0.495\textwidth}
        \includegraphics[width=\textwidth]{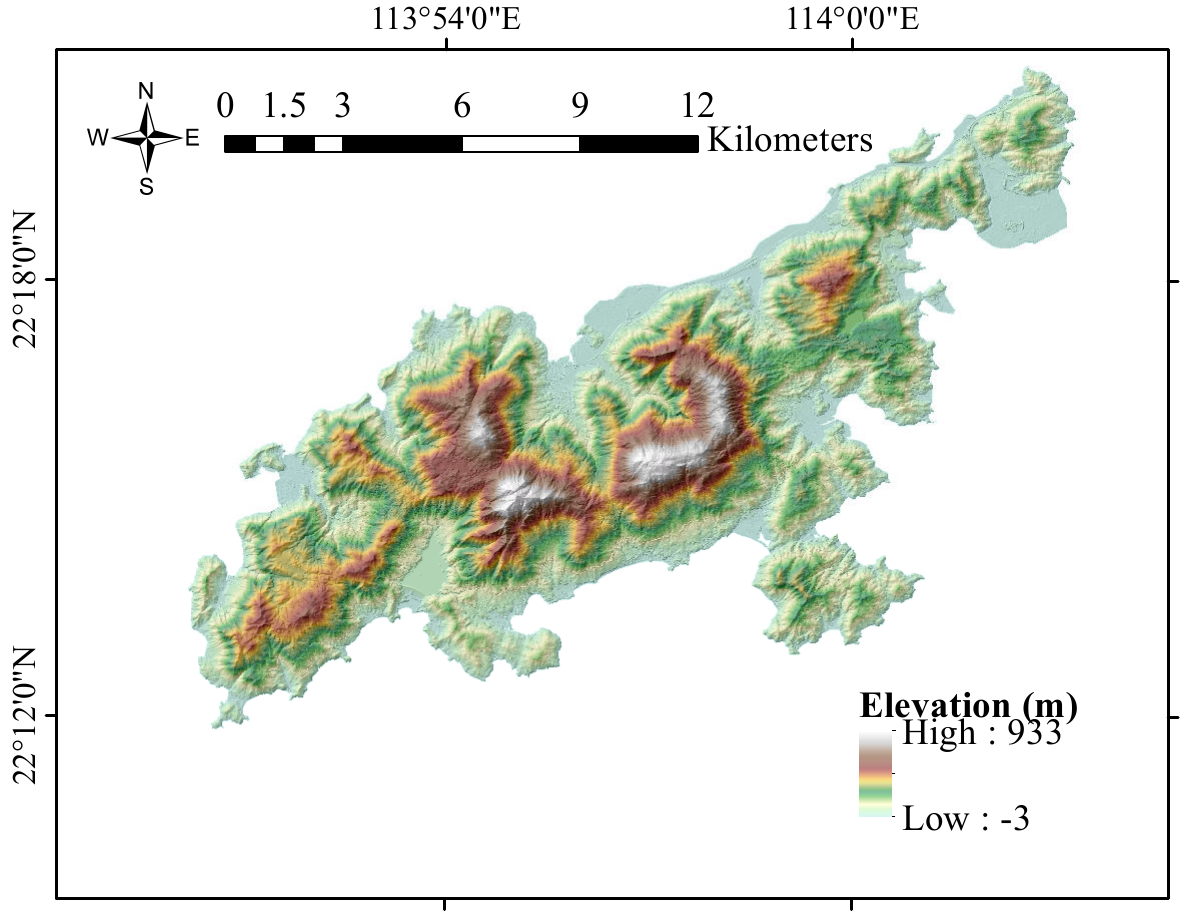}
        \caption{DEM}
    \end{subfigure}
    \begin{subfigure}{0.495\textwidth}
        \includegraphics[width=\textwidth]{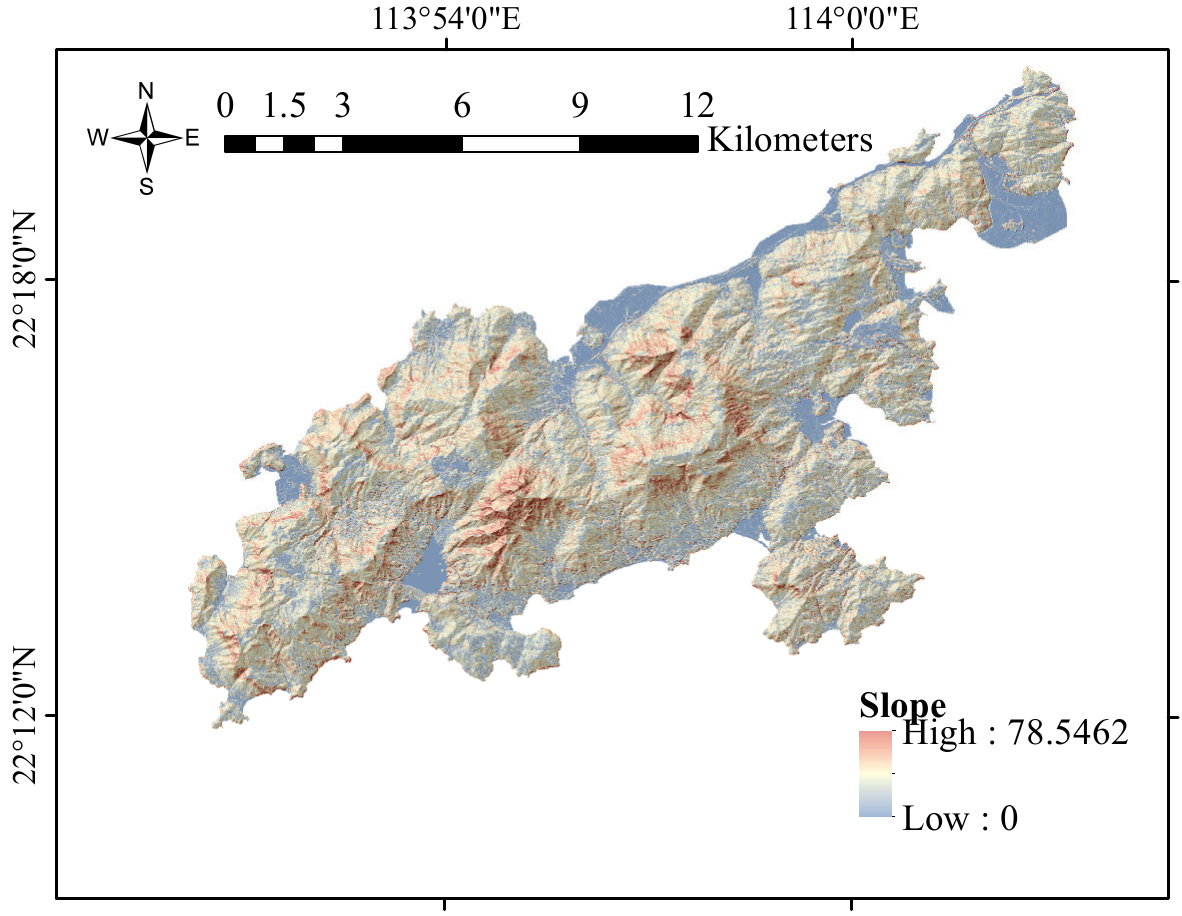}
        \caption{slope}
    \end{subfigure}

    \begin{subfigure}{0.495\textwidth}
        \includegraphics[width=\textwidth]{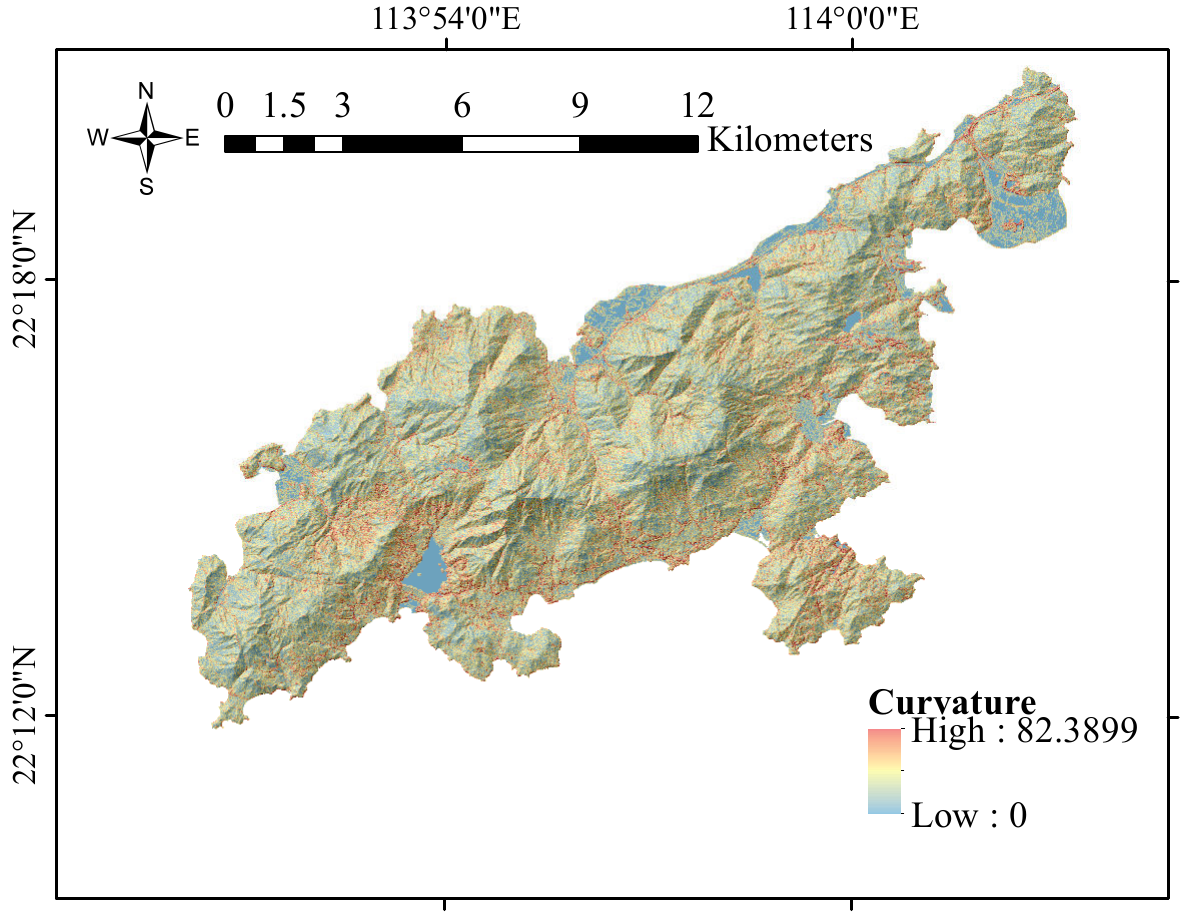}
        \caption{curvature}
    \end{subfigure}
	\begin{subfigure}{0.495\textwidth}
        \includegraphics[width=\textwidth]{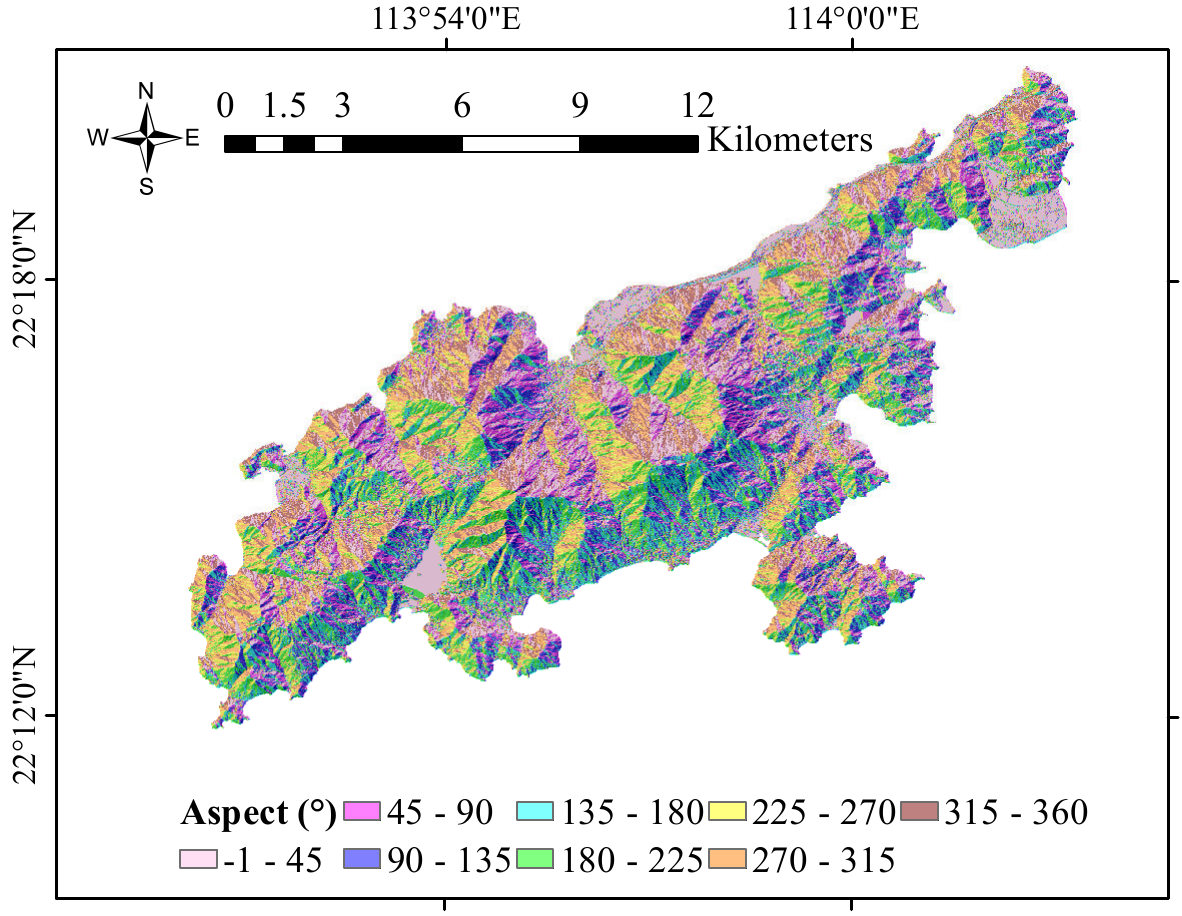}
        \caption{aspect}
    \end{subfigure}

    \caption{Landslide thematic maps. (a)Land use, (b) lithology, (c) DEM, (d) slope, (e) curvature, (f) aspect, (g) NDVI (The value is amplified 10,000 times), (h) SPI, (i) TWI, (j) catchment \& drainage, (k) road lines \& faults, (l) precipitation.}
    \label{fig:Thematic maps of Lantau Island}
\end{figure}

\begin{figure}[tbhp]
    \ContinuedFloat
    \centering
    \begin{subfigure}{0.495\textwidth}
        \includegraphics[width=\textwidth]{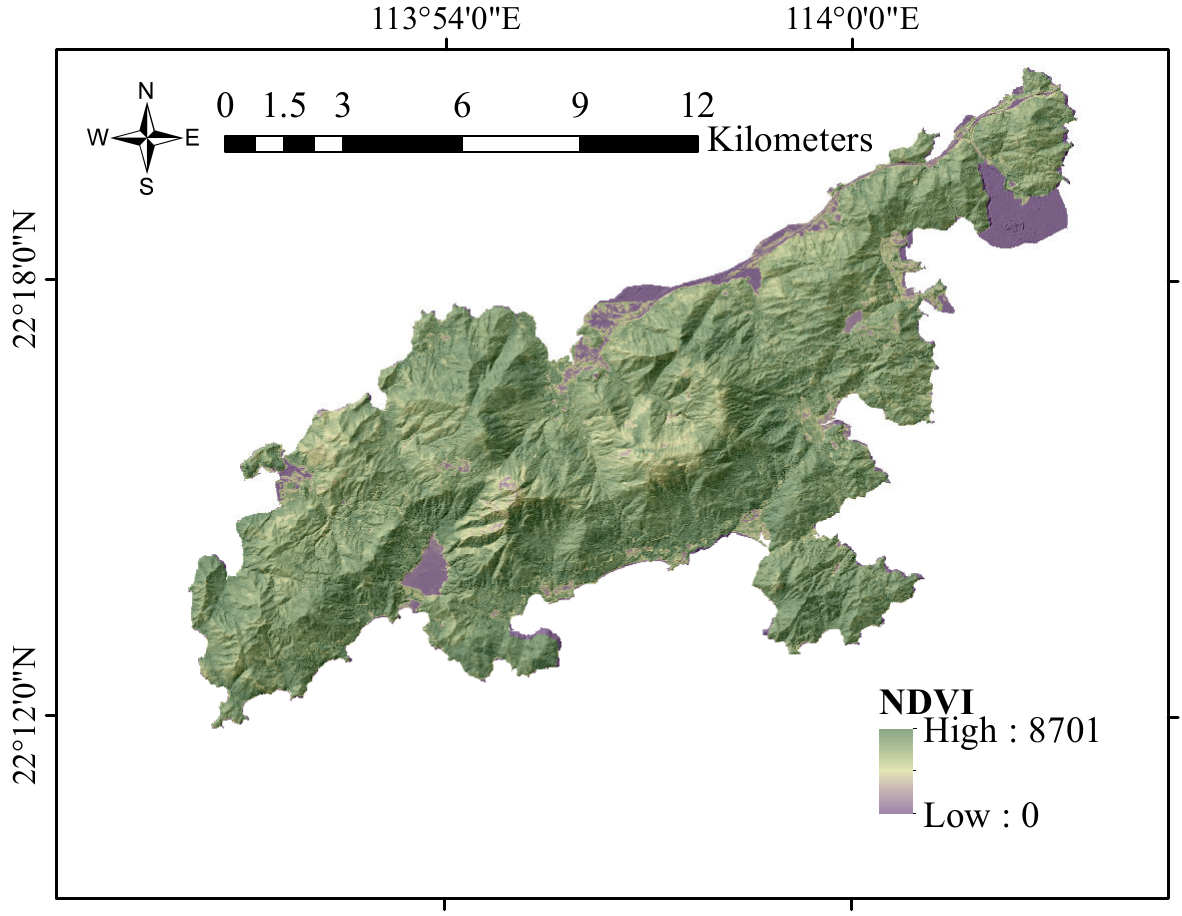}
        \caption{NDVI}
    \end{subfigure}
    \begin{subfigure}{0.495\textwidth}
        \includegraphics[width=\textwidth]{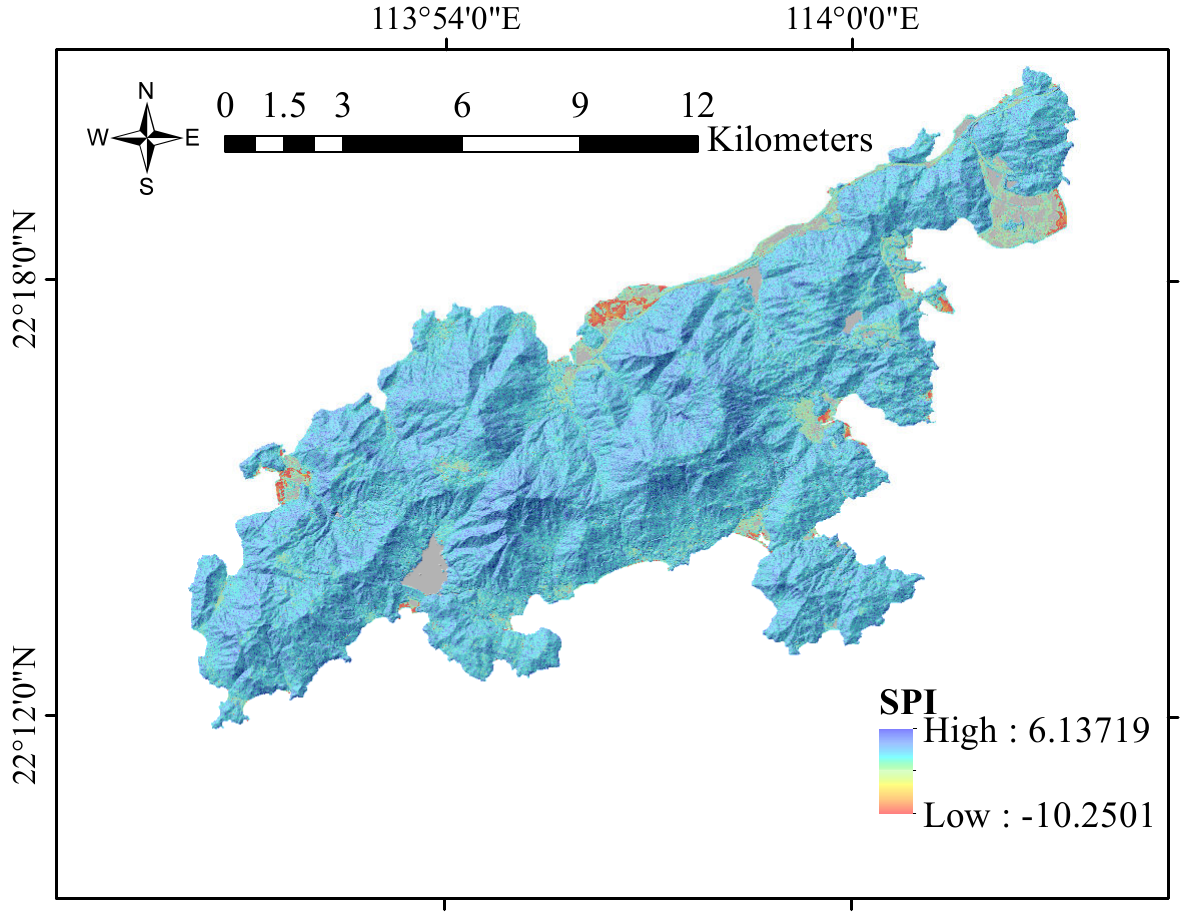}
        \caption{SPI}
    \end{subfigure}

	\begin{subfigure}{0.495\textwidth}
        \includegraphics[width=\textwidth]{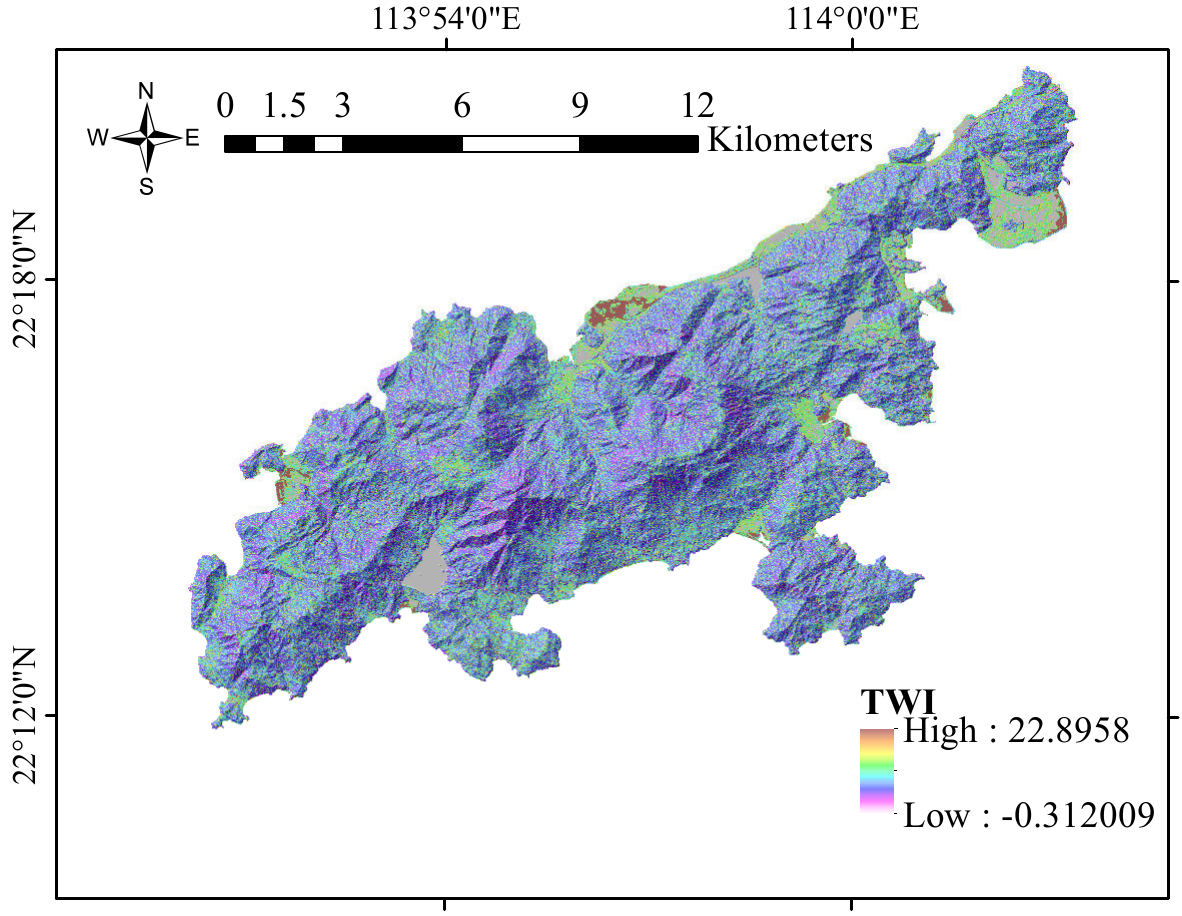}
        \caption{TWI}
    \end{subfigure}
    \begin{subfigure}{0.495\textwidth}
        \includegraphics[width=\textwidth]{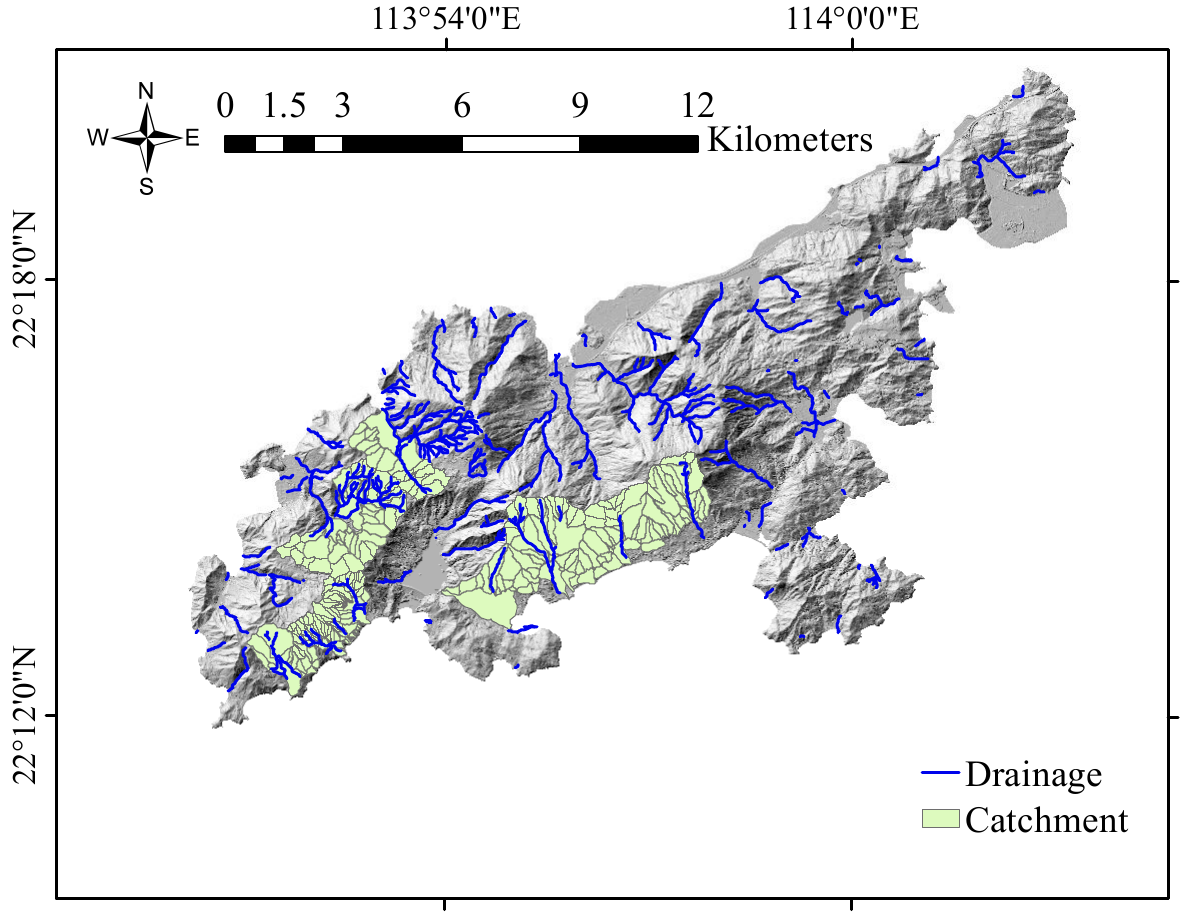}
        \caption{drainage \& catchment}
    \end{subfigure}

    \begin{subfigure}{0.495\textwidth}
        \includegraphics[width=\textwidth]{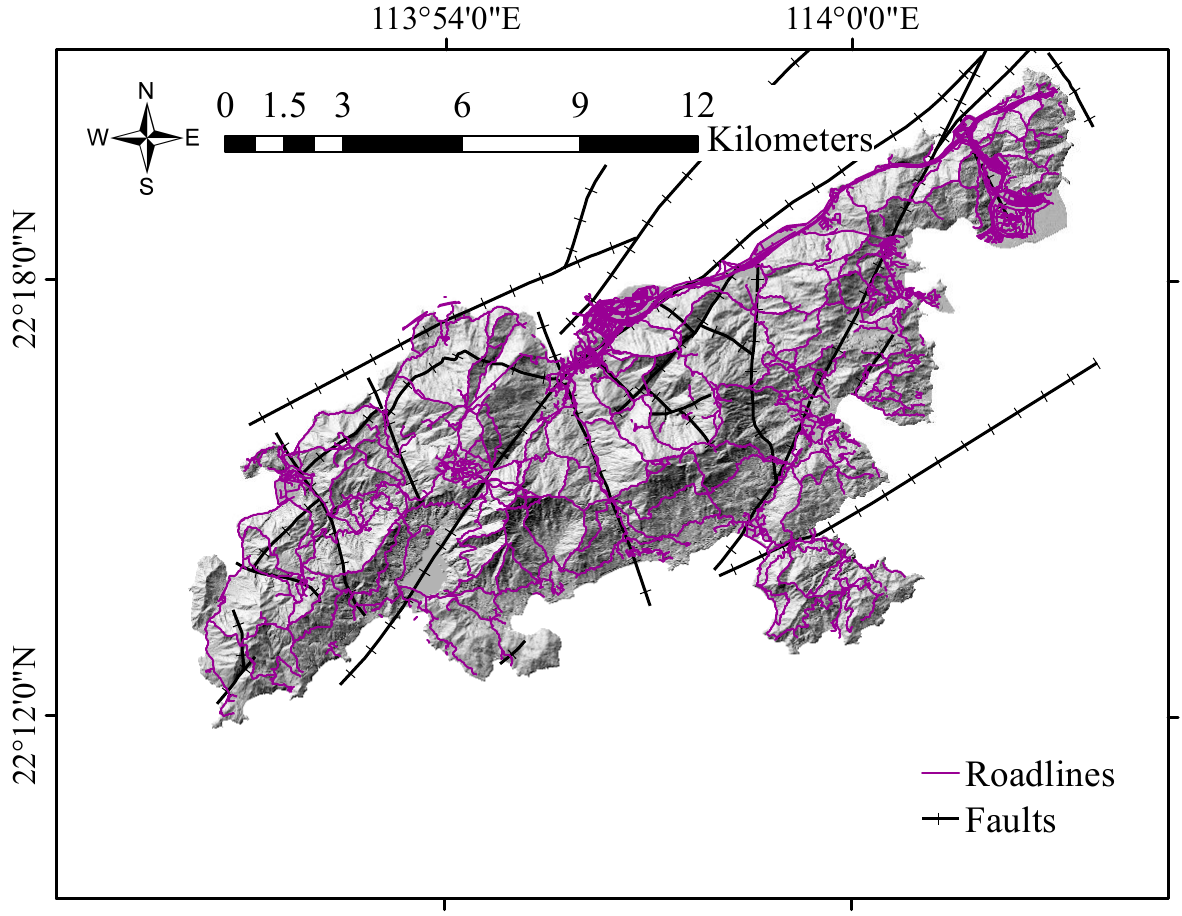}
        \caption{road lines \& faults}
    \end{subfigure}
	\begin{subfigure}{0.495\textwidth}
        \includegraphics[width=\textwidth]{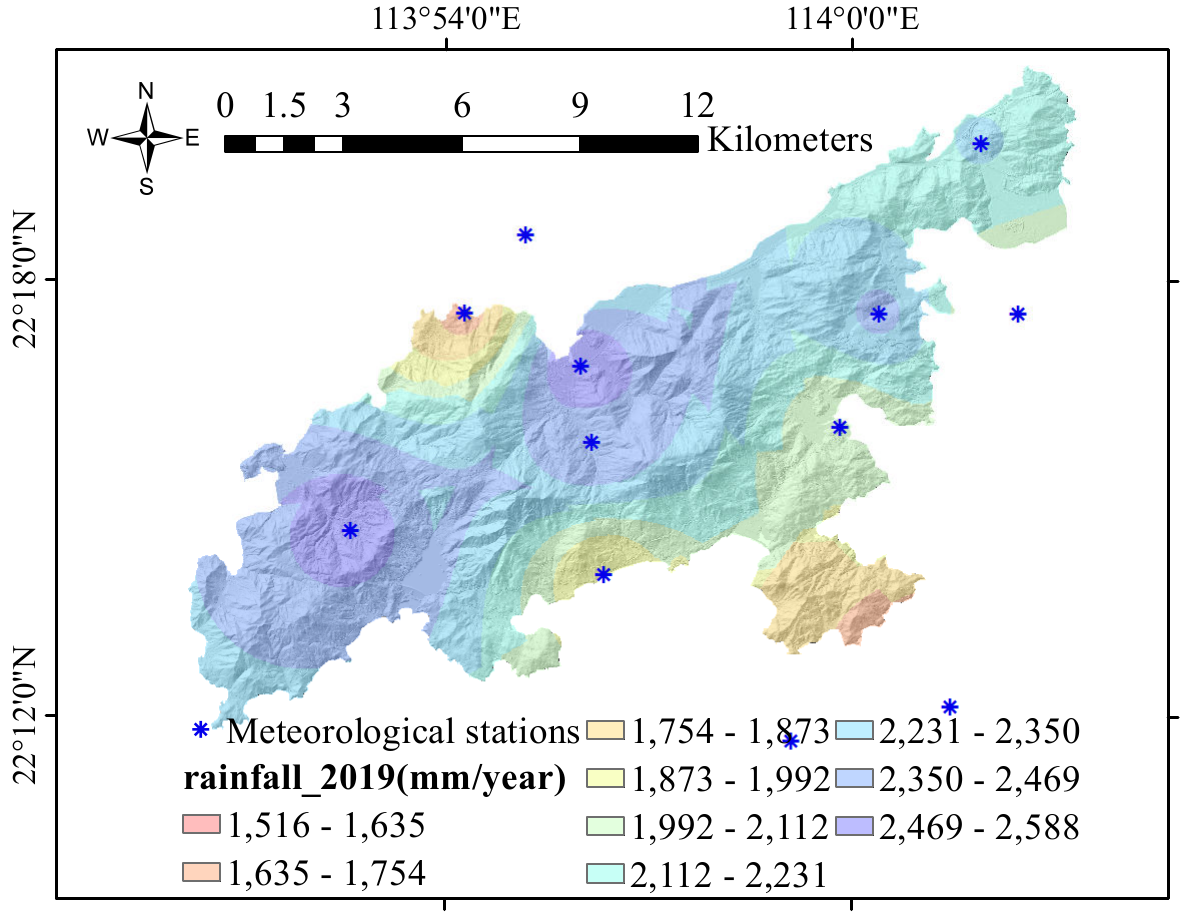}
        \caption{precipitation}
    \end{subfigure}
    \caption{(continued)}
\end{figure}

\begin{figure}[tbhp]
	\centering
	\includegraphics[width=\textwidth]{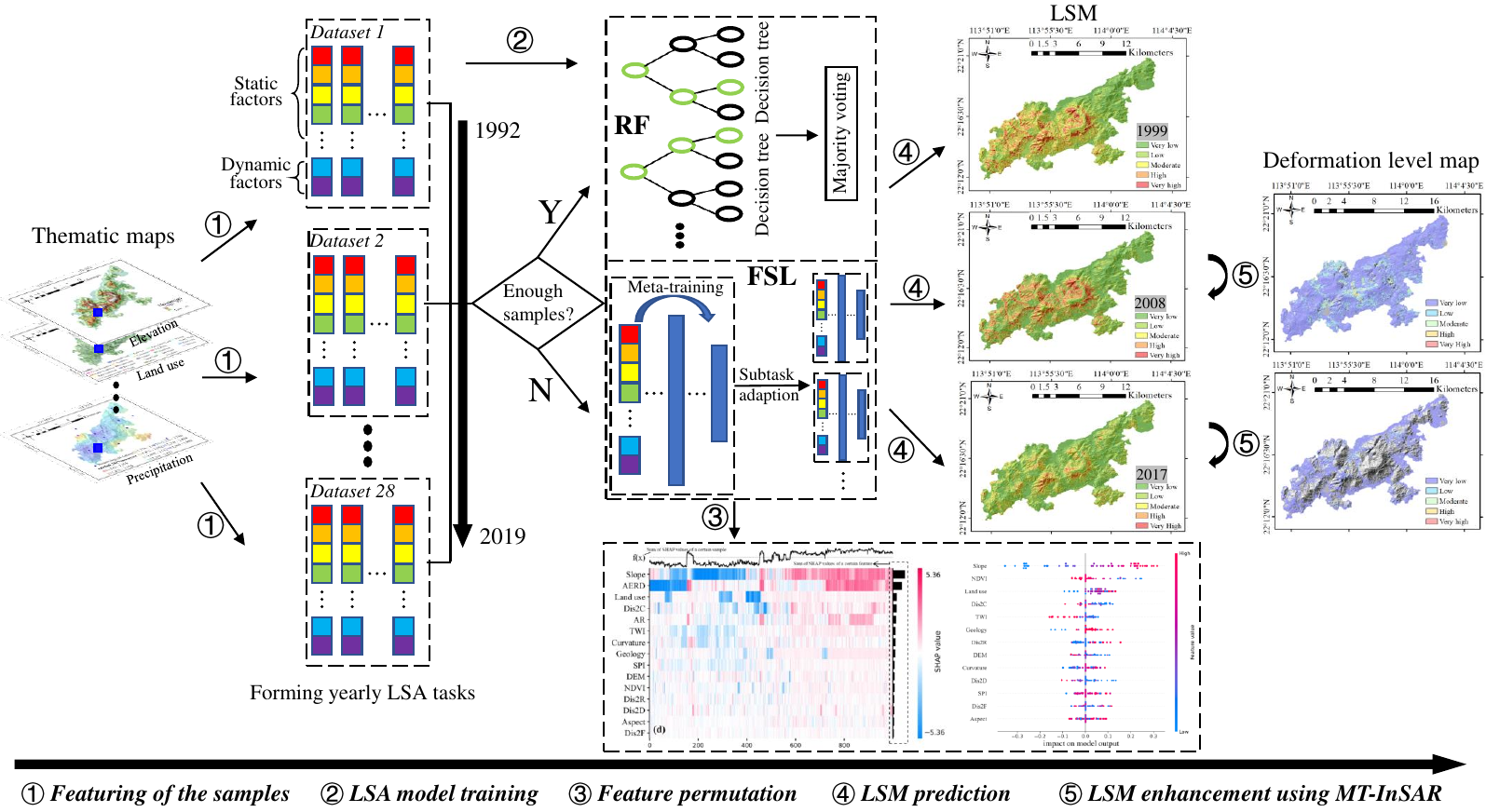}
	\caption{The overflow of the proposed method.}
	\label{fig:overflow}
\end{figure}

\section{Methodology}
\label{s:Methodology}
The overflow of the proposed method is presented in Fig. \ref{fig:overflow}, mainly containing the following components: (1) Featuring of the samples and yearly LSA tasks construction in section \ref{subsubs:3.1.1}; (2) LSA model training as is introduced in section \ref{subsubs:3.1.2} and \ref{subsubs:3.1.3}; (3) Feature permutation in section \ref{subs:3.2}; (4) Yearly landslide susceptibility mapping in section \ref{subsubs:3.1.4}; and (5) LSM enhancement using MT-InSAR technique in section \ref{subs:3.3}.

\subsection{Dynamic Landslide susceptibility mapping}
\label{subs:3.1}

The environment varies not only in spatial extent but also over time due to changing geographical and climatic conditions. Dynamic LIFs like AR and AERD change with time, making it difficult for a single model to effectively learn multi-temporal tasks in such circumstances. This paper presents the application of yearly LSA as a strategy to mitigate the negative impacts of multi-temporal environmental conditions. In years with a substantial number of available landslide samples, we employ the state-of-the-art RF-based LSA algorithm \citep{BELGIU201624} to predict the landslide susceptibility of mapping units. RF is chosen for its superior parallel training efficiency, generalization ability, and resistance to noise. However, in years when valid landslide samples are scarce, we adopt a meta-learning approach to develop an intermediate representation for few-shot learning (FSL) of the target predictive model.

\subsubsection{Data processing and yearly LSA tasks construction}
\label{subsubs:3.1.1}
Sample vectors, representing either a landslide or non-landslide record, were featured on thematic maps according to their respective spatial locations. Feature values corresponding to raster thematic maps, such as elevation, slope, aspect, curvature, SPI, TWI, and NDVI, were assigned based on their respective grid values. To value features on polyline maps, such as faults, drainage, catchment lines, and road lines, we calculated the closest distance of the samples to nearby line objects. For polygon maps, such as lithology and land use, we assigned values from 1 to 3 based on the statistical frequency of landslide occurrence in different categories (polygons). Unlike statistical data that almost remains constant, the valuing of AR and AERD features varies annually. Therefore, we assigned values based on the year of the samples. We collected precipitation information from meteorological stations and interpolate the raster thematic map using the inverse distance weighted technique. The rainfall map is shown in Fig. \ref{fig:Thematic maps of Lantau Island} (n). 

Missing values are common and inevitable, and it is important to appropriately address them. We replaced missing values with the mean value along the corresponding dimension where at least one valid value was present. Afterward, we labelled the samples as 1 if they were of a landslide, and 0 if non-landslide. For the dynamic LSA purpose, we constructed an LSA task for each year from 1992-2019 using these labelled samples. Specifically, these labelled samples were divided by year, forming tasks $\{\mathcal{T}_0, \mathcal{T}_1, ..., \mathcal{T}_{27}\}$ for each of the 28 years.

\subsubsection{Random forest}
\label{subsubs:3.1.2}
The RF algorithm is essentially an extension of the bagging methods, which combines the predictive results of multiple base classifiers to determine the final result. The results are determined through a majority vote conducted by the decision trees \citep{97458}.  The method has been widely used in numerous fields and continues to be the most prevalent approach for landslide susceptibility assessment. We choose RF mainly for the following reason: 1) The training process can be highly parallelized, leading to significant improvements in training efficiency; 2) Due to random sampling of samples and attributes, the trained model exhibits strong generalization ability; 3) There is a high tolerance for missing values in some features, which is common in LIF featuring. 

RF can utilize permutation feature importance to assess the importance of LIF. The underlying concept is straightforward: it quantifies the reduction in model performance (e.g., Accuracy and RMSE) when the feature's values are randomly shuffled, thereby identifying features that significantly influence the model's performance. However, this approach has limitations when highly correlated input landslide features are present since eliminating a correlated feature may have minimal impact on the model's performance. Therefore, in this study, we adopted SHAP \cite{WOS:000571254100009,WOS:000446910800009} for LIF permutation. SHAP has a solid foundation in game theory and will be elaborated further in section \ref{subs:3.3}.

\subsubsection{Meta-learning representations with strong generalization ability}
\label{subsubs:3.1.3}
The uneven distribution of landslide points gives rise to small sample problems in some subtasks (years). To address the problem, we implemented meta-learning features that have the capability to quickly learn from a limited number of samples and comprehend general concepts \citep{pmlr-v70-finn17a}, such as similarities in landslide causes that can be generalized among various LSA tasks. Establishing strong representations via meta-learning enabled fast learning and adaptation of models for the subtask with few samples.

The datasets for meta-learning differ from the datasets for multitask learning used for the adaptation of multiple LSA tasks. In order to enhance meta-learning and create an intermediate model with strong generalization ability, a sufficient number of meta-tasks is required. Accordingly, we further divide multiple LSA tasks with over 50 landslide samples into subtasks, to expand the pool $p\{\mathcal{T}\}$ of potential subtasks $\mathcal{T}_i\{\mathcal{D}_i, \mathcal{L}_i\}\thicksim p\{\mathcal{T}\}$, where $\mathcal{D}_i$ represent sample vectors, $\mathcal{L}_i$ represents loss function of the $i$-th task. To maintain a balance of positive and negative samples, we randomly selected an equal number of non-landslide samples, along with the landslide samples, to create meta-dataset $\mathcal{D}$. $\mathcal{L}_i$, if not specified and unless otherwise stated, refers to cross-entropy in this study. For each subtask $\mathcal{T}_i$, to produce meta-datasets, we divided $\mathcal{D}_i$ into a support set $\mathcal{S}_i$ for meta-training and a query set $\mathcal{Q}_i$ for meta-testing. The subtasks $\{\mathcal{T}_1, \mathcal{T}_2, ...,\mathcal{T}_k,...\}$ were partitioned into training ($\mathcal{D}_{train}$) and testing datasets ($\mathcal{D}_{test}$) at 3:1 ratios.

\begin{figure}[tbhp]
	\centering
	\includegraphics[width=\textwidth]{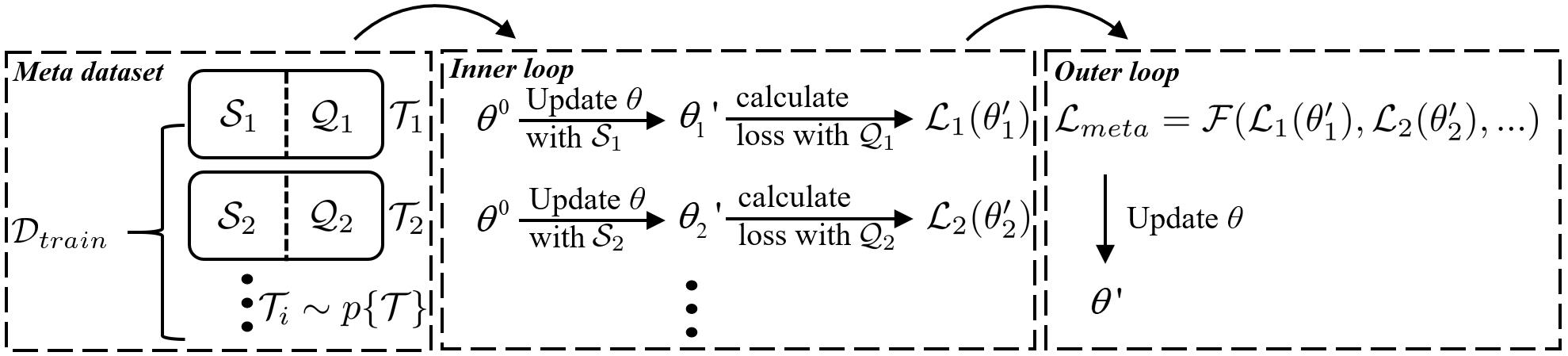}
	\caption{The proposed meta-learning strategy. $\theta_i'$ is the updated network parameters of $\mathcal{T}_i$ in the inner loop computation. $\theta'$ is the network parameters of the intermediate model in the outer loop computation. $\mathcal{F}$ is the function for meta-objective.}
	\label{fig:meta learning}
\end{figure}

Fig. \ref{fig:meta learning} shows the used meta-learning strategy based on MLP, which is designed to acquire a general intermediate model for few-shot adaptation of an LSA task. The meta-learning strategy consists of two computation loops (inner and outer loops) \citep{9428530}. The first optimizes network parameters based on the objective of each subtask, while the other optimizes aspects that guide the optimization of the inner loop, such as the hyperparameters tuning and multitask learning. Specifically, the inner update rule of LSA model $f_{\theta_i'}$ subtask is given in Eq. \ref{eq:inner computation}:
\begin{equation}
    \begin{split}
      \theta_i'=\theta^0-\alpha g_{\theta}\mathcal{L}_i,
      \label{eq:inner computation}
    \end{split}
\end{equation}
where $\theta^0$ denotes the initialized network parameter, $\alpha$ represents the inner learning rate, $g_{\theta}$ is the back propagation gradient calculated by feeding the samples in $\mathcal{S}_i$ into model. It is noteworthy that the inner learning rate $\alpha$ can be set as trainable to ensure a fast and stable training of the model. The objective of outer loop (meta objective) is to minimize the sum of the losses of a batch of meta-tasks, as is given in Eq. \ref{eq:meta-objective}:
\begin{equation}
    \begin{split}
        \mathcal{L}_{meta}=min\sum\limits_{\mathcal{T}_i \sim p(\mathcal{T})}w_i\mathcal{L}_i,
      \label{eq:meta-objective}
    \end{split}
\end{equation}
where $\mathcal{L}_i(\theta_i’)$ denotes the loss calculated by feeding the samples in $\mathcal{Q}_i$ into the model. To accomplish few-shot adaptation of $t_k(k=0,...,27)$ and obtain predictive model $f_k$, we conduct optimization in Eq. \ref{eq:inner computation} using the sample vectors present in $t_k$, with a few iterations.

\subsubsection{Yearly landslide susceptibility mapping}
\label{subsubs:3.1.4}
To perform dynamic LSA from 1992 to 2019, we first rasterized Lantau Island at a high resolution of $10 m \times 10 m$ for grid sampling. Next, we followed the sampling approach introduced in Section \ref{subsubs:3.1.1} to produce grid samples. In the years with abundant landslide samples, we trained RF as the predictive models, while in the years with few landslide samples (less than 50) that hardly support the traditional machine learning methods, we leveraged the strength of meta-learning to realize few-shot learning purpose. Finally, we predicted landslide susceptibility map for each year and analyzed the variations in landslide causation underlying dynamics of these maps and the predictive models.

\subsection{Interpretation of predictive model for landslide feature permutation}
\label{subs:3.2}
Current data-driven LSA methods generally train their models based on given datasets and then evaluate the model's performance. Despite the success, few of them take insight into how input landslide features interact with each other and impact the prediction. One of the primary ideas of this article is to interpret the learned models for each yearly LSA and conduct feature permutation to determine the significance of LIFs. Studying the evolution rule of dominant LIFs under the effect of climate change so that we can respond to the challenges of natural disasters brought by global climate anomaly. We applied SHAP \citep{WOS:000446910800009}, a game-theoretical approach that utilized Shapley values to interpret local predictions of machine/deep learning models. The aim was to determine a feature's average contribution score across all instances, as in Eq. 3:
\begin{equation}
    \begin{split}
      \phi_i(v)=min\sum\limits_{S\subseteq \{s_1,s_2...\} \setminus \{s_i\}}{\frac{|S|!(n-|S|-1)!}{n!}(v(S\cup \{s_i\})-v_s(S)) },
      \label{eq:Shapley values}
    \end{split}
\end{equation}
where $n$ represents the number of input landslide features, $s_i$ represents the $i$-th feature of sample $s$, $S$ denotes the subset of all input features, $\vert S \vert$ denotes the features used in all models, and $v_s(S)$ is the predicted value of the sample vector $s$ from $S$.

Specifically, we calculated the Shapley values of input landslide features for each input sample vector following Eq. \ref{eq:Shapley values}. Next, we summed the absolute Shapley values associated with each feature and computed the mean to determine the feature's level of contribution. The input features were sorted in descending order according to their mean absolute Shapley values. Besides the feature permutation, the Shapley values can also uncover the mutual relationship between each pair of LIFs and the effect of each input feature on the model output of a single sample prediction.

\subsection{LSM enhancement and validation by MT-InSAR techniques}
\label{subs:3.3}
\subsubsection{The principle of applied InSAR techniques}
Landslide-prone areas are subjected to temporal decorrelation and atmospheric noise, leading to challenges in accurate InSAR result deduction. Accordingly, we employed a two-tier network utilizing a robust deformation estimator to detect Persistent Scatterer (PS) and Distributed Scatterer (DS) points \citep{SHI2019111231}. The InSAR images were processed to derive the line of sight (LOS) surface movements. In the first-tier network, the most reliable PS points were detected based on strict thresholds. We mainly use amplitude dispersion to select PS candidates. After eliminating the effects of the atmospheric phase screen, the SAR imaging signal model can be expressed as:
\begin{equation}
      y=A\gamma ,
      \label{eq:signal model}
\end{equation}
where $y = [y_1, ..., y_N]^T$ ($N$ is the number of observations, $(\cdot)^T$ is the transpose operation) represents the complex values of differential interferograms, $A$ is the sensing matrix containing the steering vector $a(\Delta h, \Delta v)$ as columns, and $\gamma$ is the reflectivity to be reconstructed. We jointly used the beamforming and robust M-estimator to determine $\Delta h$ and $\Delta v$, which are the relative height and mean deformation velocity, respectively. If the temporal coherence of a PS candidate is larger than a given threshold, we preserve the arc connecting two adjacent candidates and the corresponding $\Delta h$ and $\Delta v$ as the preliminary estimates; otherwise, the arc is rejected. The preliminary estimates were used to unwrap the temporal phase of the preserved arcs. Then, we introduced a robust M-estimator to calculate the final estimates. The M-estimator iteratively reduces the weight of large phase residuals probably due to the larger noise of images and thus improves the robustness of estimation.

\begin{figure}[tbhp]
	\centering
	\begin{subfigure}{0.32\textwidth}
        \includegraphics[width=\textwidth]{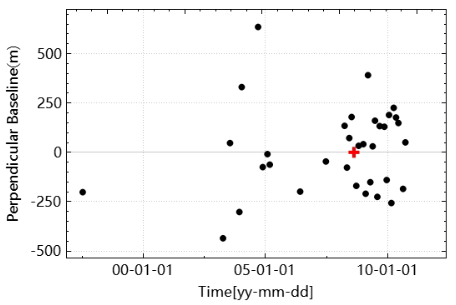}
        \caption{ASAR}
    \end{subfigure}
    \begin{subfigure}{0.32\textwidth}
        \includegraphics[width=\textwidth]{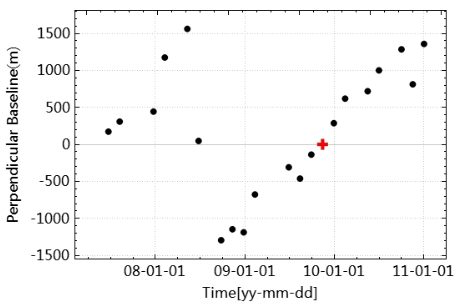}
        \caption{ALOS}
    \end{subfigure}
    \begin{subfigure}{0.32\textwidth}
        \includegraphics[width=\textwidth]{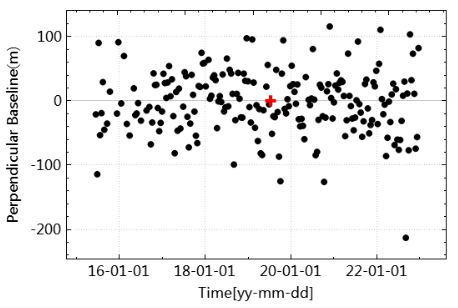}
        \caption{Sentinel-1A}
    \end{subfigure}
	\caption{The distribution of spatial and temporal baselines.}
	\label{fig:baselines}
\end{figure}

After resolving the relative estimates at the preserved arcs, we integrated them by network adjustment and calculated the absolute estimates using one reference point in the study area. To address possible ill-conditioned problems of the inversion of the adjustment matrix, we applied a ridge estimator to integrate relative estimates. A more detailed description of the verification of the robustness of the M-estimator and ridge estimator was given in \cite{7329983}.
In the second-tier network, we used the PS detected in the first-tier network as reference points and extended the first-tier network by detecting the remaining PS and all DS points based on relatively less strict thresholds. For the detection of DS points, the Kolmogorov-Smirnov test was first applied to identify statistically homogeneous pixels and then the complex covariance matrix (CCM) was calculated. SqueeSAR uses the Broyden-Fletcher-Goldfarb-Shanno algorithm to reconstruct the optimal phase, which requires the inversion of CCM. The problem is that CCM is rank deficient and thus the inversion is not stable when the SHP number is less than $N$. To address this issue, we revised the phase optimization problem \citep{MA2019111282} and improved the robustness of estimation by assigning a larger weight to the higher-coherence phase and avoiding matrix inversion. Then, we used a more efficient phase-linking method to obtain the optimal phase \citep{4685949}, which is called a coherence-weighted phase-linking method. A more detailed description of the phase-linking method in this study is presented in \cite{ZHANG2019157}. Finally, we employed the reconstructed optimal phase to identify DS points by temporal coherence threshold.

\subsubsection{Landslide susceptibility enhancement and validation}
\label{subsubs:3.3.2}
Almost all recorded landslides in Lantau Island are identified through the interpretation of optical images and on-site investigations, resulting in a data-driven model prone to predict high landslide susceptibility under post-landslide geographical environment, but insensitive to pre-failures with slope dynamics. In fact, slope dynamics also matter as many major landslides are preceded by signs of slow movement \citep{BEKAERT2020111983}. Therefore, this study applied MT-InSAR techniques for the detection and monitoring of the slow-moving slopes in Lantau Island and conducted LSM enhancement. The enhancement of the initial landslide susceptibility map involves the following screening, leveling, interpolation, and LSM refinement steps: (1) Exclude InSAR points located on slopes facing south or north, as they are more sensitive to displacement in the eastern or western directions; and exclude InSAR points located on slopes with a low gradient. In this study, we excluded the measurement points in the slope aspect ranging from (-11.25$^{\circ}$ - 11.25$^{\circ}$) and (168.75$^{\circ}$ - 191.25$^{\circ}$) from this study. (2) Assign a deformation level for each InSAR measurement point based on the magnitude of deformation velocity. Specifically, we defined the points with a velocity ranging from (-$\infty$, -10) mm/year as level 4; those ranging from (-10, -8) mm/year as level 3; those ranging from (-8, -4) mm/year as level 2; those ranging from (-4, -2) mm/year as level 1; and all other points as level 0. (3) Assign a deformation level for each mapping unit using the nearest neighbor interpolation method. (4) Enhance initial LSM with a deformation level map using the matrix in Table \ref{tab:matrix}. 

\begin{table*}[h]
    \caption{LSM enhancement matrix. L0-L4 represents initial LSM level (very low to very high susceptibility); D0-D4 is the defined deformation level. The table values represent the final enhanced landslide susceptibility of the mapping units.}
    \centering
    \footnotesize
    \renewcommand\arraystretch{1.25}  
    \setlength\tabcolsep{7.5pt}  
    \begin{tabular}{cccccc}
        \hline
        Level      &L0                        &L1                        &L2                        &L3                        &L4                    \\ \hline
        D0         &\cellcolor{green!30}0     &\cellcolor{lime!50}1      &\cellcolor{yellow!30}2    &\cellcolor{orange!30}3    &\cellcolor{red!30}4   \\ \hline
        D1         &\cellcolor{lime!50}1      &\cellcolor{lime!50}1      &\cellcolor{yellow!30}2    &\cellcolor{orange!30}3    &\cellcolor{red!30}4   \\ \hline
        D2         &\cellcolor{yellow!30}2    &\cellcolor{yellow!30}2    &\cellcolor{orange!30}3    &\cellcolor{orange!30}3    &\cellcolor{red!30}4   \\ \hline
        D3         &\cellcolor{orange!30}3    &\cellcolor{orange!30}3    &\cellcolor{red!30}4       &\cellcolor{red!30}4       &\cellcolor{red!30}4   \\ \hline
        D4         &\cellcolor{red!30}4       &\cellcolor{red!30}4       &\cellcolor{red!30}4       &\cellcolor{red!30}4       &\cellcolor{red!30}4   \\ \hline
    \end{tabular}
    \label{tab:matrix}
\end{table*}

Slope movement is a representation of the interaction of landslide-inducing factors. Landslide susceptibility assesses the likelihood of landslide occurrence under the interaction of these factors. Based on this, cross-validation can be performed between ground deformation maps and landslide susceptibility maps.

\subsection{Implementation details}
\label{subs:implementation details}

\textbf{Non-landslide samples.}
We selected an equal number of non-landslide points as landslide points through ArcGIS delineation. 50\% of the points were randomly sampled, while the rest were selected based on prior knowledge and statistical data. We prioritized points from flat and urban areas, as well as major points in lithology and land categories that historically had a low frequency of landslides. These non-landslide points correspond one-to-one with the landslide points after shuffling, and were featured by sampling static and dynamic thematic information for corresponding years.

\textbf{The data-driven model settings.}
The model was constructed and trained in the \textit{TensorFlow} environment ($tf=2.10$, $python=3.9$). Initially, during the meta-learning stage, the inner and outer (meta-) learning rates were set at 0.01 and 0.001, respectively. It’s noteworthy that the inner learning rate can be set trainable. The number of meta-training iterations was set to 3000. We conducted 5 gradient update and used half of the samples in the subtask for inner optimization.  Both the inner and outer batch sizes were set to 16. For the few-shot adaptation of task $t_i$, all the samples in task $t_i$ participated the fine-tuning process, applying 5 gradient updates for optimization. More details are available at \url{https://github.com/CLi-de/D_LSM}.

\section{Results}
\label{Results}
\subsection{Initial LSM based on LIFs}
\begin{figure}[tbhp]
    \centering
    \begin{subfigure}{0.242\textwidth}
        \includegraphics[width=\textwidth]{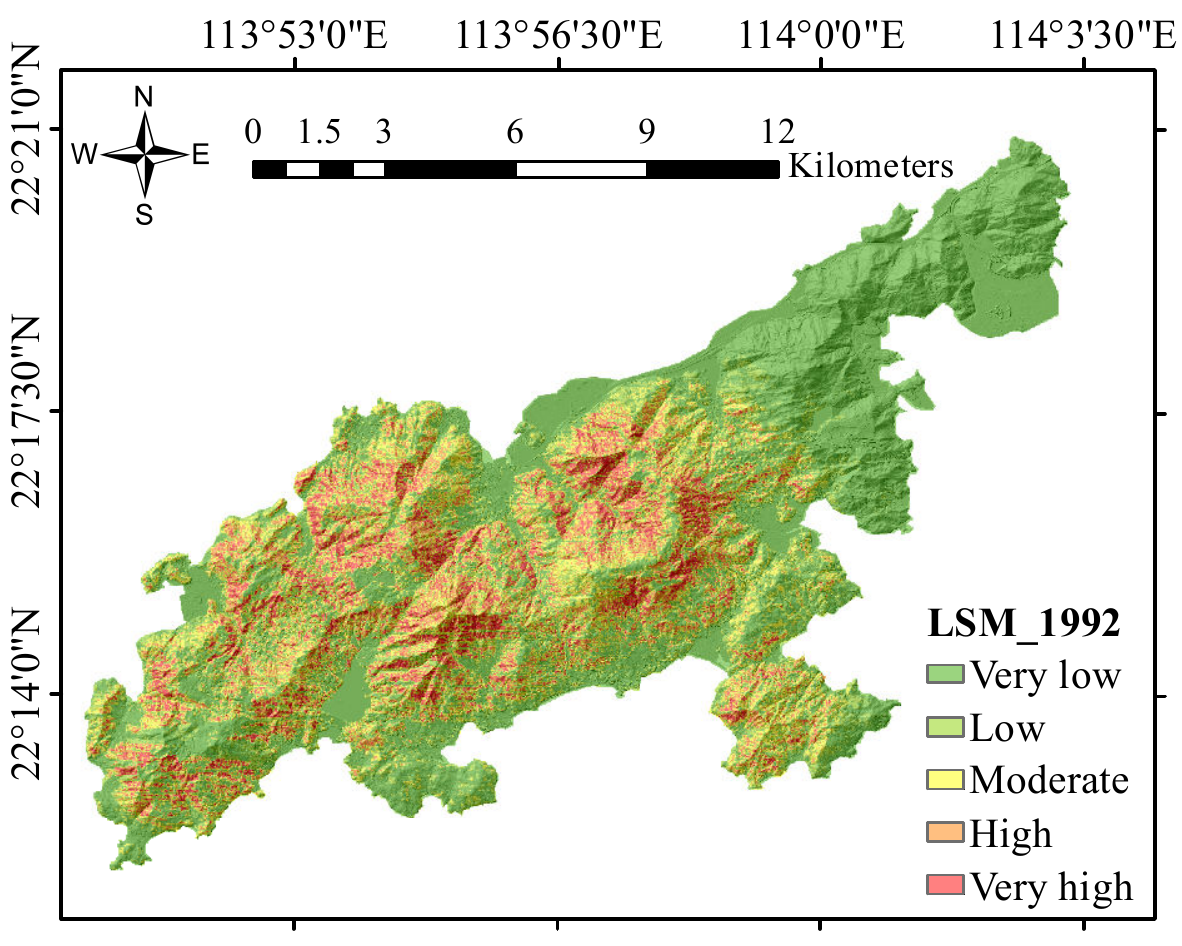}
    \end{subfigure}
    \begin{subfigure}{0.242\textwidth}
        \includegraphics[width=\textwidth]{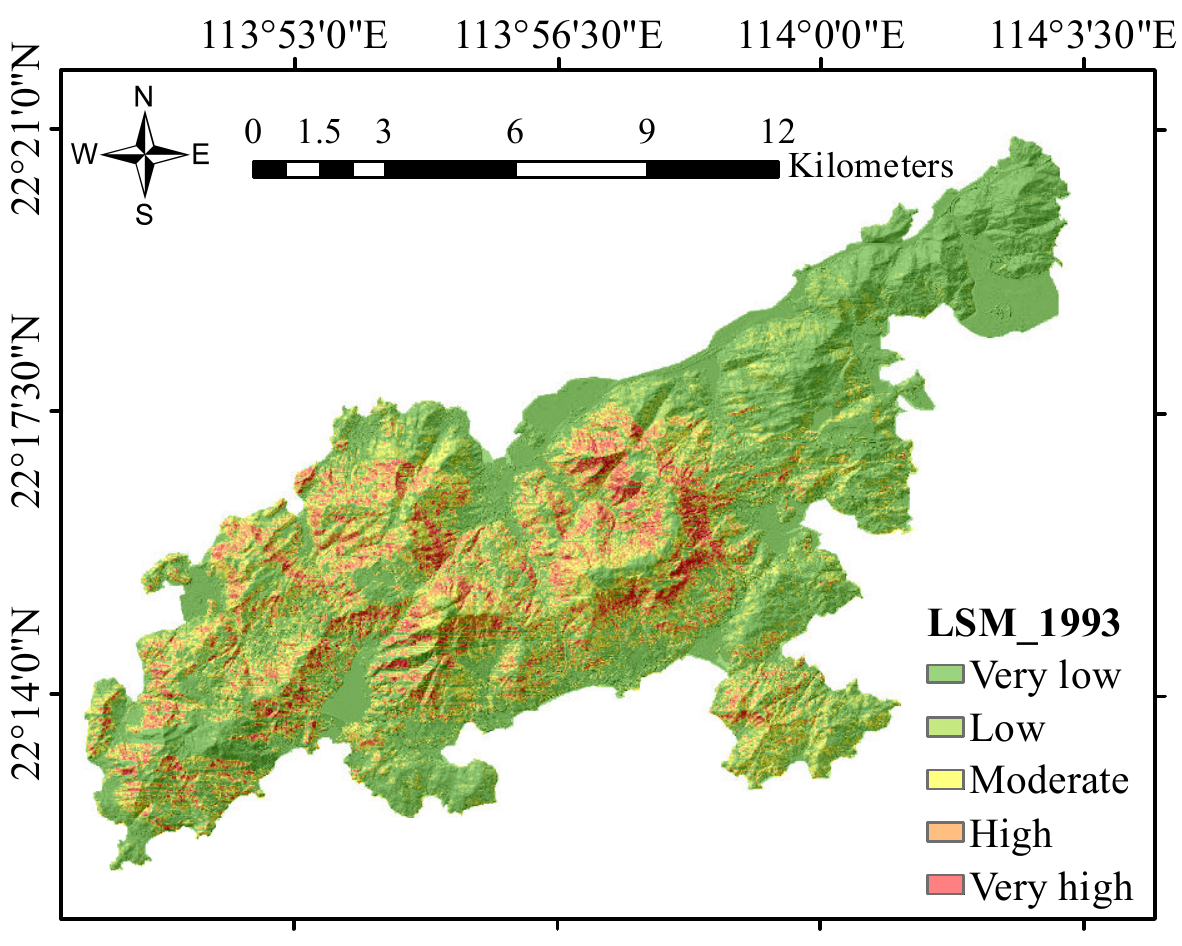}
    \end{subfigure}
	\begin{subfigure}{0.242\textwidth}
        \includegraphics[width=\textwidth]{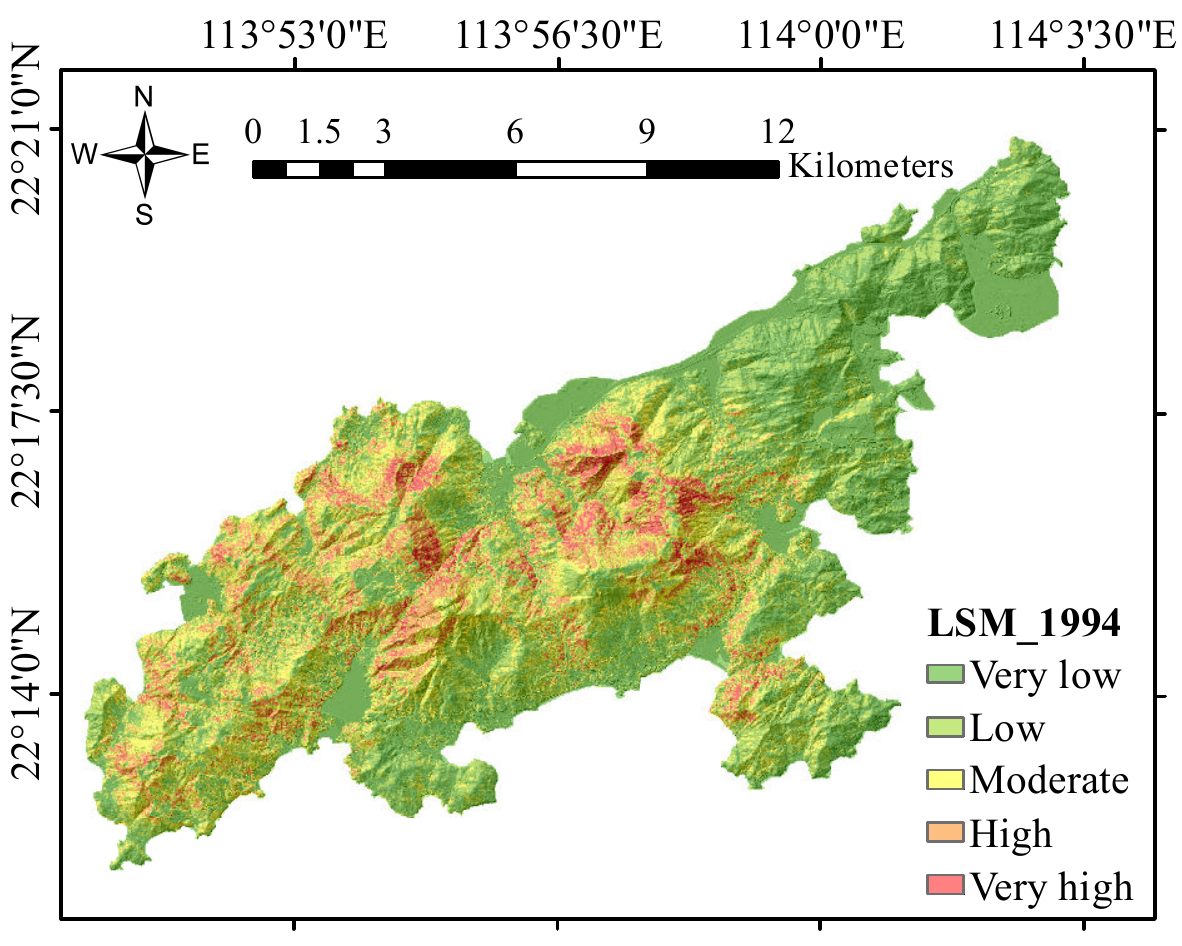}
    \end{subfigure}
    \begin{subfigure}{0.242\textwidth}
        \includegraphics[width=\textwidth]{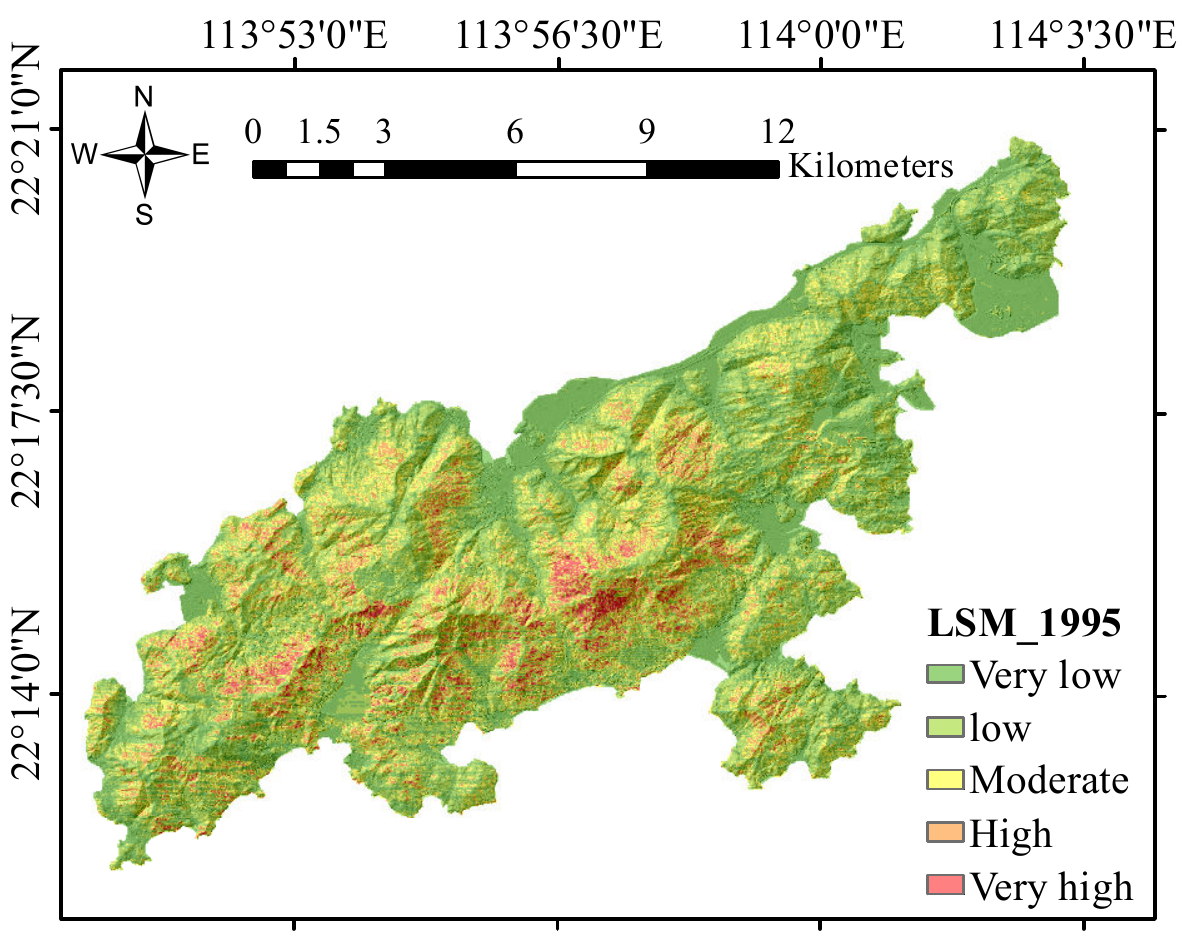}
    \end{subfigure}

    \begin{subfigure}{0.242\textwidth}
        \includegraphics[width=\textwidth]{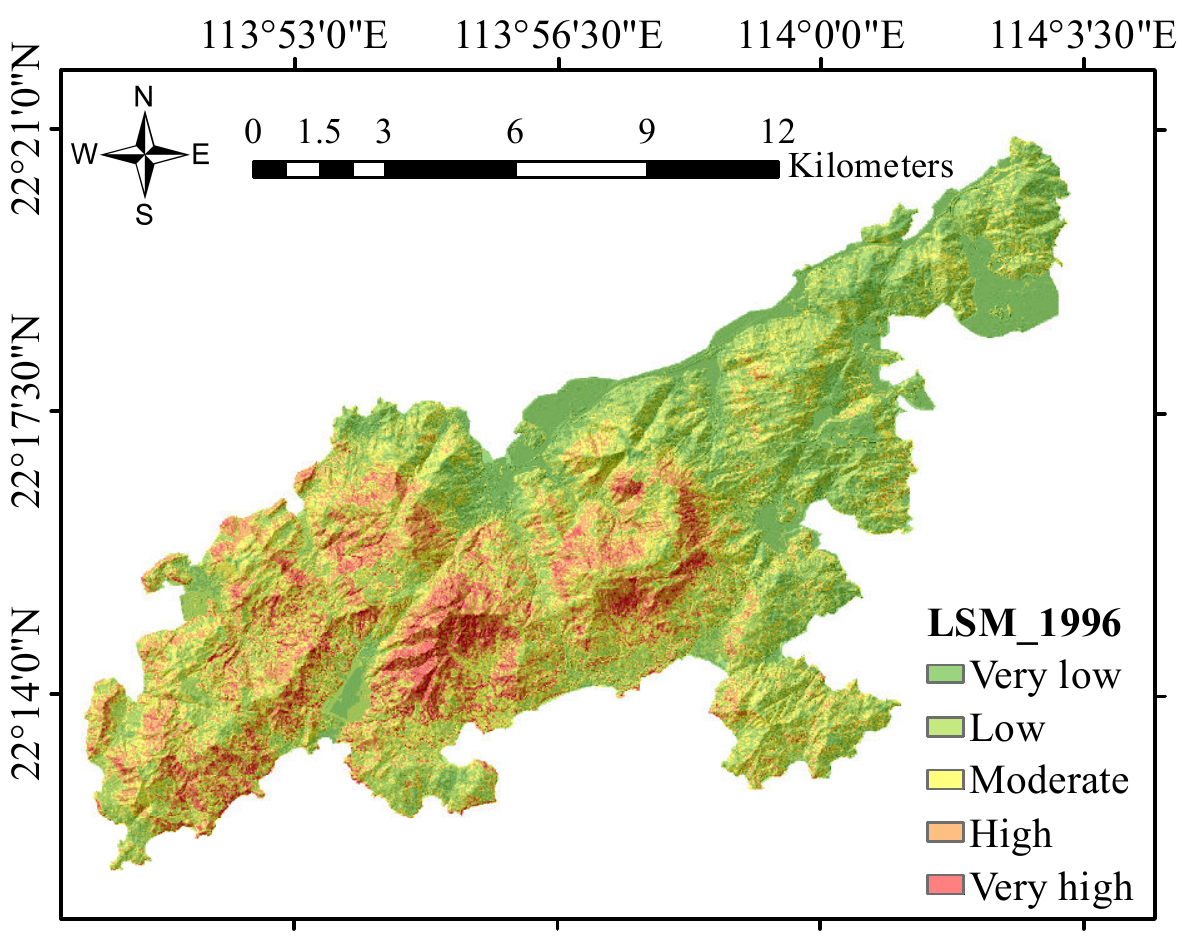}
    \end{subfigure}
    \begin{subfigure}{0.242\textwidth}
        \includegraphics[width=\textwidth]{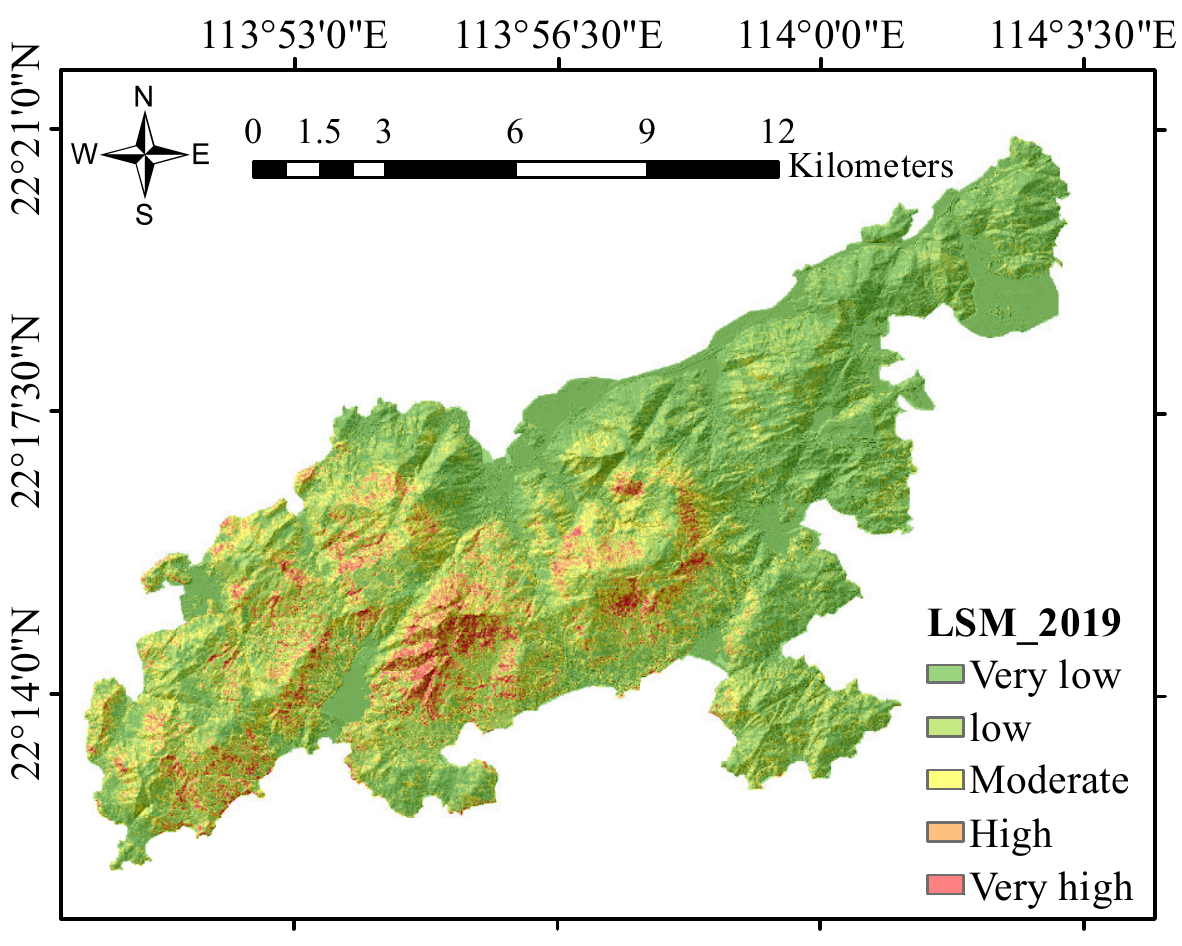}
    \end{subfigure}
	\begin{subfigure}{0.242\textwidth}
        \includegraphics[width=\textwidth]{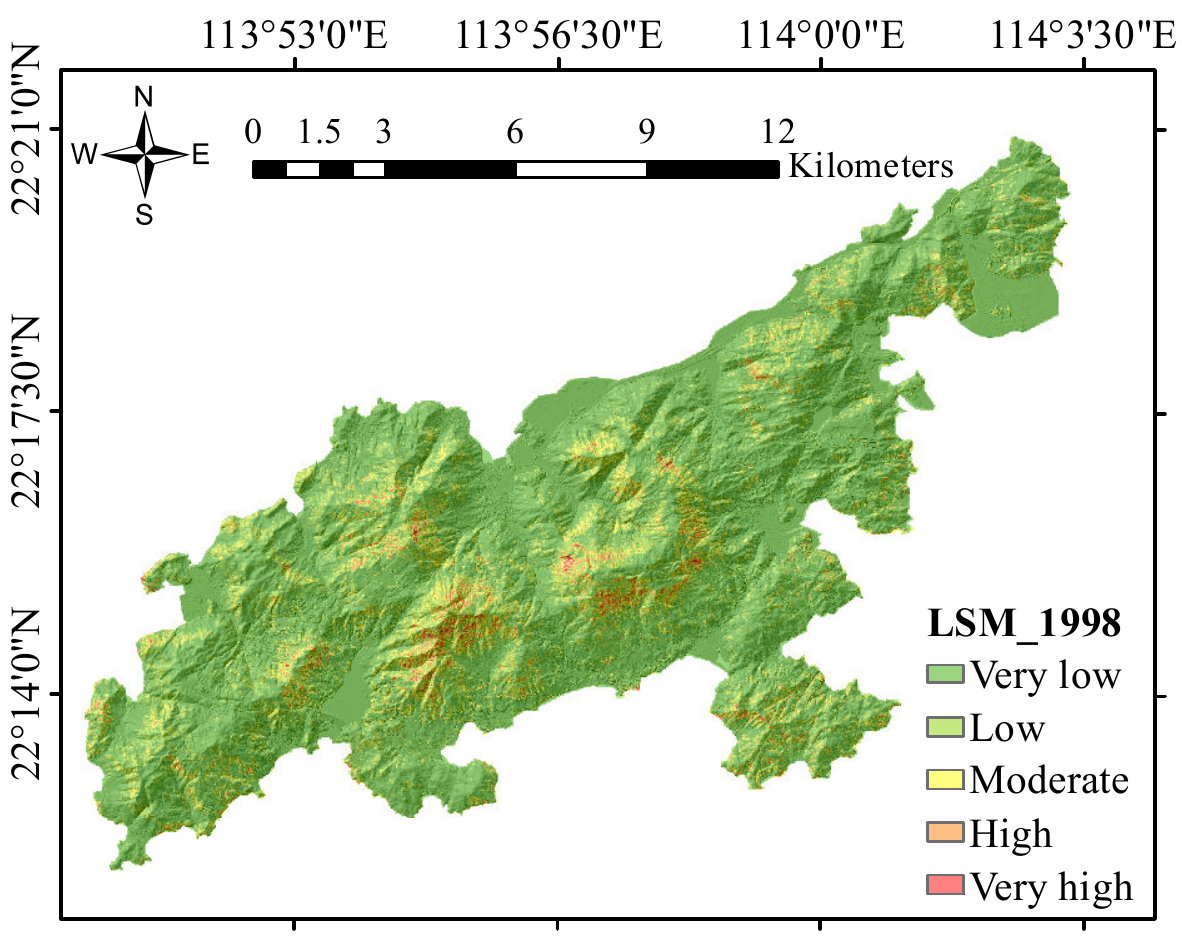}
    \end{subfigure}
    \begin{subfigure}{0.242\textwidth}
        \includegraphics[width=\textwidth]{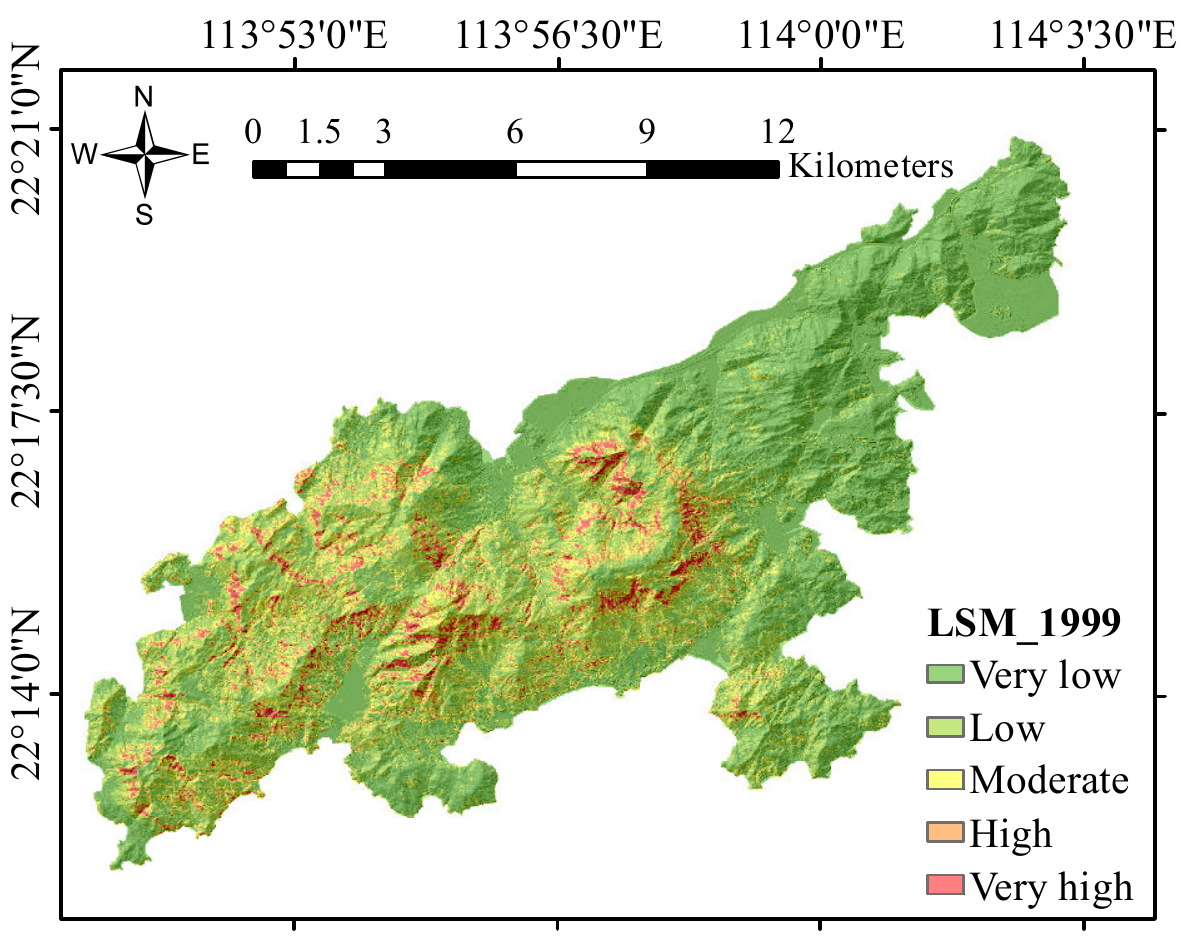}
    \end{subfigure}

    \begin{subfigure}{0.242\textwidth}
        \includegraphics[width=\textwidth]{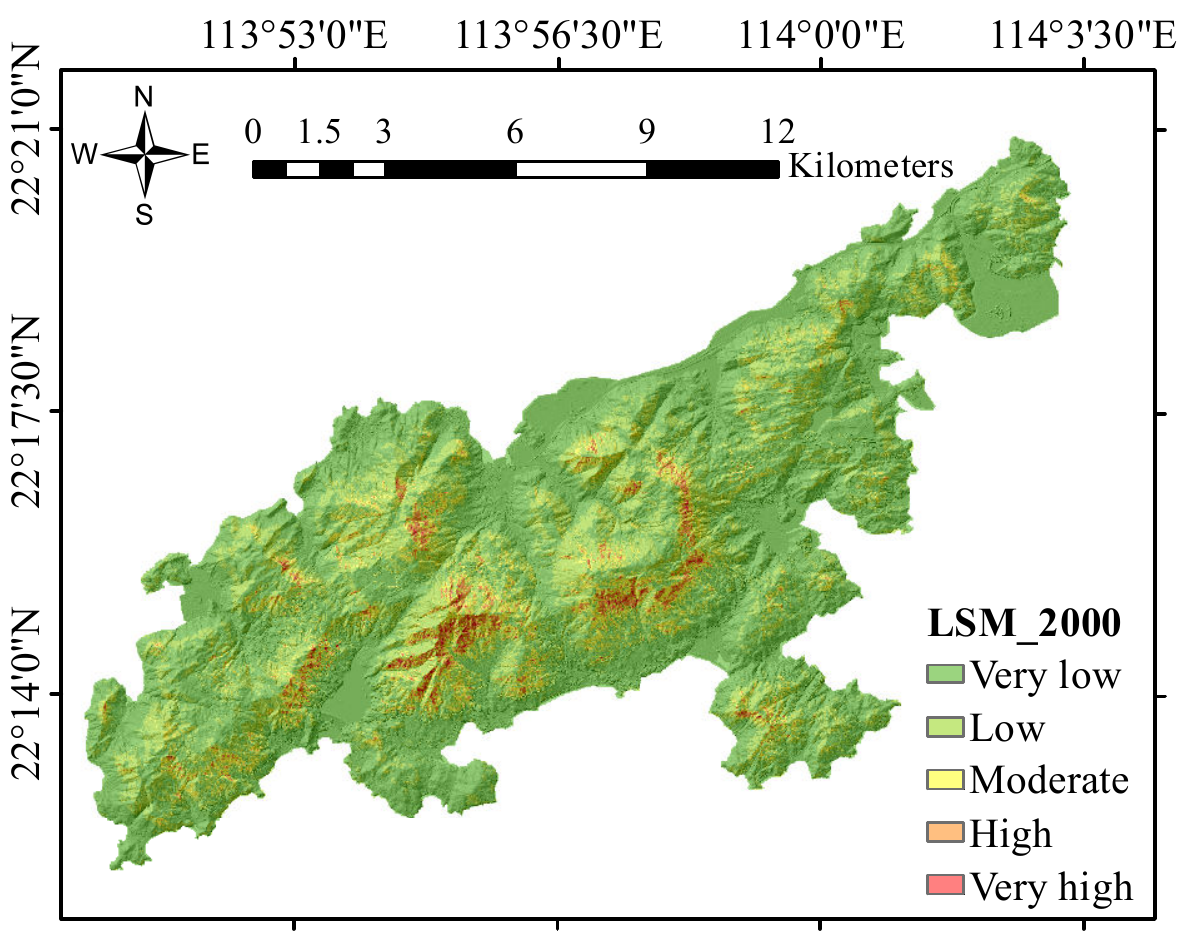}
    \end{subfigure}
    \begin{subfigure}{0.242\textwidth}
        \includegraphics[width=\textwidth]{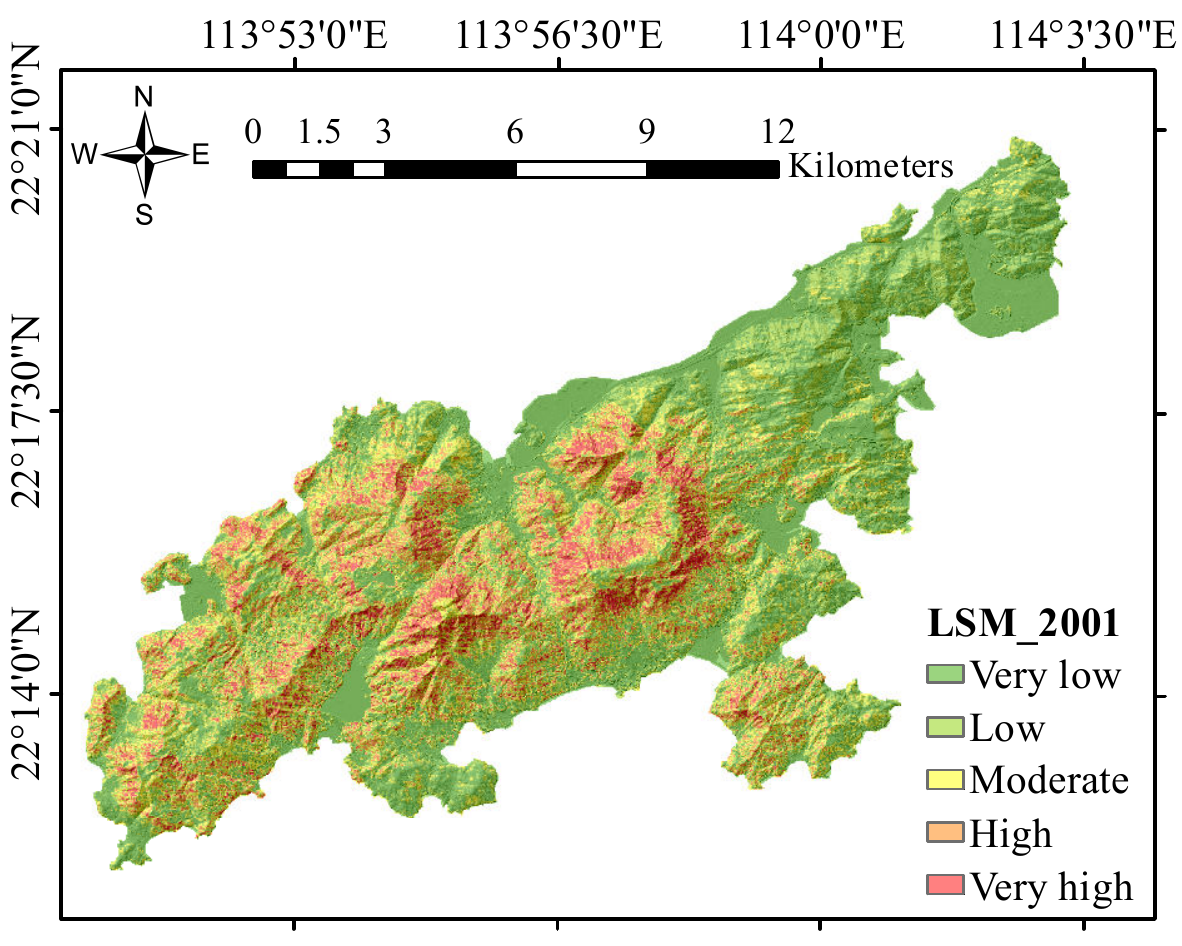}
    \end{subfigure}
	\begin{subfigure}{0.242\textwidth}
        \includegraphics[width=\textwidth]{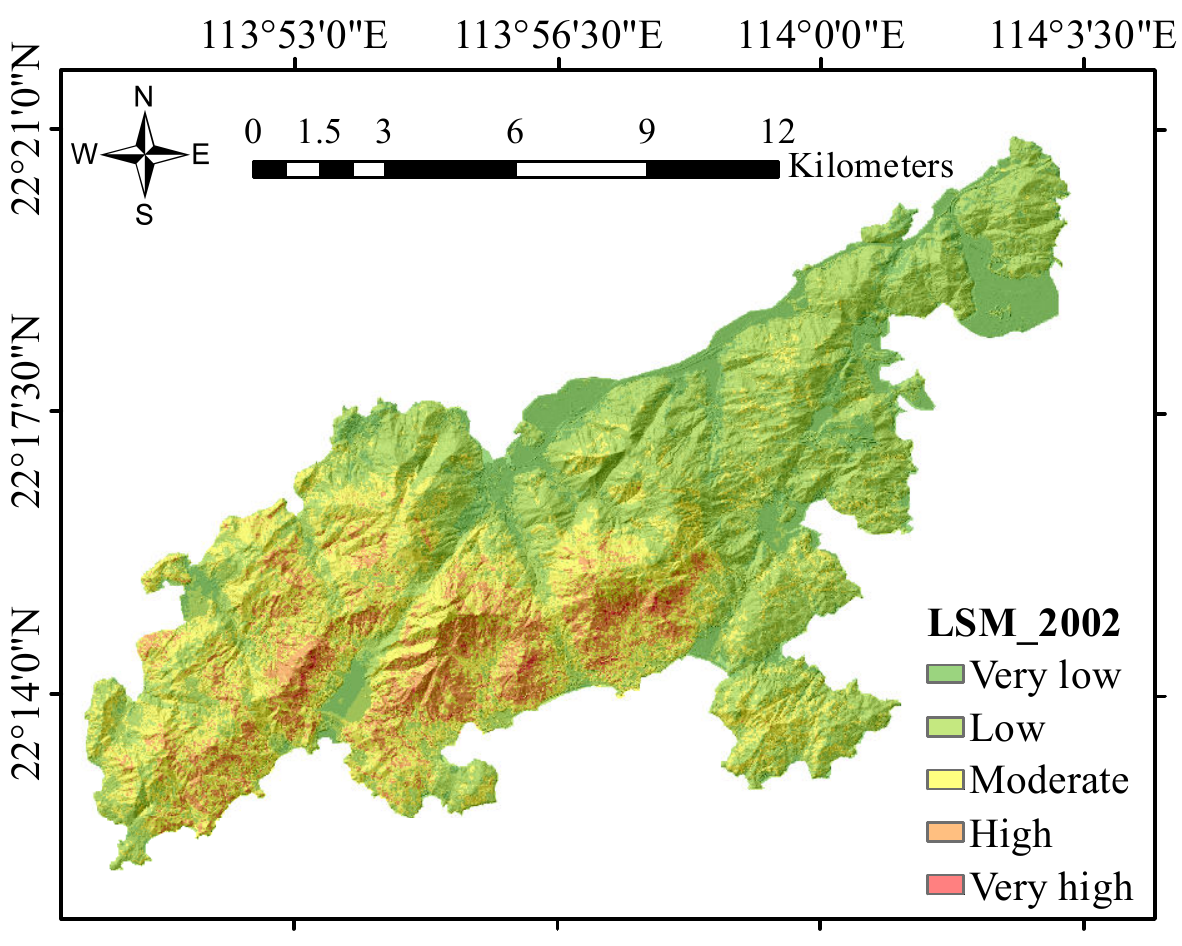}
    \end{subfigure}
    \begin{subfigure}{0.242\textwidth}
        \includegraphics[width=\textwidth]{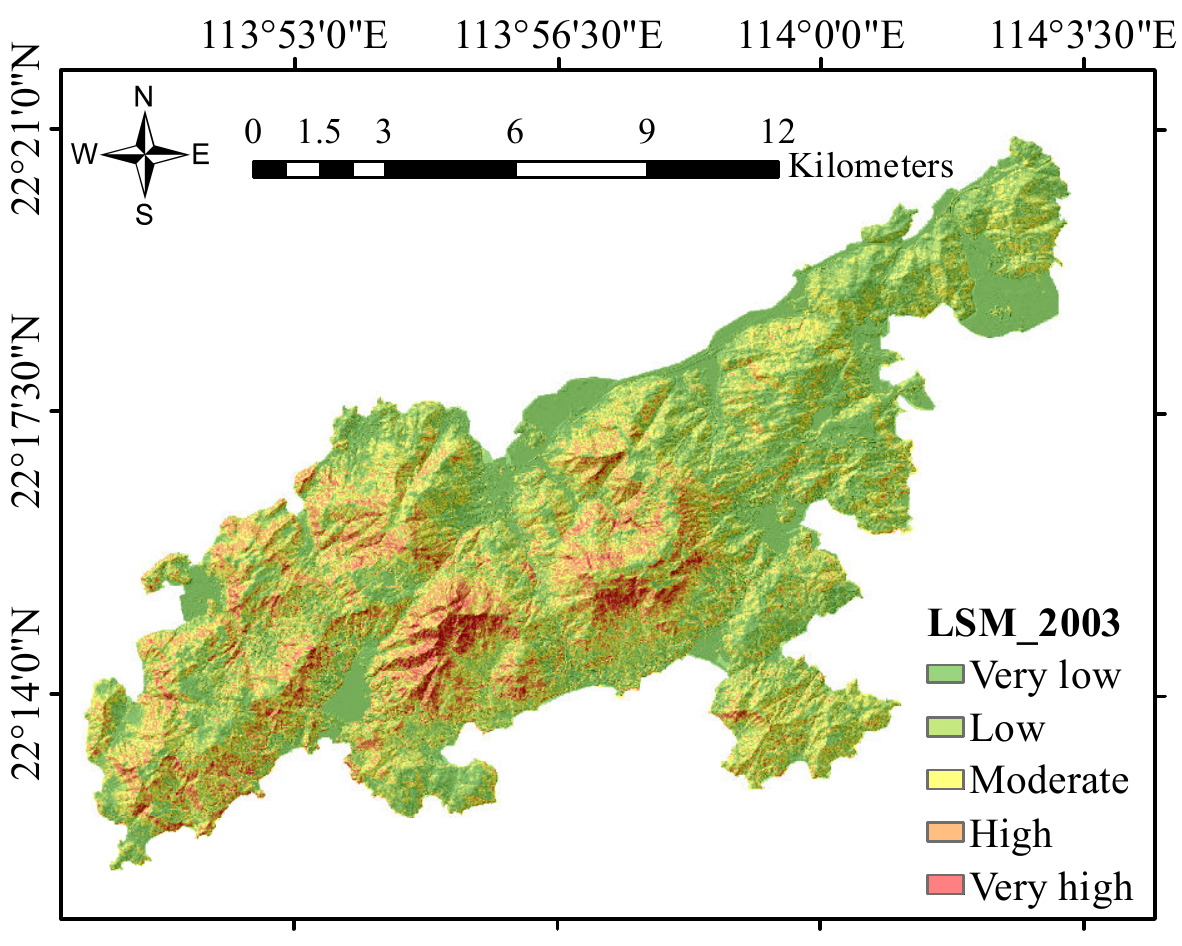}
    \end{subfigure}

    \begin{subfigure}{0.242\textwidth}
        \includegraphics[width=\textwidth]{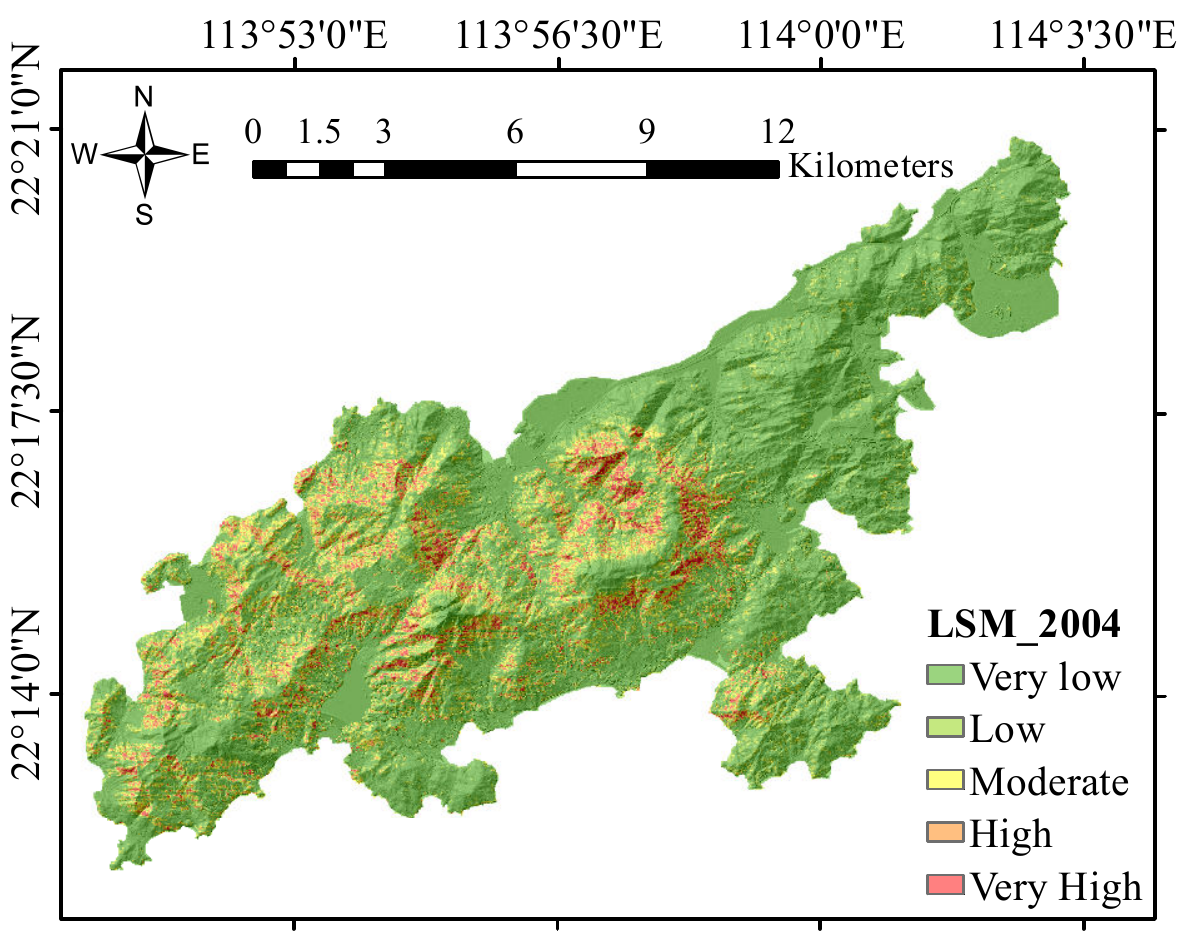}
    \end{subfigure}
    \begin{subfigure}{0.242\textwidth}
        \includegraphics[width=\textwidth]{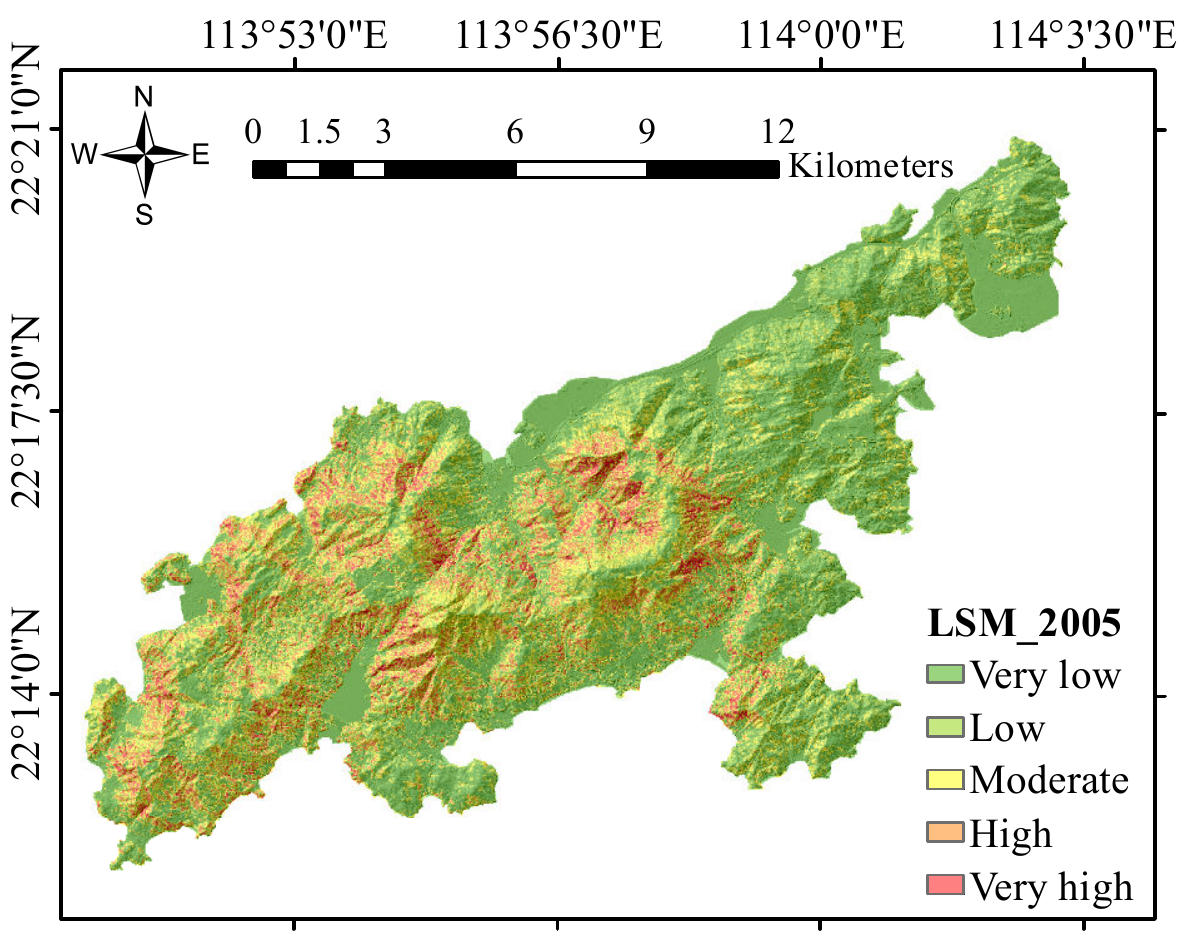}
    \end{subfigure}
	\begin{subfigure}{0.242\textwidth}
        \includegraphics[width=\textwidth]{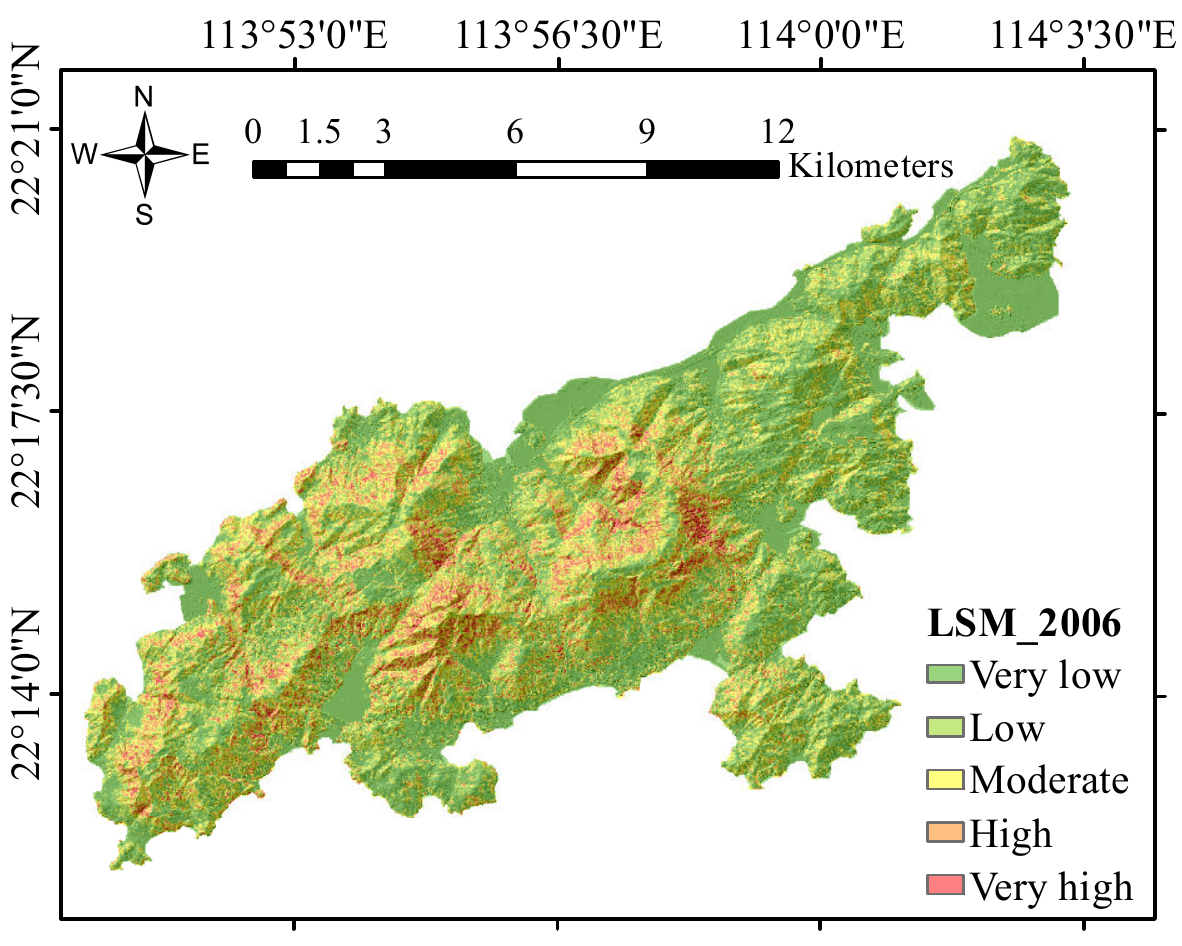}
    \end{subfigure}
    \begin{subfigure}{0.242\textwidth}
        \includegraphics[width=\textwidth]{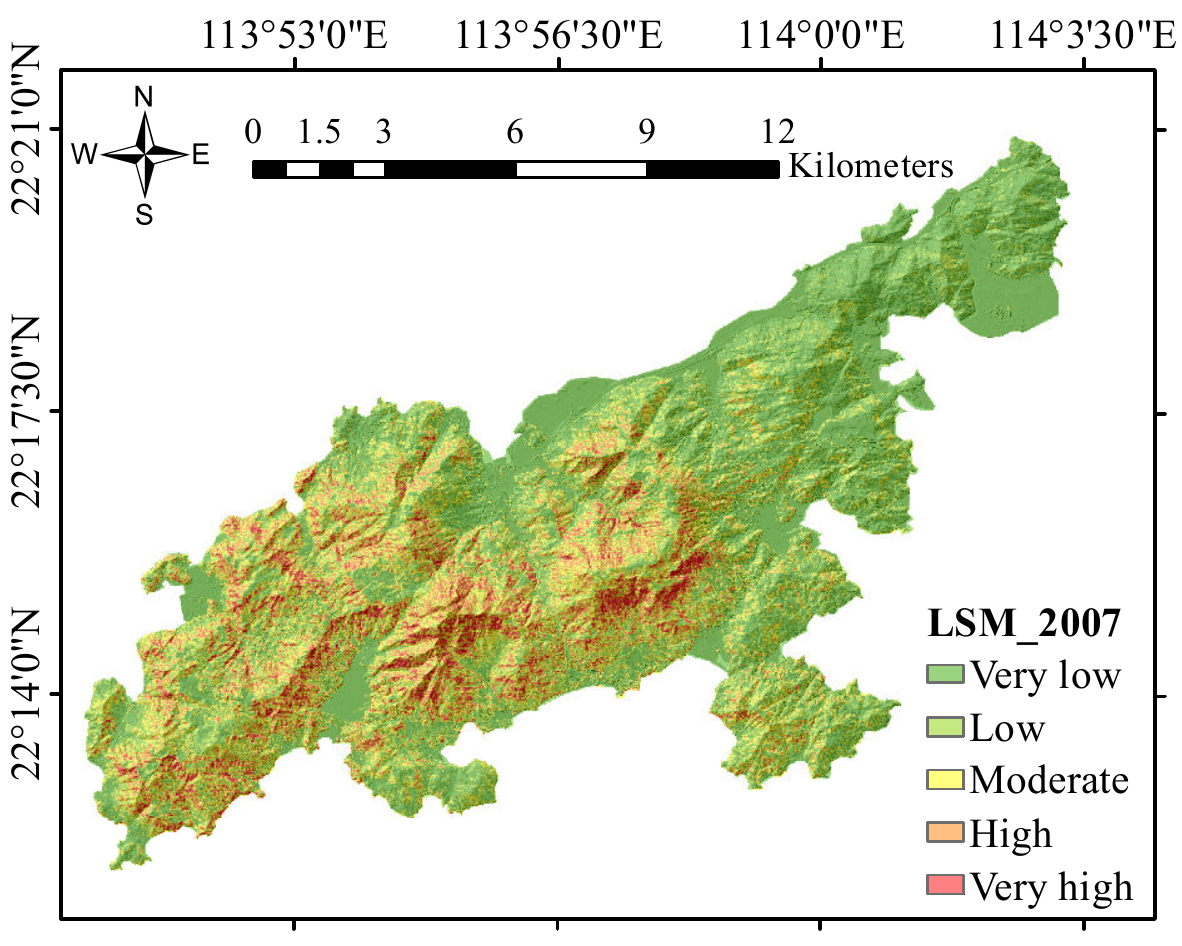}
    \end{subfigure}

    \begin{subfigure}{0.242\textwidth}
        \includegraphics[width=\textwidth]{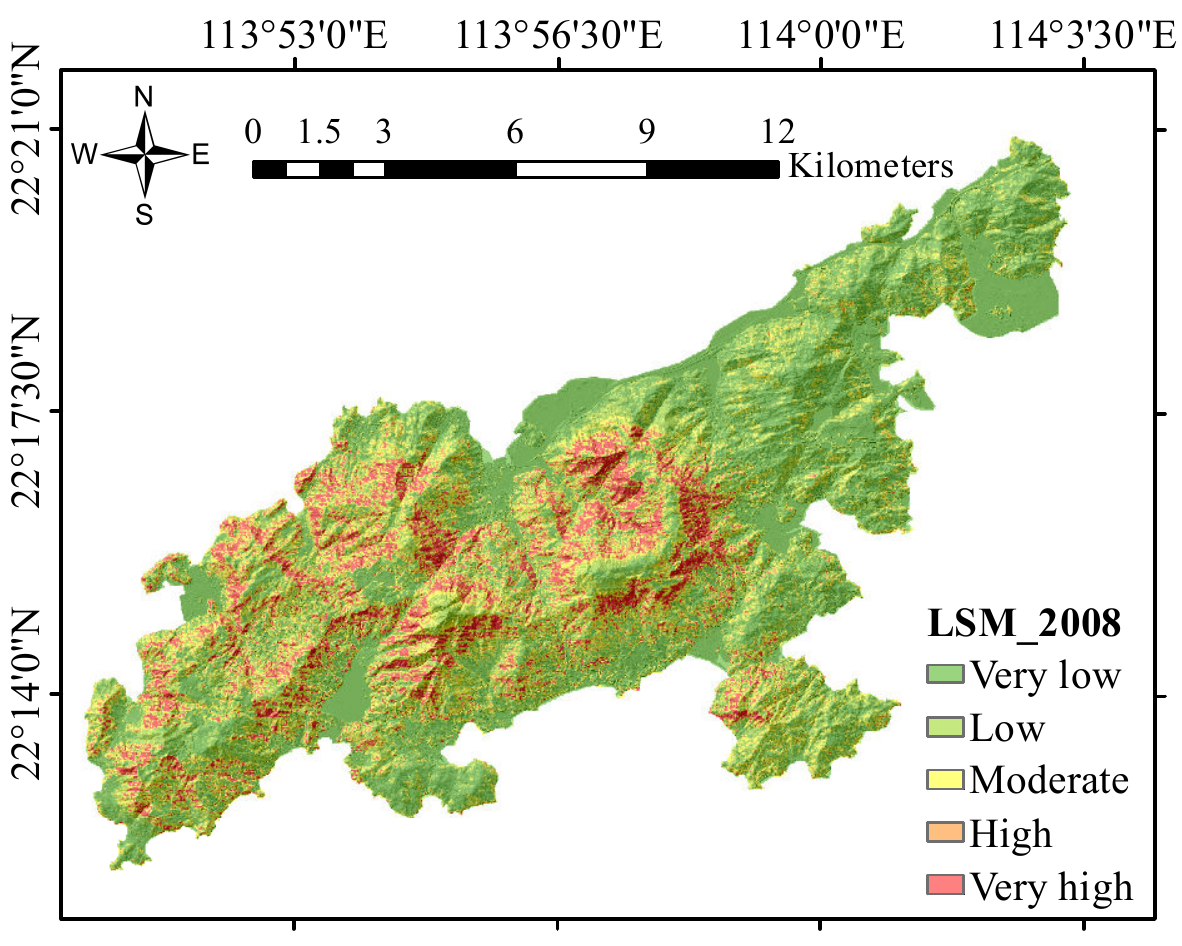}
    \end{subfigure}
    \begin{subfigure}{0.242\textwidth}
        \includegraphics[width=\textwidth]{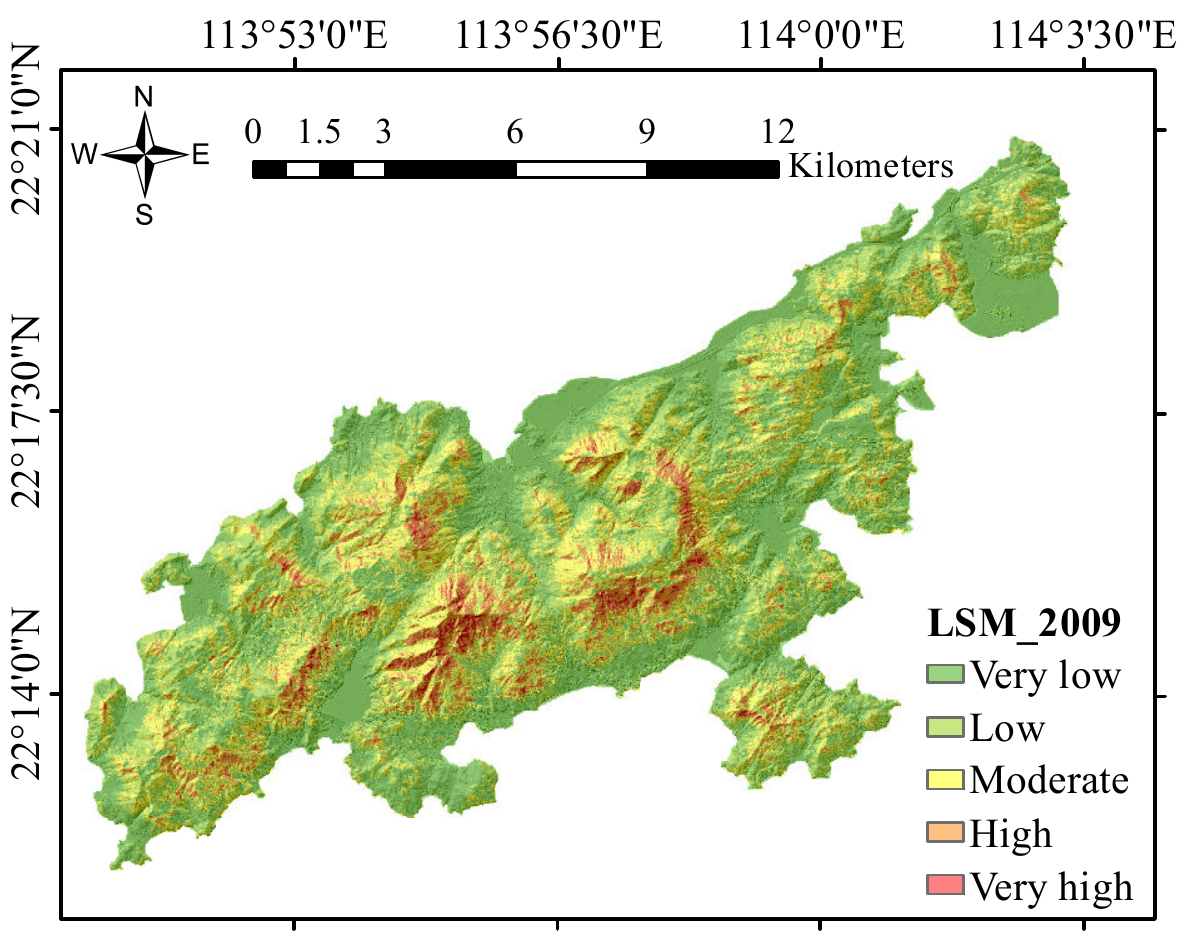}
    \end{subfigure}
	\begin{subfigure}{0.242\textwidth}
        \includegraphics[width=\textwidth]{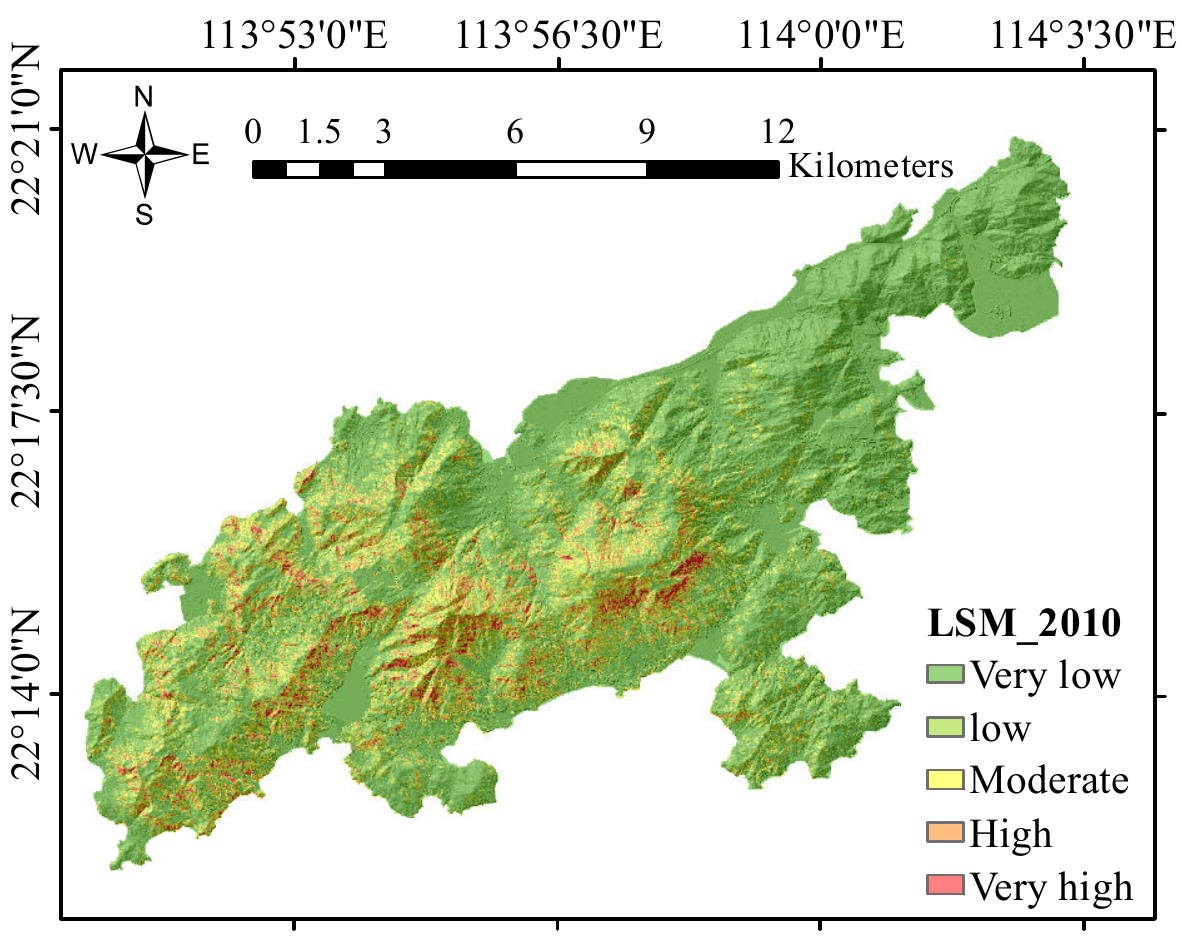}
    \end{subfigure}
    \begin{subfigure}{0.242\textwidth}
        \includegraphics[width=\textwidth]{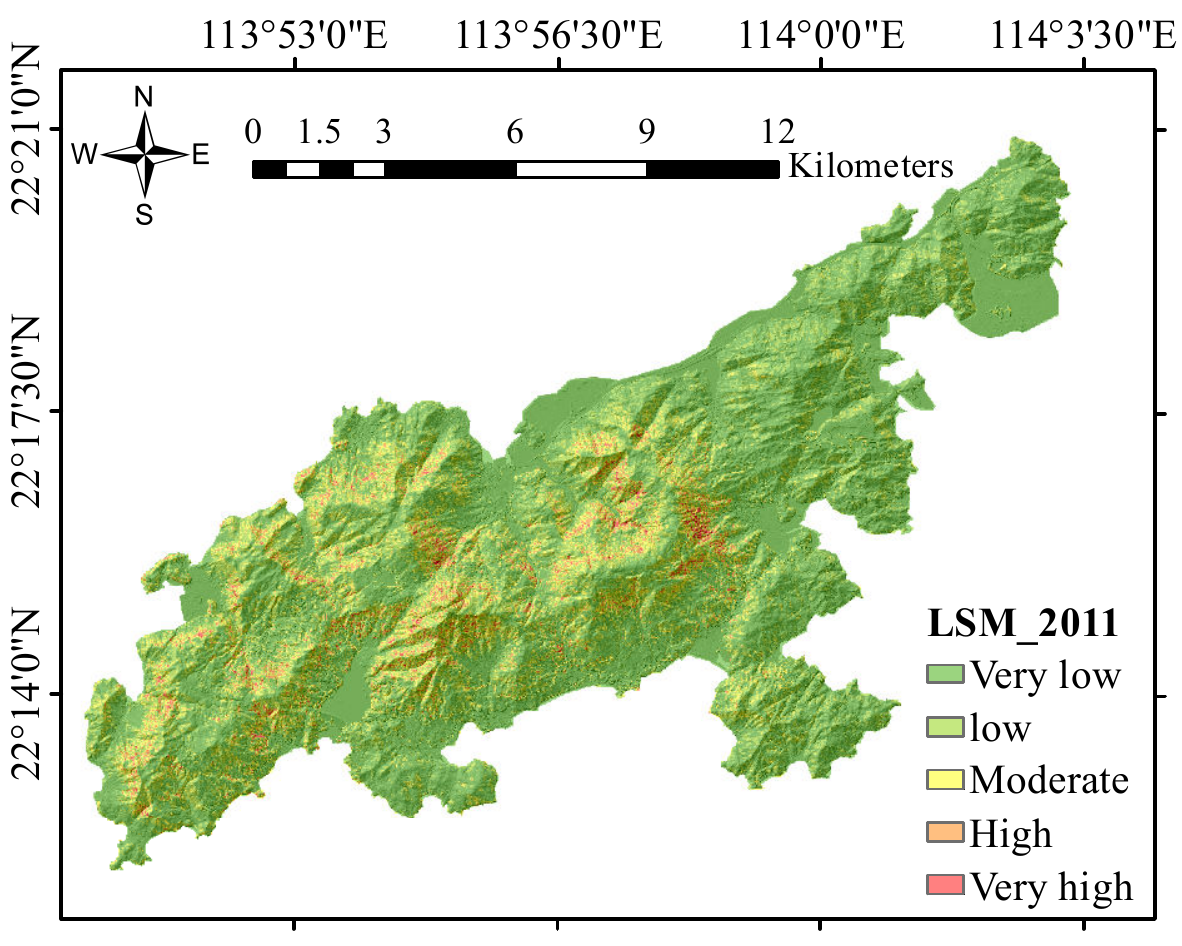}
    \end{subfigure}

    \begin{subfigure}{0.242\textwidth}
        \includegraphics[width=\textwidth]{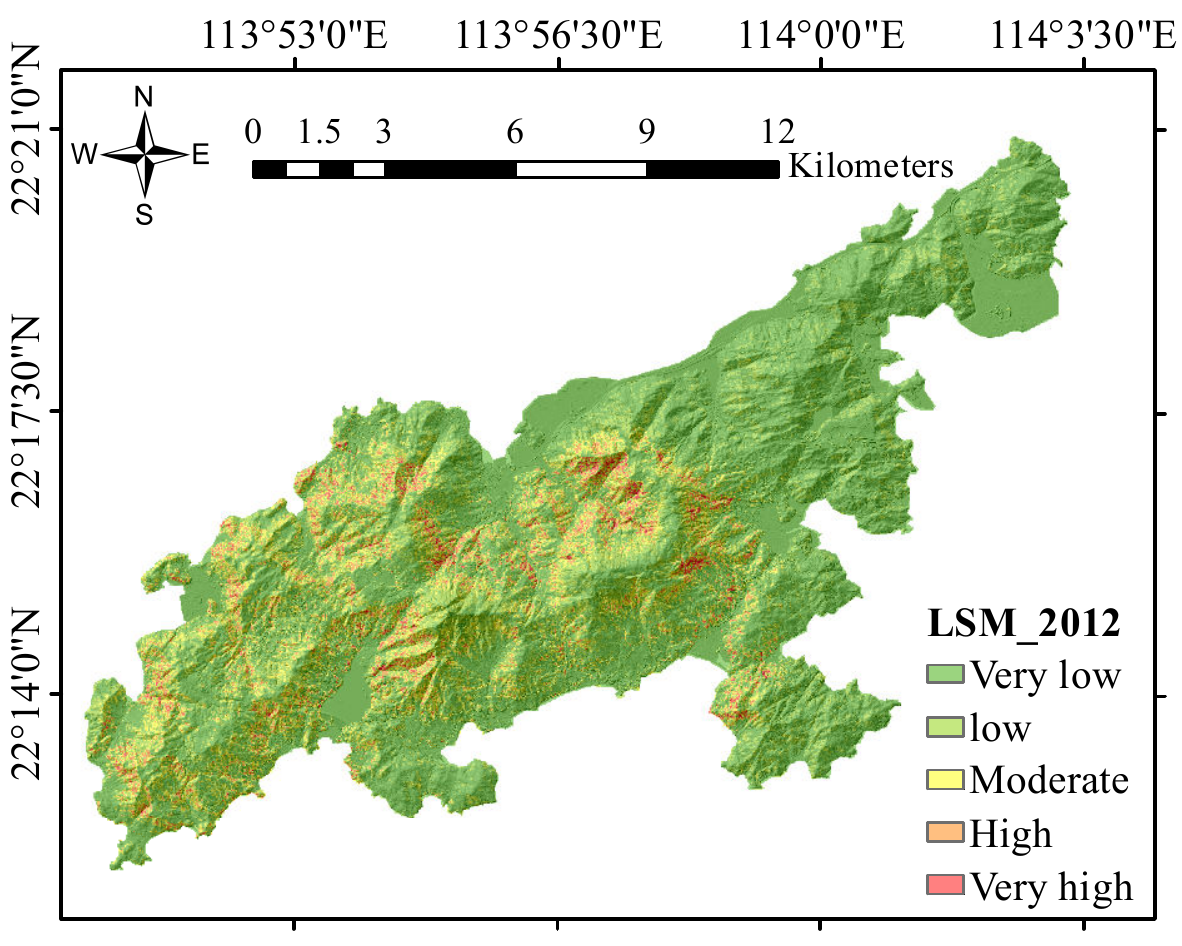}
    \end{subfigure}
    \begin{subfigure}{0.242\textwidth}
        \includegraphics[width=\textwidth]{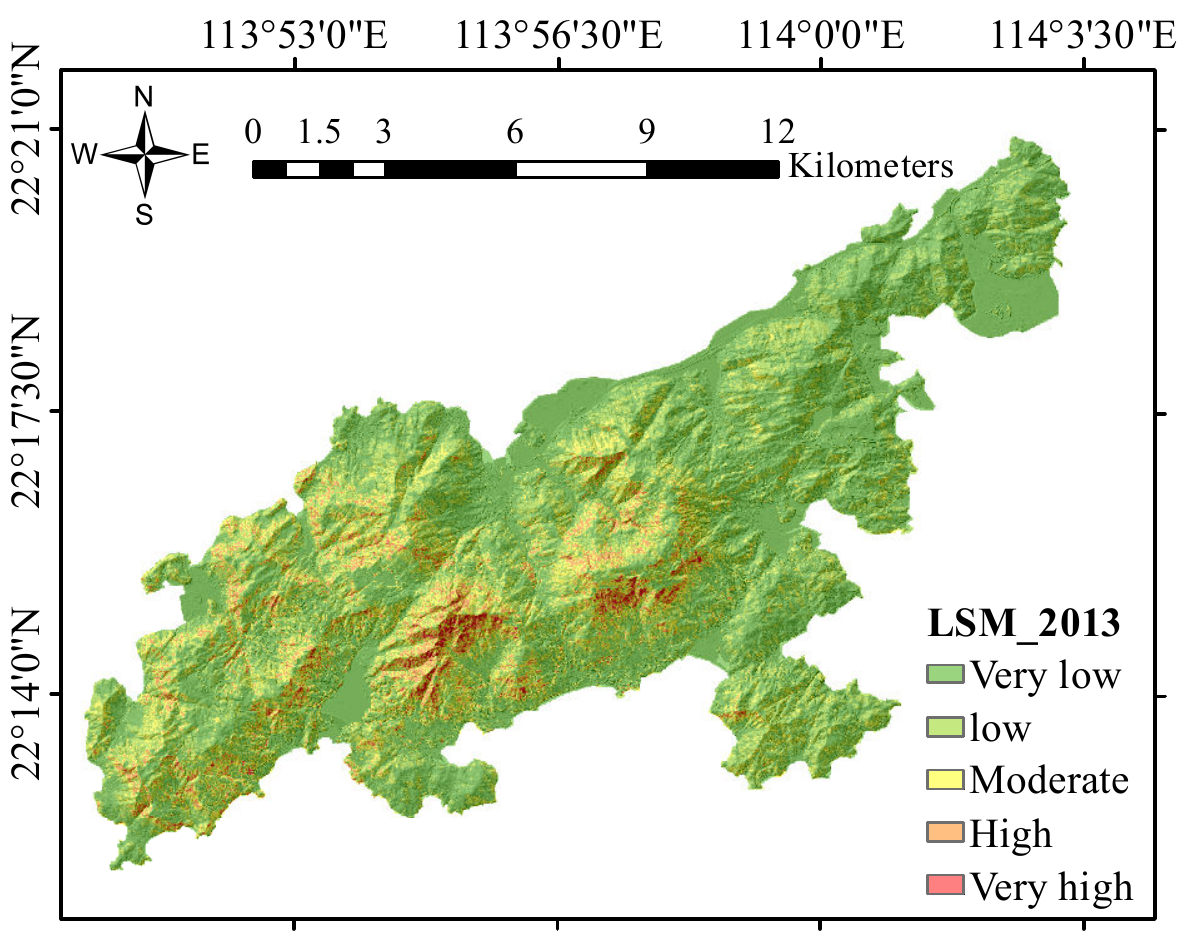}
    \end{subfigure}
	\begin{subfigure}{0.242\textwidth}
        \includegraphics[width=\textwidth]{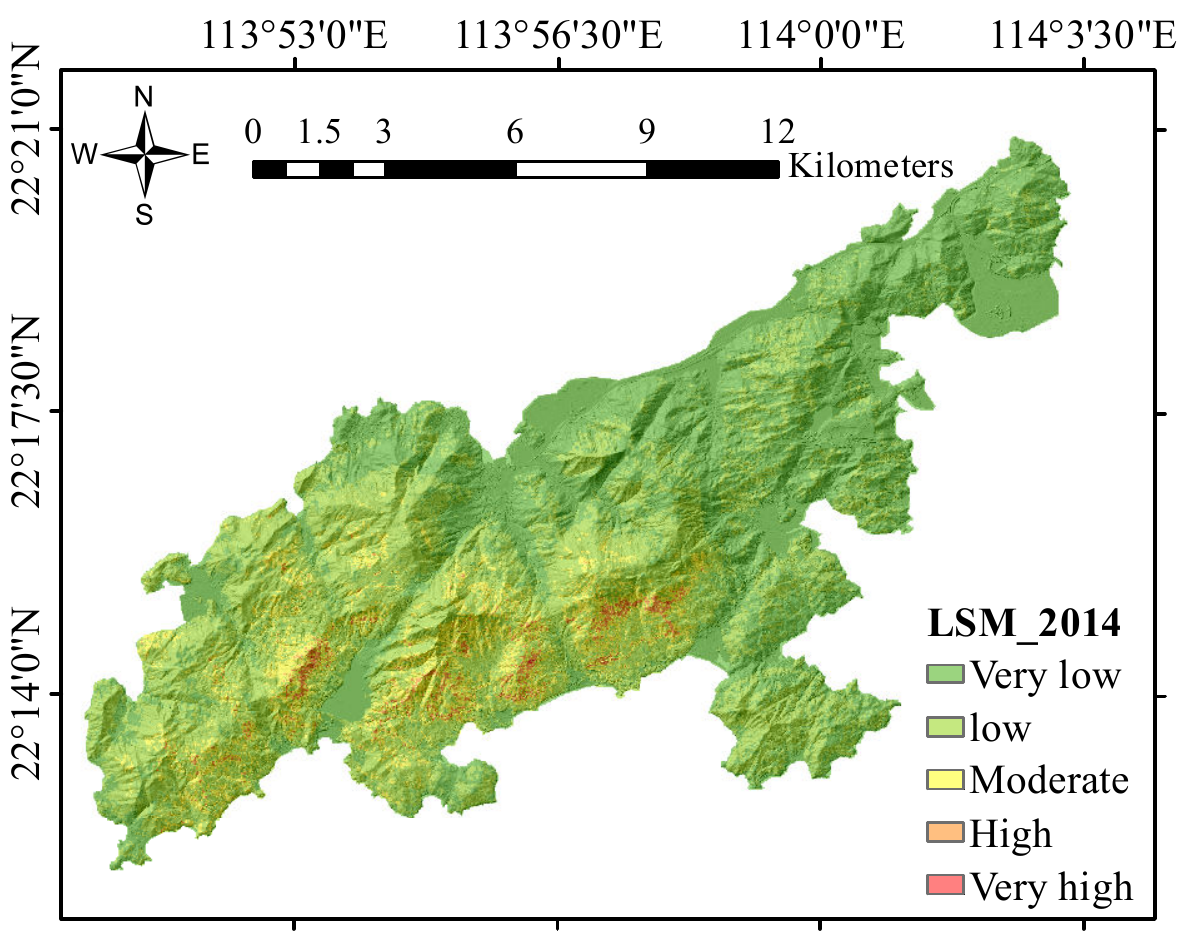}
    \end{subfigure}
    \begin{subfigure}{0.242\textwidth}
        \includegraphics[width=\textwidth]{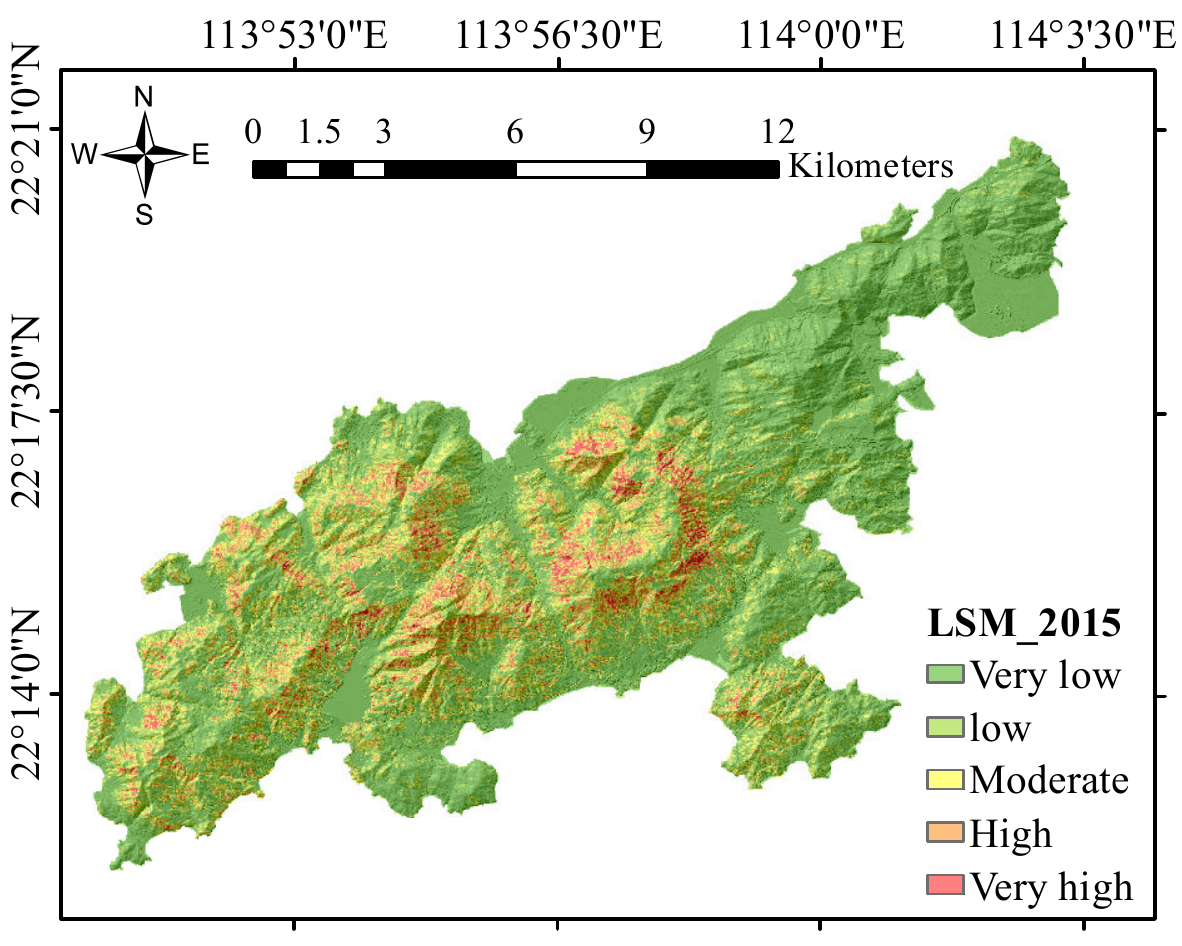}
    \end{subfigure}

    \begin{subfigure}{0.242\textwidth}
        \includegraphics[width=\textwidth]{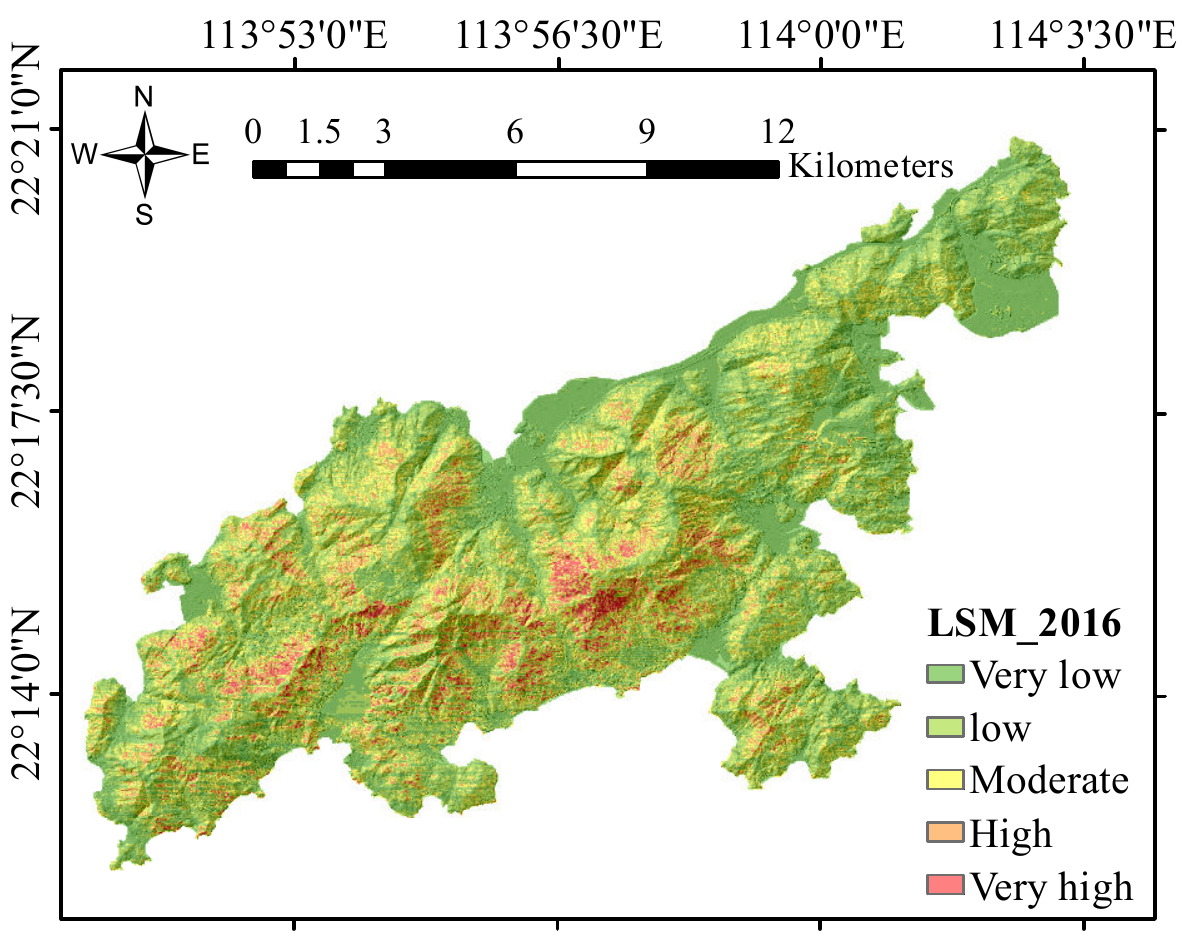}
    \end{subfigure}
    \begin{subfigure}{0.242\textwidth}
        \includegraphics[width=\textwidth]{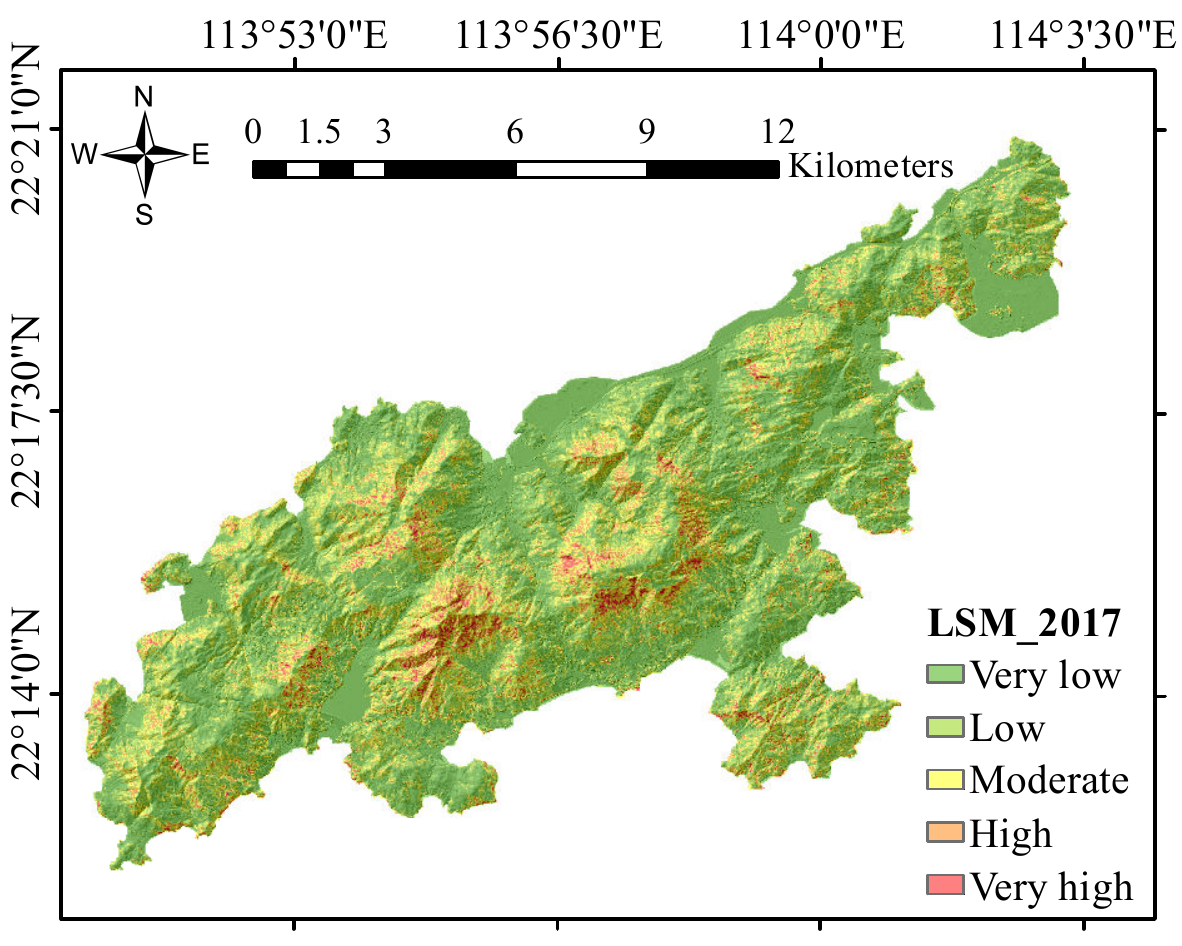}
    \end{subfigure}
	\begin{subfigure}{0.242\textwidth}
        \includegraphics[width=\textwidth]{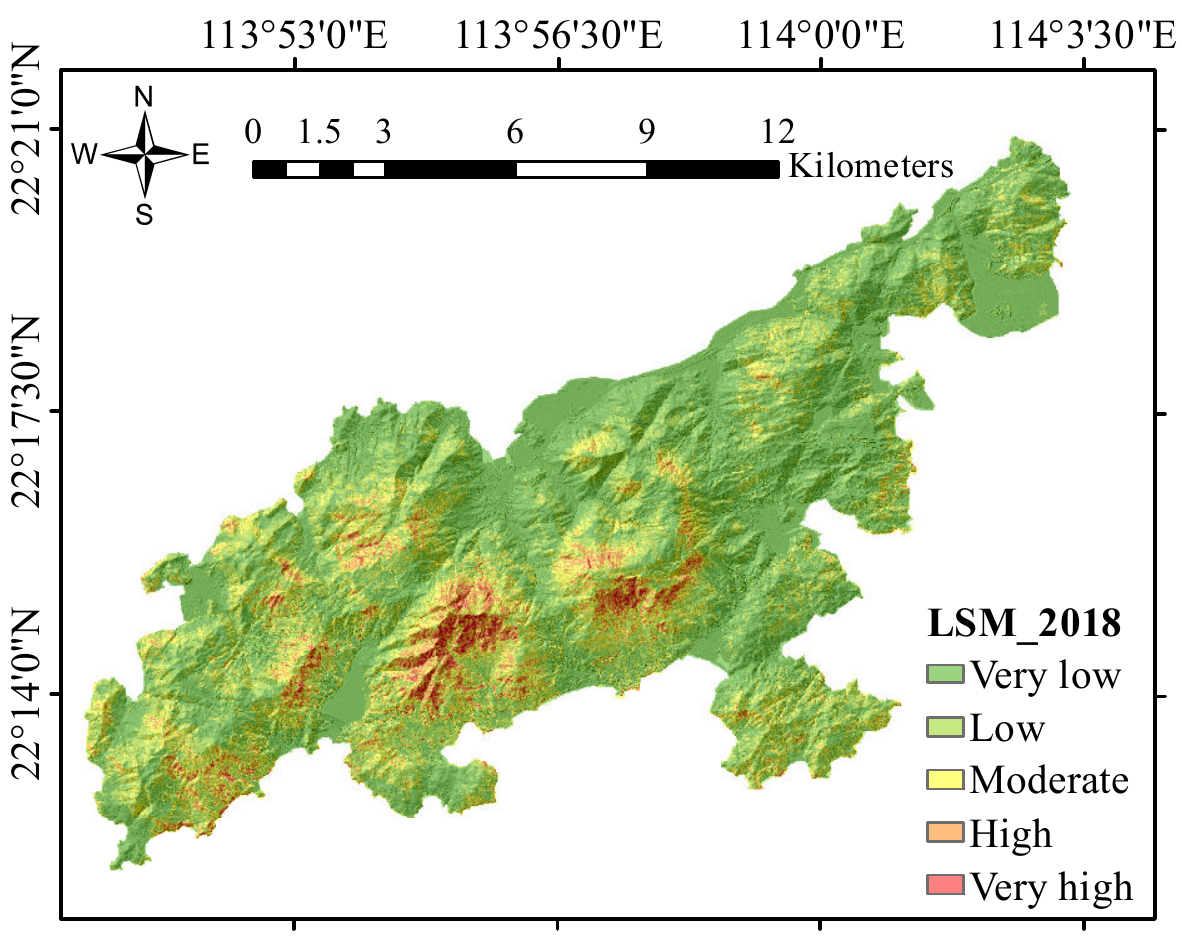}
    \end{subfigure}
    \begin{subfigure}{0.242\textwidth}
        \includegraphics[width=\textwidth]{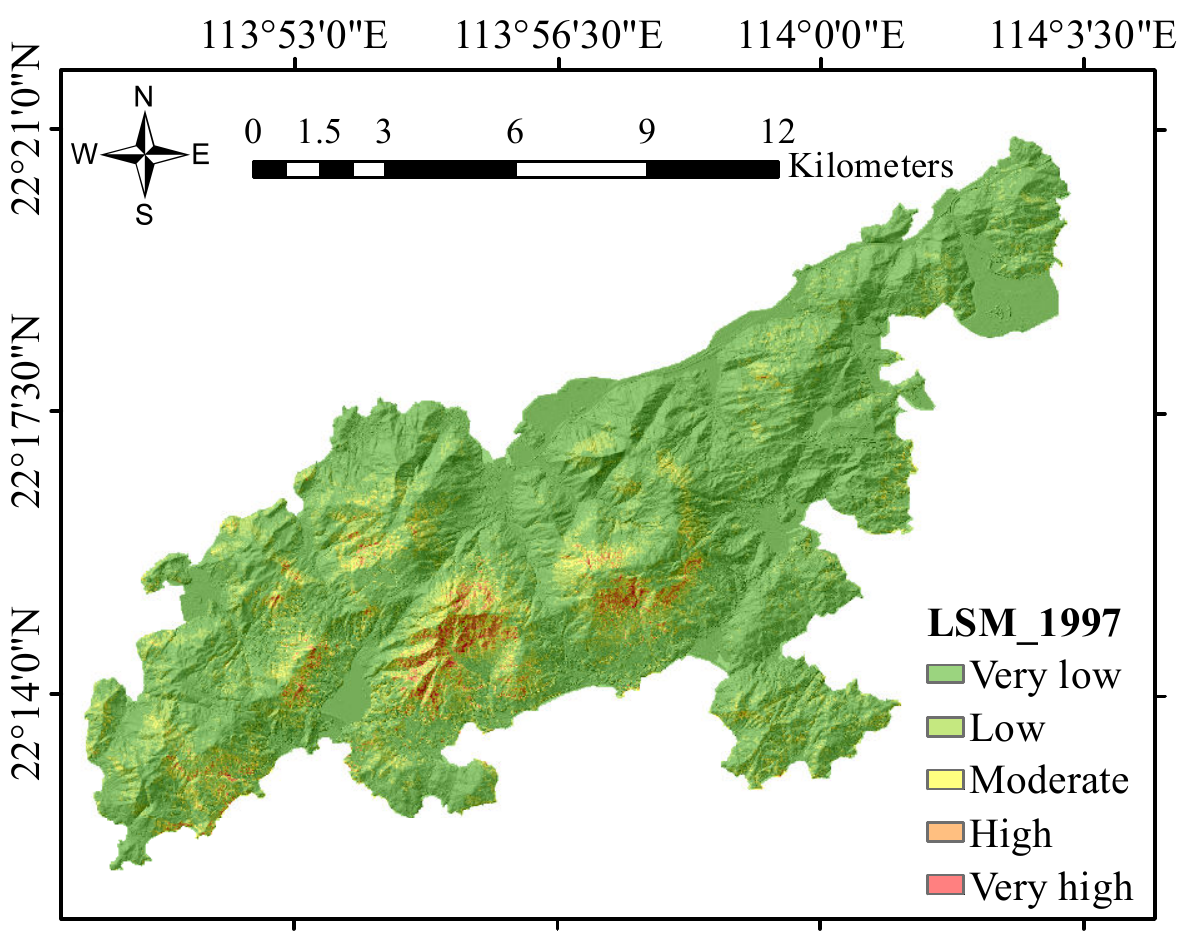}
    \end{subfigure}

    \caption{LSMs covering the years from 1992 to 2019.}
    \label{fig:LSMs}
\end{figure}

We rasterized the Lantau Island with a $10 m \times 10 m$ resolution, and generated sample vectors from each gird for landslide susceptibility prediction. When assigning values to the dimensions of a sample vector connecting the static LIFs, all samples utilize the same sampling method and thematic maps. AR and AERD maps are produced annually for vector sampling of dynamic LIFs due to significant annual variability in rainfall conditions. Some years featured an ample amount of samples, for which the RF method was employed. In contrast, the years that had a limited number of samples, necessitated the use of the proposed meta-learning approach. We fed these sample vectors and predicted probability as introduced in section \ref{s:Methodology}. Then, we classified the grids into five categories that include very low, low, moderate, high, and very high levels. This study presents the LSMs for the years from 1992 to 2019, as shown in Fig. \ref{fig:LSMs}. 

The variation between these landslide susceptibility maps can be attributed to the independent training of each model using different datasets. Overall, our research reveals large-scale areas of high landslide susceptibility at the end of the 20th century, the beginning of the 21st century, possibly driven by external triggering factors like extreme weather conditions, as a result of global climate change. From each of the LSM, we can find an imbalance in landslide susceptibility between the eastern and western regions of the study area, primarily resulting from the uneven distribution of landslide samples. This imbalance is also attributed to the influence of AERD (explored in Section \ref{subs:4.2}) that can only be identified at a low resolution. The distribution of rainfall gauges in use was found to be roughly oriented in an east-west direction. Also, it has been found that the landslide susceptibility on the windward slope in eastern Lantau Island is considerably greater than that on the leeward slope. This is primarily due to the higher amount of rainfall on the windward slope, caused by the southeast monsoon during summer.

In 2008, Hong Kong experienced the most severe rainfall in nearly a century. Through the landslide susceptibility map (LSM) of the same year, a noteworthy increase and aggregation are observed in areas with very high (red) landslide susceptibility, particularly around the ridge of Yi Tung Shan and Lantau Peak. Following 2008, there has been a substantial decrease in the number of areas highly prone to landslides in Hong Kong. On one hand, the Hong Kong Observatory reports a reduction in extreme rainfall conditions. On the other hand, the implementation of LPMitP by CEDD has had a significant effect since 2010.

\subsection{LIF permutation}
\label{subs:4.2}
We used 28 trained models to predict landslide susceptibility over a period of 28 years, respectively. Using feature permutation as introduced in section \ref{subs:3.2}, we identified the dominant LIFs and examined how they contribute to landslides.

\subsubsection{LIF permutation with overall datasets}
\begin{figure}[tbhp]
    \centering
    \begin{subfigure}{0.61\textwidth}
        \includegraphics[width=\textwidth]{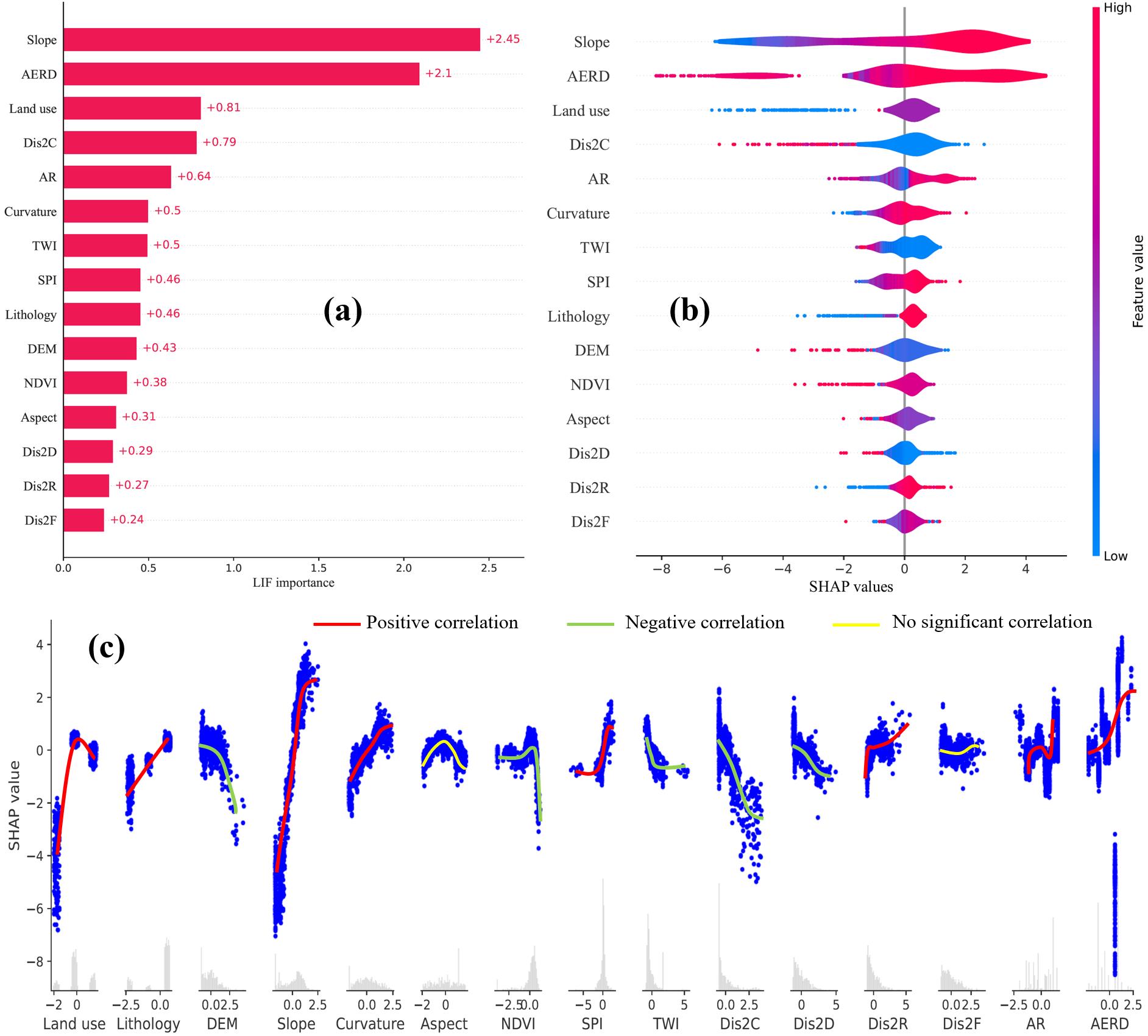}
        \includegraphics[width=\textwidth]{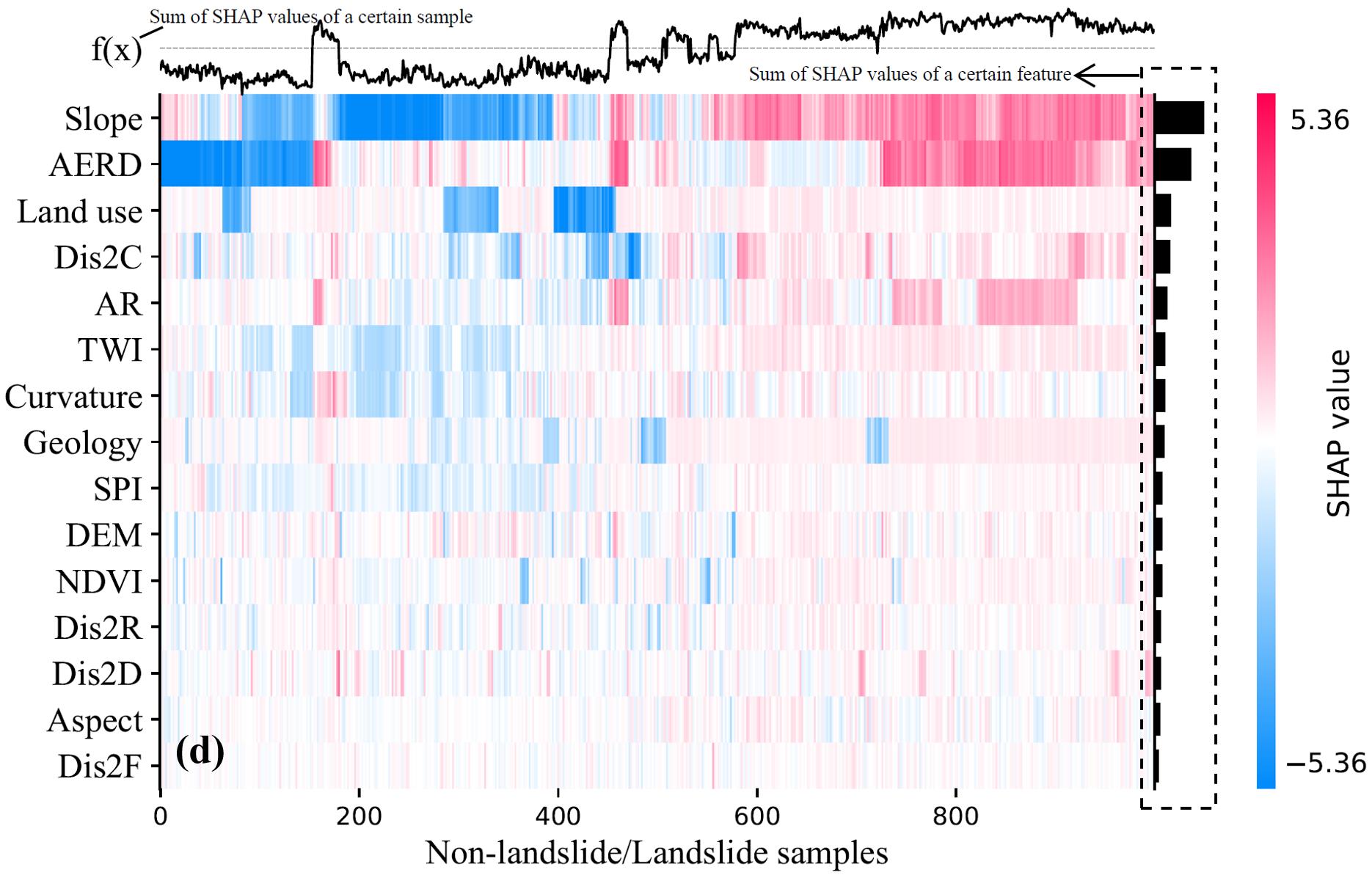}
    \end{subfigure}
    \begin{subfigure}{0.34\textwidth}
        \includegraphics[width=\textwidth]{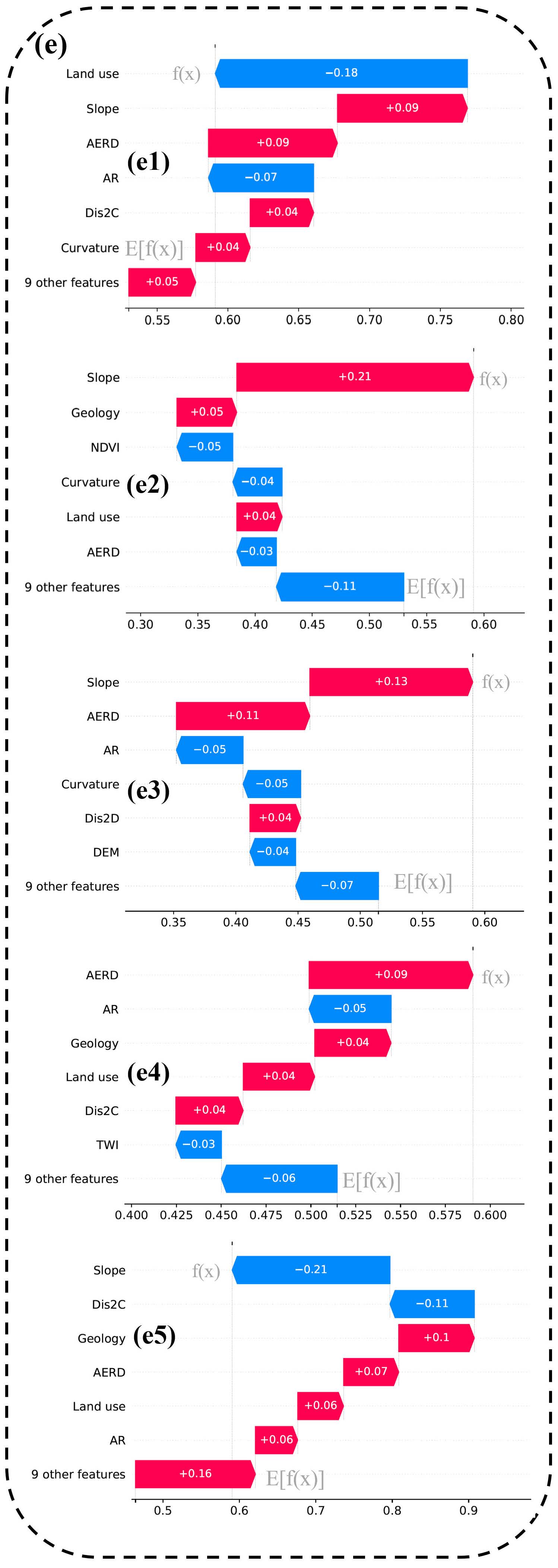}
    \end{subfigure}
    \caption{Effects of LIFs on model prediction. (a) The sum of SHAP values for various LIFs, which indicates their relative importance in predicting landslide susceptibility. Larger summation values correspond to greater influence. (b) The violin plot displays the impact of each LIF on the model output. (c) The scatter plot displays SHAP values of each feature for each sample. The histogram below summarizes the distribution of the SHAP values. The colored curve indicates the positive and negative correlation between LIF and SHAP values. (d) The heatmap plot displays the sum of SHAP values of samples along each feature, and the sum of SHAP values of features for each given sample. (e) The influence of single-sample features (e1-4) on landslide prediction. f(x) denotes the predicted value of a single sample; while E[f(x)] represents the base value - the expected value of f(x) for all samples.}
\label{fig:feature permutation overall}
\end{figure}

First, we conducted overall feature permutation by training the standard XGBoost model with samples from all years (1992 - 2017). Fig. \ref{fig:feature permutation overall} (a) shows that the most influential LIF is slope, followed by AERD. This is consistent with the geographical environment of Hong Kong which is featured mainly with hilly mountains and occasionally suffered from severe rainfall. In recent decades, climate change has resulted in more frequent extreme weather events, with abnormal heavy rainfall being the most common triggering factor. Fig. \ref{fig:feature permutation overall} (b) shows that higher slope and AERD values correspond to higher SHAP values, indicating higher landslide susceptibility. Other feature values that positively correlate with landslide causes include land use, AR, curvature, SPI, lithology, and Dis2R. The correlation between landslide features and landslide susceptibility is further illustrated in Fig. \ref{fig:feature permutation overall} (c). The features that exhibit a negative correlation with landslide susceptibility include distance to catchment (Dis2C), TWI, DEM, NDVI, and distance to drainage (Dis2D). Aspect and distance to faults (Dis2F) were found to have no significant correlation with landslides. In Fig. \ref{fig:feature permutation overall} (d), a heatmap plot is displayed with instances plotted on the x-axis, model inputs on the y-axis, and SHAP values represented by the color bar. This visualization is generated by passing a matrix of SHAP values to the appropriate function. To fully understand Fig. \ref{fig:feature permutation overall} (e), take Fig. \ref{fig:feature permutation overall} (e3) for example, the base value E[f(x)] is initially at around 0.525. 9 other features nearly have no impact on the prediction of the sample, while minor adjustments being made by distance to faults and curvature towards prediction shifting to the left. Lithology and NDVI have a significant effect, positively influencing the sample prediction to shift towards the right. Notably, the two most influential factors are slope and AERD, leading to positive and negative predictions respectively. It is possible for the sample to have a large slope but low AERD values.

\subsubsection{LIF permutation for each year in the span from 1992 to 2019}
Fig. \ref{fig:feature permutation overall} can reflect the cause of the landslide as a whole, but there are limitations when applied to specific years. This is because the LIE is subject to vary every year under the influence of global climate change. Therefore, we analyzed the primary landslide causes for each year between 1992 and 2019. In order to save space in the article, the feature permutation results for each year are presented in the supplemental materials. It’s apparent that the impact of LIFs on landslides changes every year. The study of such evolution will be further discussed in section \ref{subs:5.3}. 

\begin{figure}[tbhp]
    \centering
    \begin{subfigure}{0.9\textwidth}
        \includegraphics[width=\textwidth]{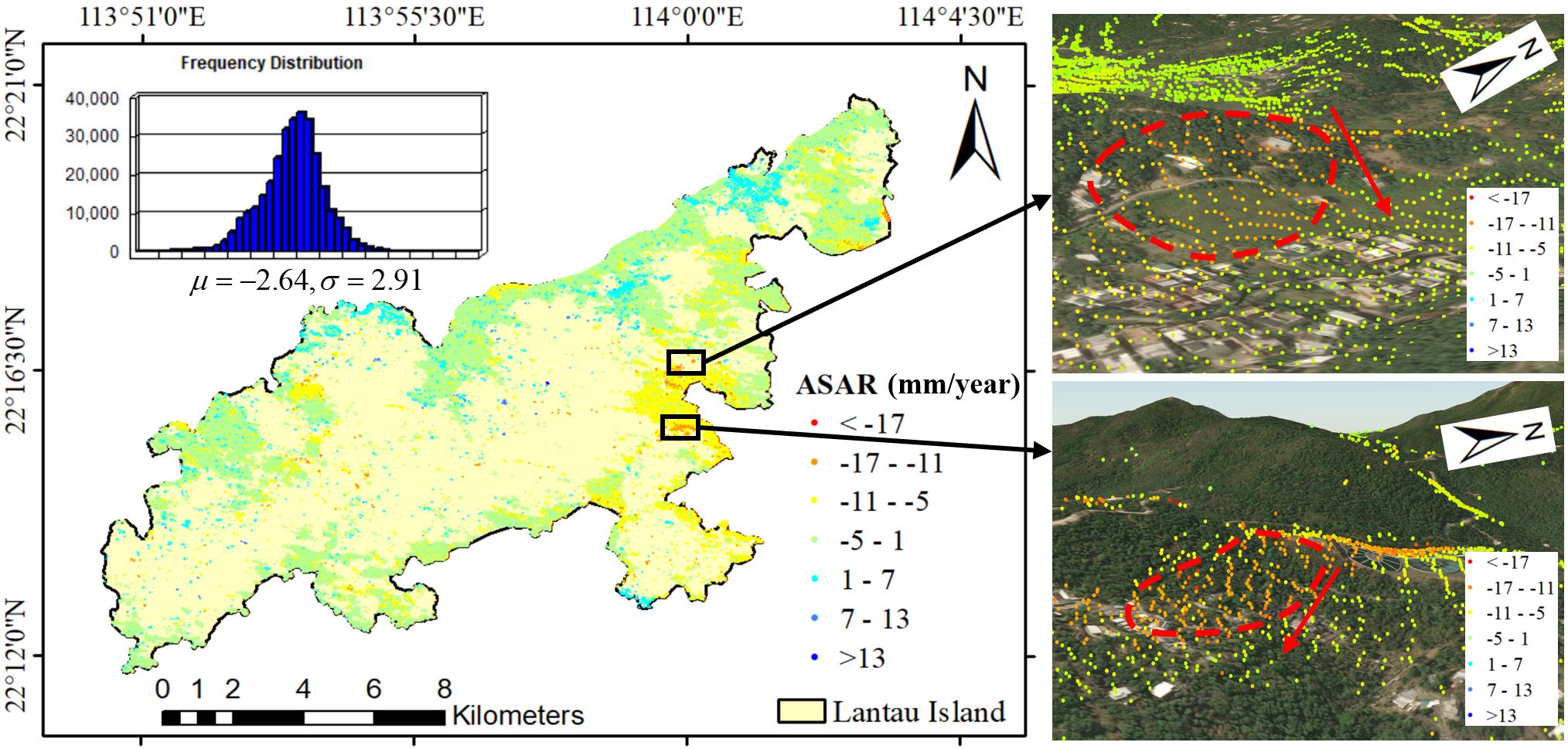}
    \end{subfigure}
    \begin{subfigure}{0.9\textwidth}
        \includegraphics[width=\textwidth]{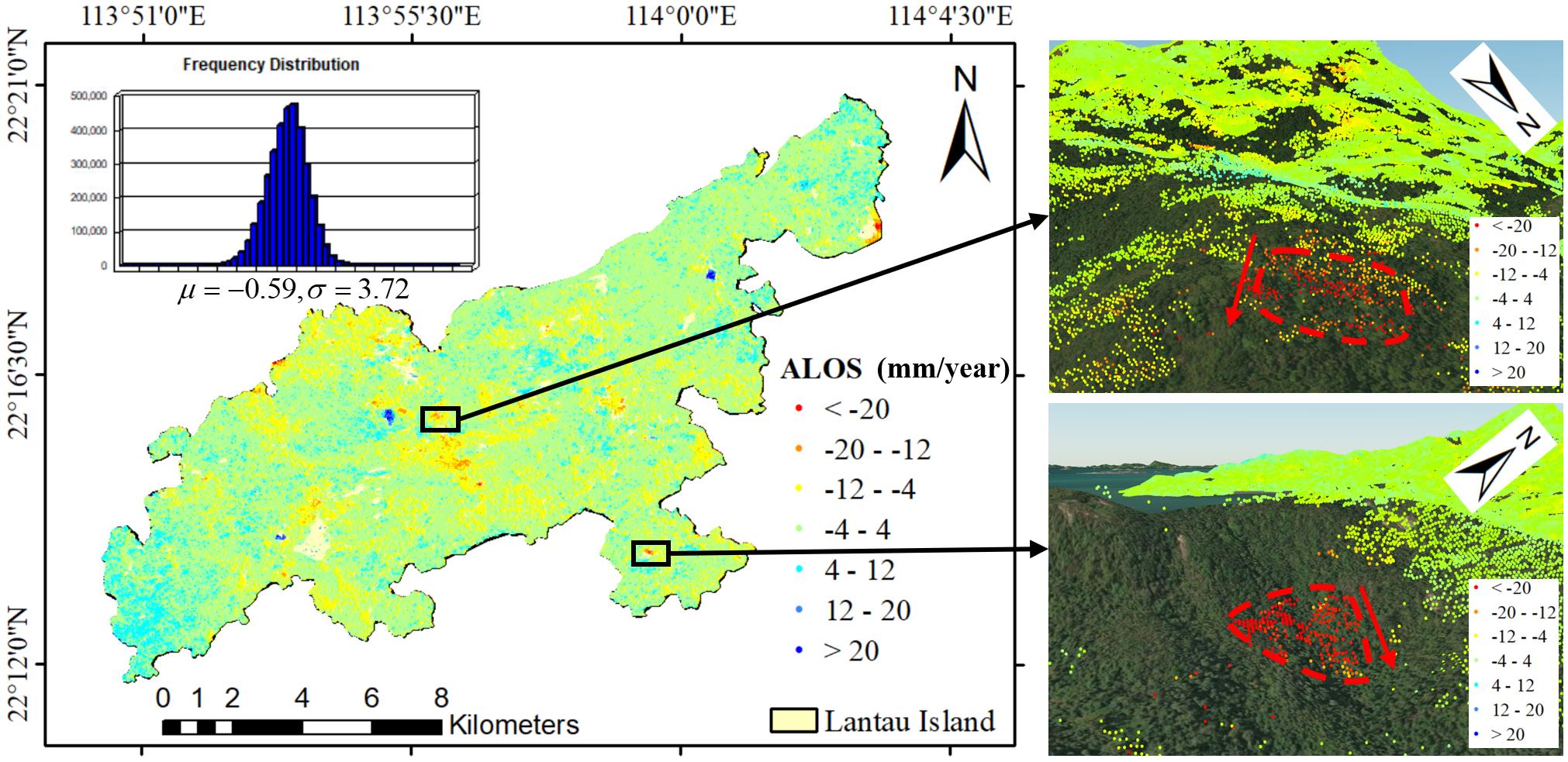}
    \end{subfigure}
    \begin{subfigure}{0.9\textwidth}
        \includegraphics[width=\textwidth]{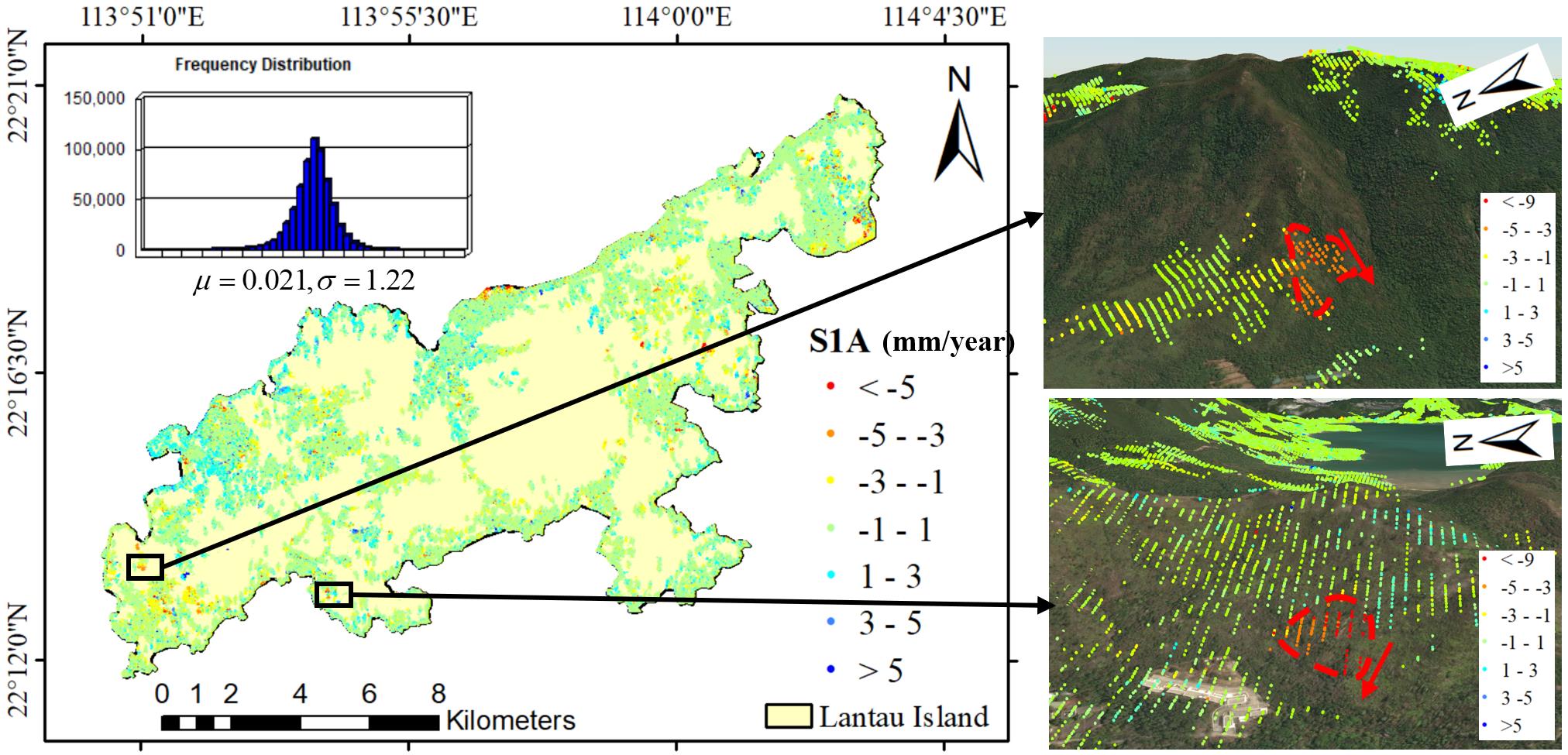}
    \end{subfigure}
    \caption{The deformation velocity maps (a)-(c) in 2003.04-2010.09 (period 1), 2007.06-2011.01 (period 2) and 2015.06-2019.12 (period 3). The upper left shows the frequency distribution, where $\mu$ is the mean value, $\theta$ is the standard deviation. The subfigures give a 3D visualization for zoomed slow-moving slopes along the east or west direction with evident ground deformation.}
    \label{fig:deformation map}
\end{figure}

\subsection{Ground deformation}
The deformation velocity maps for the three periods were derived as is shown in Fig. \ref{fig:deformation map}. The figures present a total of 311099, 3547634, and 659909 measurement points in period 1-3, respectively. The figures also provide the mean and standard deviation of these TS points. The deformation velocity ranges for the three periods are (-17.48, 13.83), (-31.1, 30.08), and (-9, 7.2), respectively. ALOS has more TS points because its long-wave characteristics allow for better penetration through forests. However, its temporal resolution is low and the echoes are unstable. In certain instances, optimizing InSAR results may involve a tradeoff between the quality and quantity of TS points. In period 1, severe deformation regions are concentrated on the eastern part of Lantau Island. In period 2, there are multiple deformation regions that concentrate on the central mountainous area of Lantau Island. Period 2 exhibited the most severe ground deformation among the three periods, with numerous regions experiencing severe ground deformation, closely associated with exceptionally high rainfall. This topic will be further discussed in section \ref{subs:5.3}. In contrast, the observed ground deformation in period 3 has apparently decreased, possibly due to the reduction of extreme rainfall events as well as the effectiveness of the Landslide Prevention and Mitigation Programme that consolidates many of the slopes in Hong Kong. This assertion will be further substantiated in section \ref{subs:5.4}. Due to the limitations of InSAR techniques in areas with dense vegetation \citep{7329983}, a two-tier network strategy utilizing the MT-InSAR method was implemented to measure surface deformation.

In this article, we have visualized all measurement points in a 3D environment using \textit{Cesium}, overcoming the high rendering costs associated with importing \textit{*.kml} files into \textit{Google Earth Engine}, which had limitations in handling a large number of data points. The 3D visualization affords an overall perspective under the virtual geographical environment, thus facilitating the fast and reliable identification of potential slow-moving slopes. 

\subsection{LSM enhancement and validation}
\begin{figure}[tbhp]
	\centering
	\includegraphics[width=\textwidth]{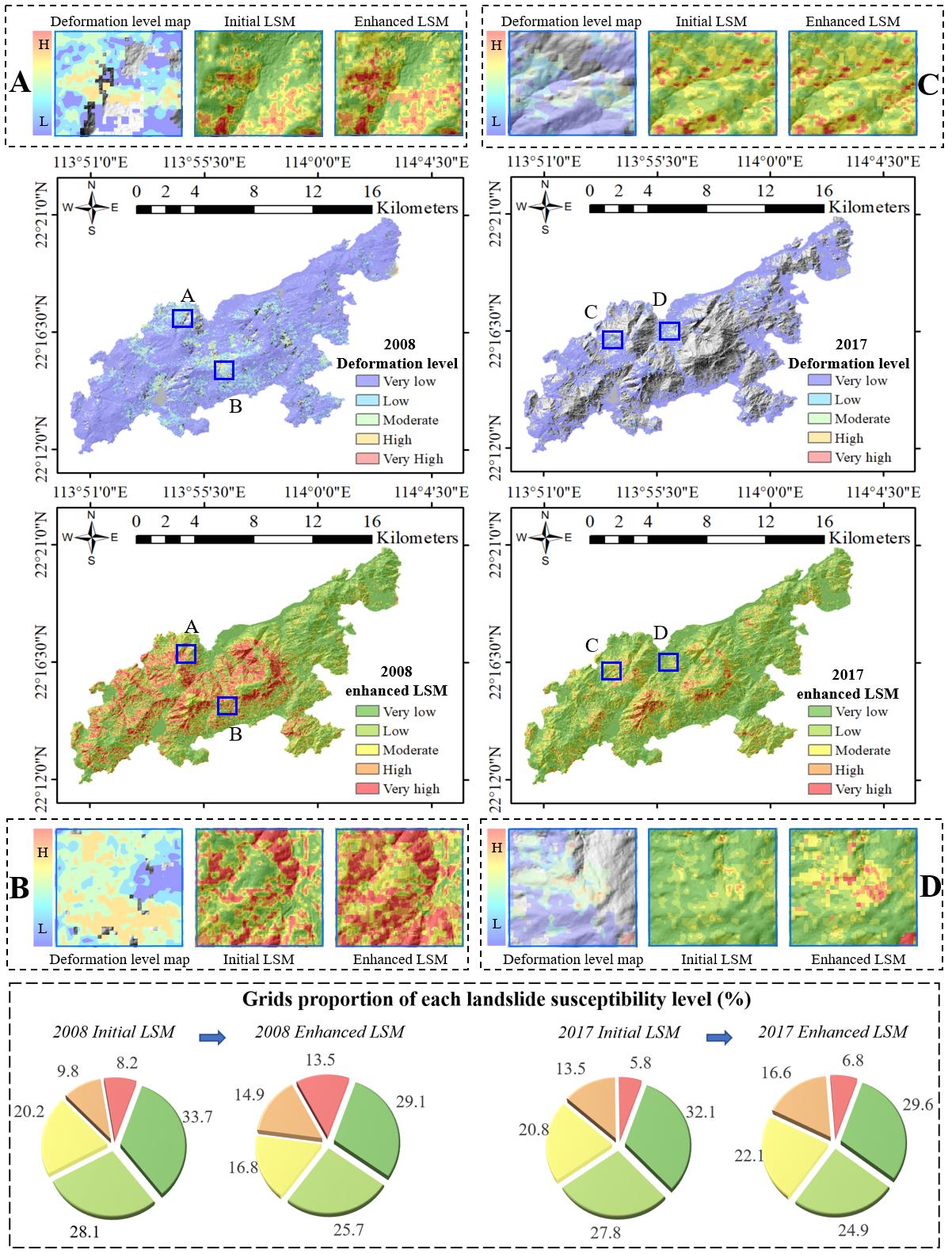}
	\caption{The LSM enhancement result of 2008 (left) and 2017 (right). Above the enhanced LSM are deformation level maps. Regions A, B, C, and D are zoomed to provide an intuitive visualization of the enhancement of LSM by combining the initial LSM and deformation map. L and H in the color bar represent low and high, respectively. The lowest subfigure shows the proportion of grid cells at each landslide susceptibility level.}
	\label{fig:enhanced LSM}
\end{figure}

When significant slope deformations occur in areas initially classified as having low landslide susceptibility, it is necessary to refine the initial predictions according to the deformation specifics. The deformation velocity derived from the MT-InSAR technique was applied to enhance the initial landslide susceptibility map. Fig. \ref{fig:enhanced LSM} shows the LSM enhancement result of 2008 and 2017. The deformation level maps are derived from deformation velocity as is introduced in section \ref{subsubs:3.3.2}. In regions A, B, D, and D, from left to right, the zoomed figures represent deformation level map, initial LS map, and enhanced LS map. In 2008, there was a significant increase in both deformation and landslide susceptibility levels in the mountainous areas of central and northern Lantau Island. This was due to the steep slopes in the mountainous areas, which were highly susceptible to collapse or slow movement under intense rainfall. In 2017, the predicted high-risk areas largely decreased, which can be attributed to the reduced occurrence of extreme rainfall events throughout the year. The implementation of the LPMitP by the Hong Kong government since 2010 has also played a role in this decreasement. Through the proposed LSM enhancement, the proportion of grids with high landslide susceptibility increased from 9.8\% to 14.9\% in 2008 and from 13.5\% to 16.6\% in 2017, while very high landslide susceptibility increased from 8.2\% to 13.5\% in 2008 and from 5.8\% to 6.8\% in 2017. 

\begin{figure}[tbhp]
	\centering
	\includegraphics[width=0.75\textwidth]{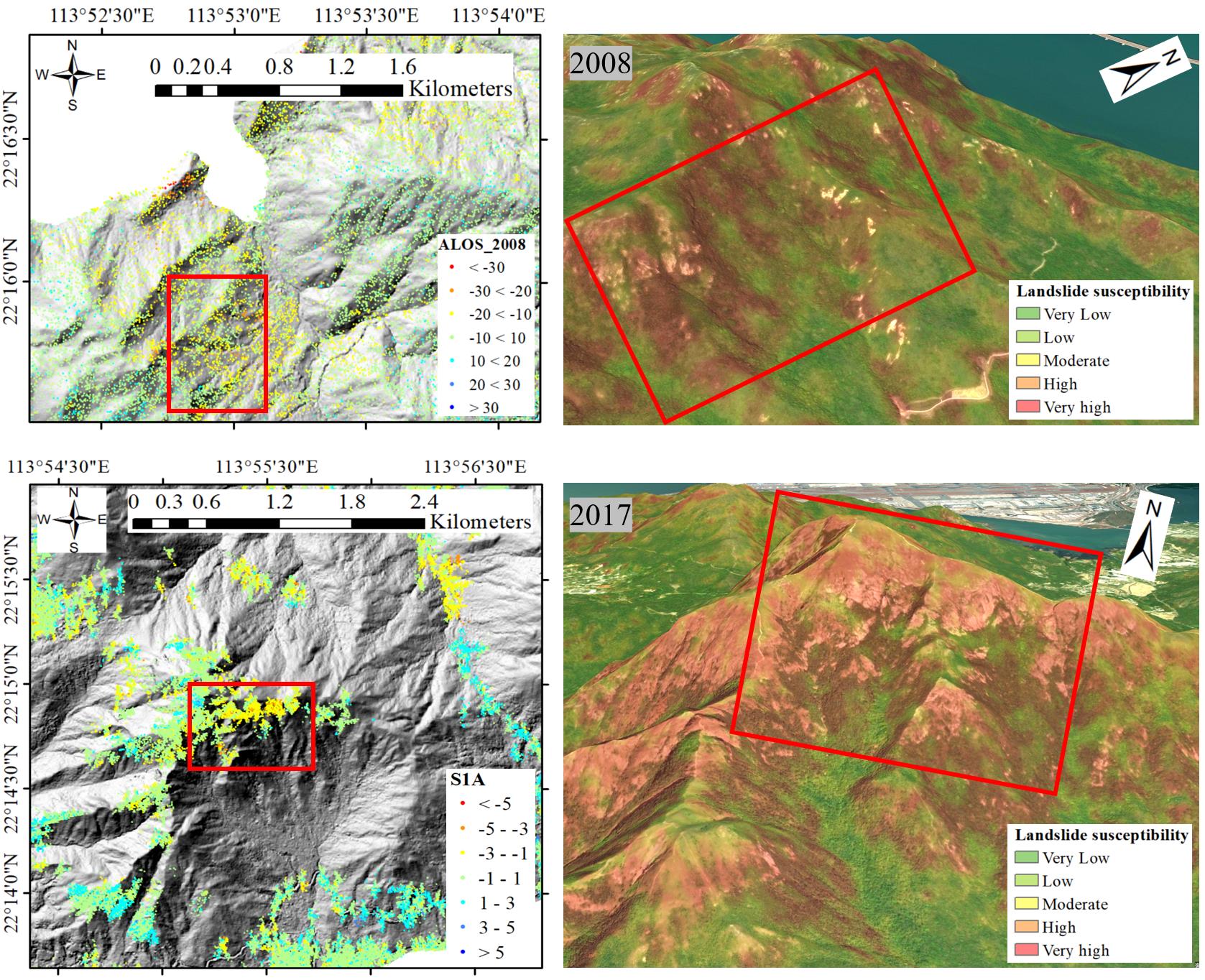}
	\caption{Cross validation of ground deformation map (left) and initial landslide susceptibility map (right) in 2008 and 2017.}
	\label{fig:cross validation}
\end{figure}

Fig. \ref{fig:cross validation} gives two areas on Lantau Island that exhibit a significant overlap in terms of high landslide susceptibility and ground deformation levels in both 2008 and 2017. The corresponding optical images, represented as imagelayers on the 3D terrain (shown on the right), exhibit multiple signs of landslides, also providing validation for the LSM and InSAR results and their correlation in slope dynamics.

\section{Discussion}
\subsection{Performance comparison}
\begin{figure}[tbhp]
    \centering
    \begin{subfigure}{0.45\textwidth}
        \includegraphics[width=\textwidth]{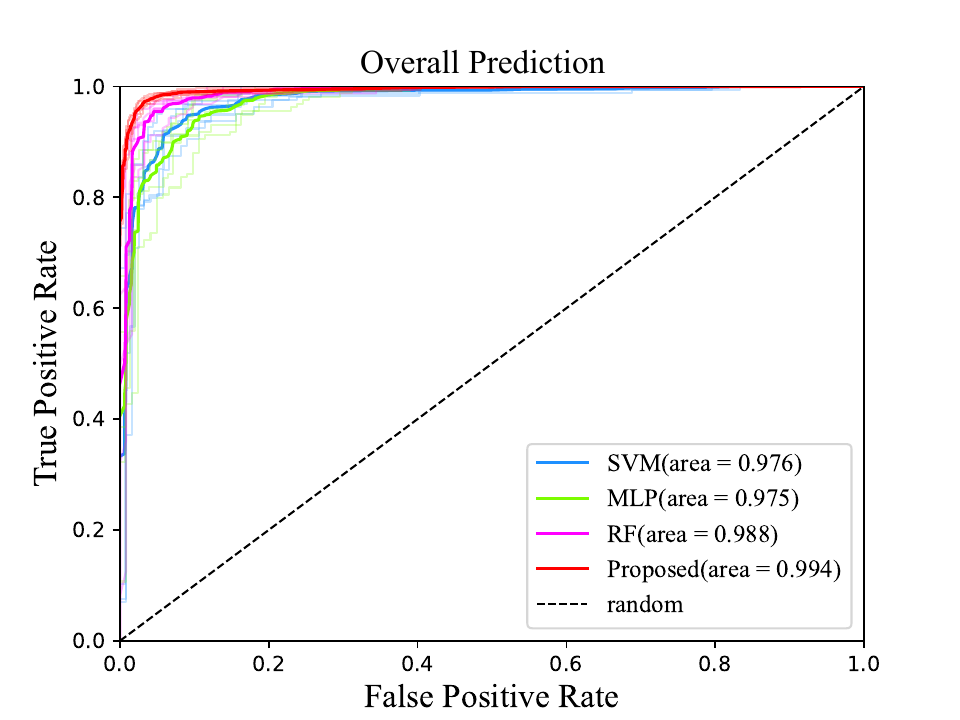}
    \end{subfigure}
    \begin{subfigure}{0.45\textwidth}
        \includegraphics[width=\textwidth]{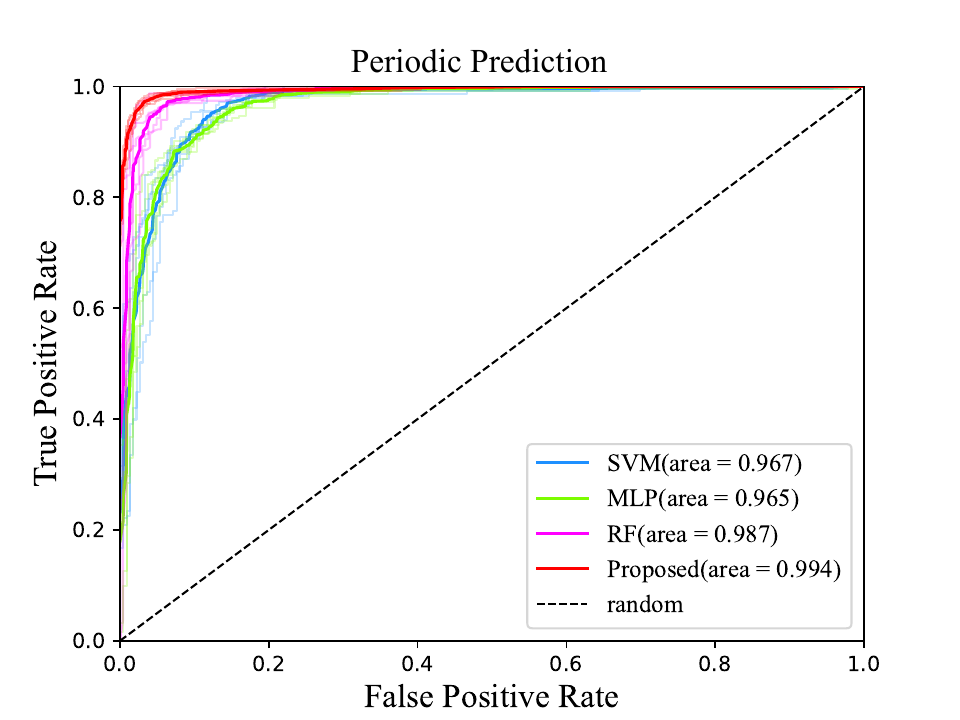}
    \end{subfigure}
    \caption{The ROC of various methods under overall prediction (left) and periodic prediction (right). The dashed lines indicate ROC curves obtained by using 5 groups of shuffled input data. The solid lines represent the average ROC curves. The AUROC is provided in the lower right corner.}
    \label{fig:ROC}
\end{figure}

The indicators for the performance evaluation are listed as follows:
\begin{equation}
    \begin{split}
    \begin{aligned}
        Accuracy&=\frac{TP+TN}{TP+FP+TN+FN},\\
        Precision&=\frac{TP}{TP+FP},\\
        Recall&=\frac{TP}{TP+FN},\\
        \operatorname{F1-score}&=\frac{2TP}{2TP+FP+FN},
    \end{aligned}
  \label{eq:measurement}
\end{split}
\end{equation}
where FP, FN, TP, TN represent the false positive, false negative, true positive and true negative, respectively. Data-driven methods are known to perform well when sufficient and reliable labelled data is available. In contrast to SVM \citep{WANG2019104217}, MLP \citep{huang2020landslide}, and RF-based \citep{SAHA2021142928} approaches that directly perform supervised regression using the dataset, this study conducts LSA for each year and learned multiple predictive models. Specifically, the Random Forest (RF) algorithm was utilized to perform LSA when an adequate number of samples were available for that particular year. In the cases where only a limited number of landslide samples were available, we fine-tuned the meta-learned model with only a few samples and gradient descent iterations. The fast adaptation not only reduced computational expenses, but also accomplished efficient modeling for dynamic LSA. Despite using a fast-learning strategy, the proposed approach outperformed other methods in accuracy (3\%-7\%), precision (2\%-9\%), recall (3\%-5\%), and F1-score (2\%-7\%), as is shown in Table \ref{tab:statistical measures}. The performance of SVM, MLP, and RF models is better in overall landslide prediction than in predicting landslides periodically. This outcome is due mainly to the lack of available landslide records in certain years, which in turn results in insufficient data for model training. We tested the stability of the proposed model by plotting ROC curves of the various methods. Our findings show that the proposed method performed well across all five groups of shuffled input, with a mean AUROC of 0.994 (as shown in Fig. \ref{fig:ROC}), which is higher than that of other methods.

\begin{table*}[tbhp]
    \caption{Statistical performance of various methods with the original and augmented landslide inventories.}
    \centering
    \footnotesize
    \setlength\tabcolsep{2.5pt}  
    \begin{tabular}{clccccc}
      \hline
      Datasets                                          & Models           & Accuracy(\%)     &Precision(\%)     & Recall(\%)      & F1-score(\%)      & AUROC(\%)          \\
      \hline
      \multirow{3}{*}{\makecell[c]{Overall \\ prediction}}                 & SVM              & 91.9             & 90.0            & 94.3              & 92.1               & 0.976              \\
                                                        & MLP              & 90.4             & 88.8             & 92.1            & 90.5              & 0.975              \\
                                                        & RF               & 95.8             & 97.3             & 93.9            & 95.5              & 0.988              \\ \hline
      \multirow{4}{*}{\makecell[c]{Periodic \\ prediction}}                & SVM              & 89.3             & 87.1            & 92.2              & 90.7               & 0.967              \\
                                                        & MLP              & 87.2             & 87.0             & 90.5            & 88.9              & 0.965              \\
                                                        & RF               & 94.1             & 95.8             & 93.2            & 95.1              & 0.987              \\ 
                                                        & Proposed         & \textbf{97.3}    & \textbf{98.2}    & \textbf{96.9}   & \textbf{97.5}     & \textbf{0.994}     \\ \hline
    \end{tabular}
    \label{tab:statistical measures}
\end{table*}

\subsection{Few-shot adaption performance}
\begin{figure}[bhp]
	\centering
	\includegraphics[width=\textwidth]{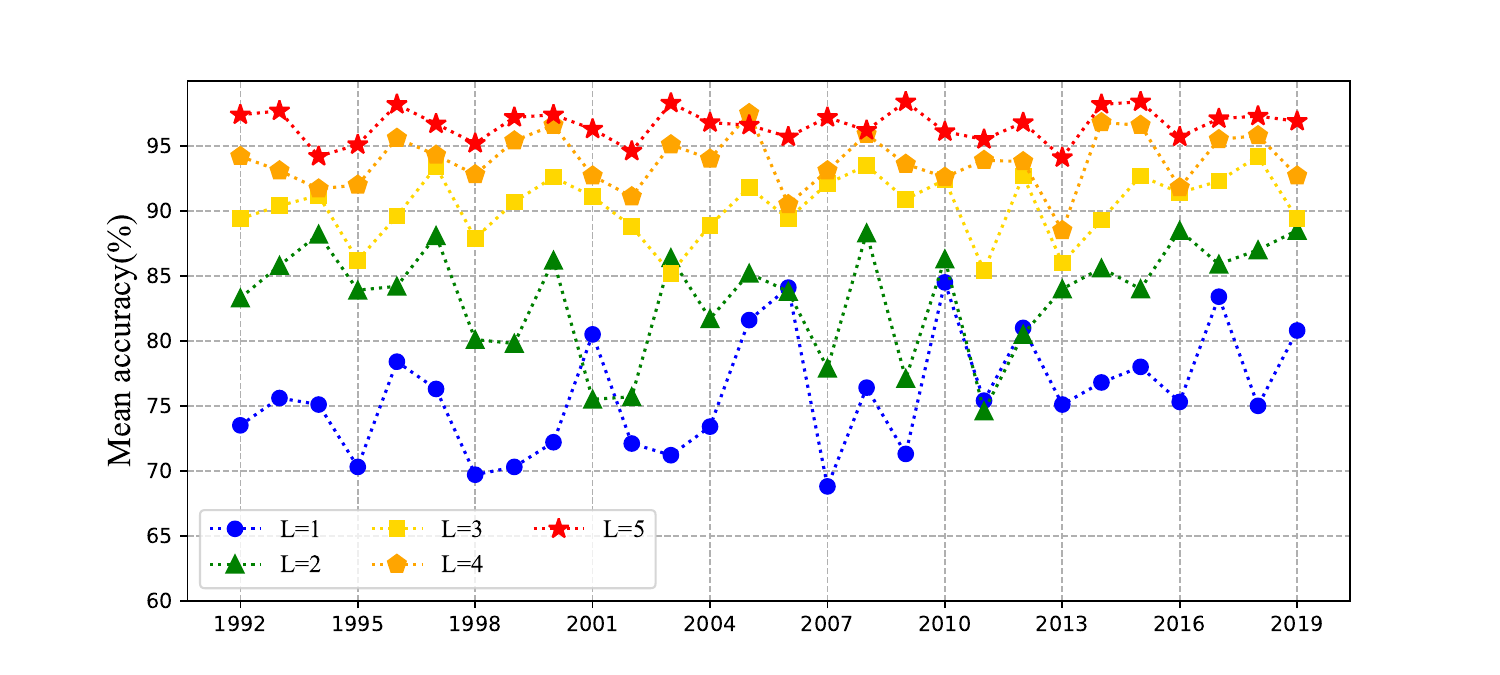}
	\caption{Each marked point represents the mean OA performance of a fast-adapted model on an LSA task for a specific year. L refers to the number of iterations used for model adaptation.}
	\label{fig:FSL scatter}
\end{figure}

\begin{figure}[tbhp]
	\centering
	\includegraphics[width=0.45\textwidth]{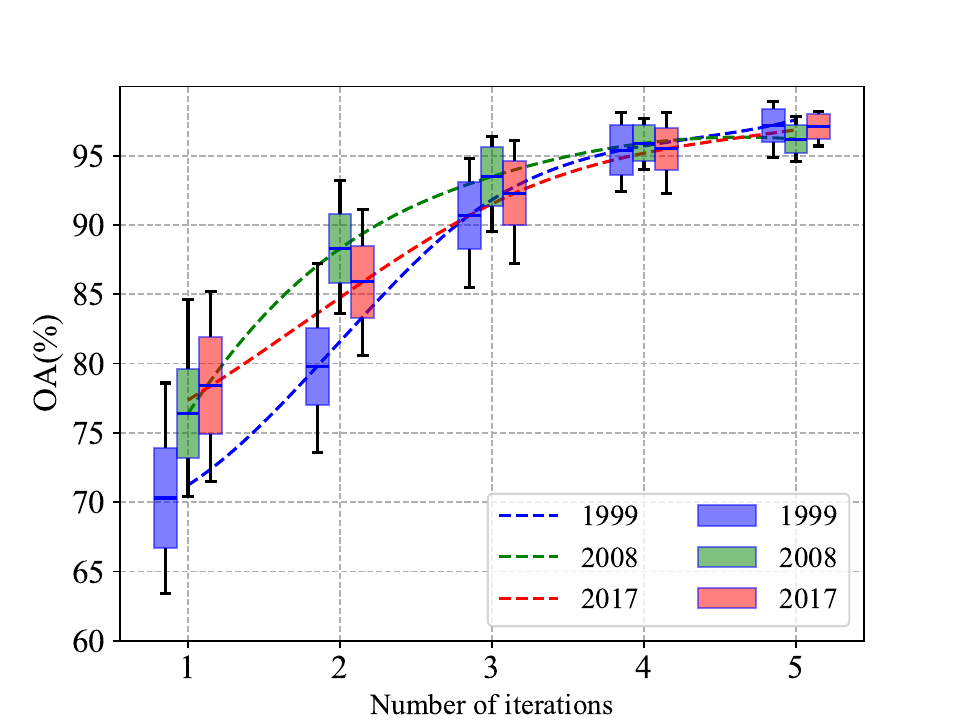}
	\caption{The OA performance of 1999, 2008, and 2017, with varying numbers of gradient descent updates. The bars indicate the range of the OA values, where the mean OA is represented by the middle line, and the height of the bar indicates two times the standard deviation. Additionally, the horizontal lines convey the maximum and minimum OA.}
	\label{fig:FSL candle}
\end{figure}

One superiority of the proposed method is that it allows for fast adaptation of the predictive model using a few samples and gradient descent updates in years with limited landslide records. To fully exploit the potential of the meta-learned representations, we directly employ these representations to adapt the LSA models for other years, as the representation learning also involves samples from these years. Fig. \ref{fig:FSL scatter} illustrates the overall accuracy (OA) performance of the proposed method on the LSA task for each year. The batch size of 8 samples was used for each iteration. As the number of iterations L increases, the OA performance significantly improves. When $L=5$, satisfactory OA performance is obtained for all LSA subtasks. Furthermore, candle charts for the years 1999, 2008, and 2017 with different numbers of updating iterations were also plotted, as is shown in Fig. \ref{fig:FSL candle}, which shows a fast improvement in both OA and stability (a reduction in standard deviation), indicating the validity of the proposed few-shot adaptations. 

\subsection{The vulnerability of slopes to extreme rainfall events}
\label{subs:5.3}
\begin{figure}[tbhp]
	\centering
	\includegraphics[width=0.65\textwidth]{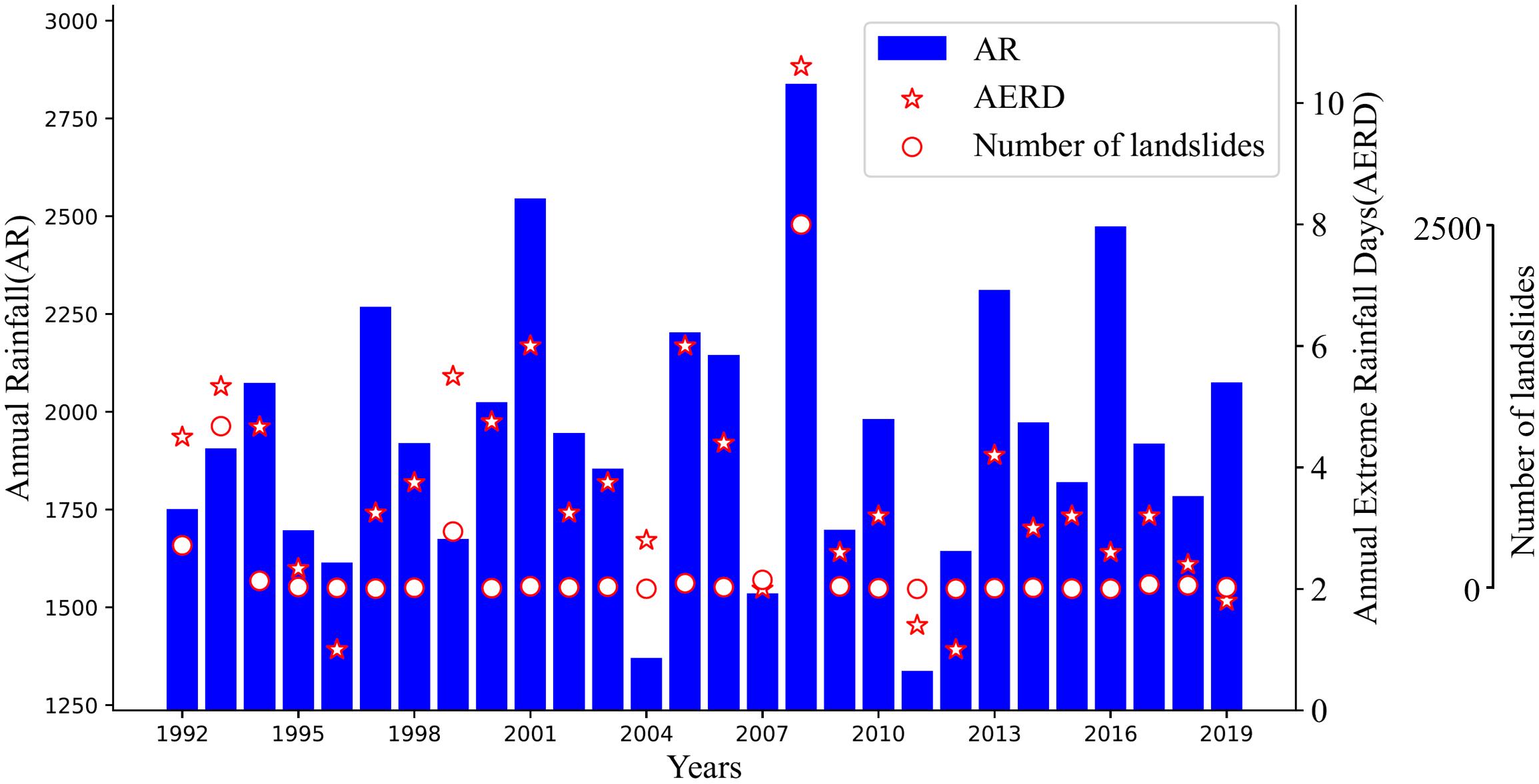}
	\caption{AR, AERD, and number of landslides from 1992 to 2019.}
	\label{fig:histogram}
\end{figure}

The causes of landslide disasters can be attributed to factors such as surface lithology, rugged topography, climate change, and human intervention. These causes can be divided into environmental factors and triggering factors, among which rainfall is the most influential external force in landslide occurrence \citep{DOU2019332}. Fig. \ref{fig:histogram} showcases the average annual rainfall (AR) and annual extreme rainfall days (AERD) spanning from 1992 to 2019. Additionally, the figure also depicts the number of landslides for each year. We observed that in 2008, both AR and AERD reached their highest levels in nearly decades. This year, due to the great damage caused by triggered landslides, GEO traced and recorded a large number of occurred landslides. Additionally, examining the period from 1992 to 2019, we noticed that there was no significant trend of AR, while AERD exhibited a certain declining trend. From a visual perspective, we found that AR and AERD are somewhat related but do not have a strict positive correlation. There is a stronger positive correlation between the recorded number of landslides and AERD compared to the positive correlation with AR. This can be also reflected in Fig. \ref{fig:feature permutation overall} that AERD has a greater contribution to landslides compared to AR, as indicated by the interpretation of the data-driven model. 

\begin{figure}[tbhp]
	\centering
	\includegraphics[width=\textwidth]{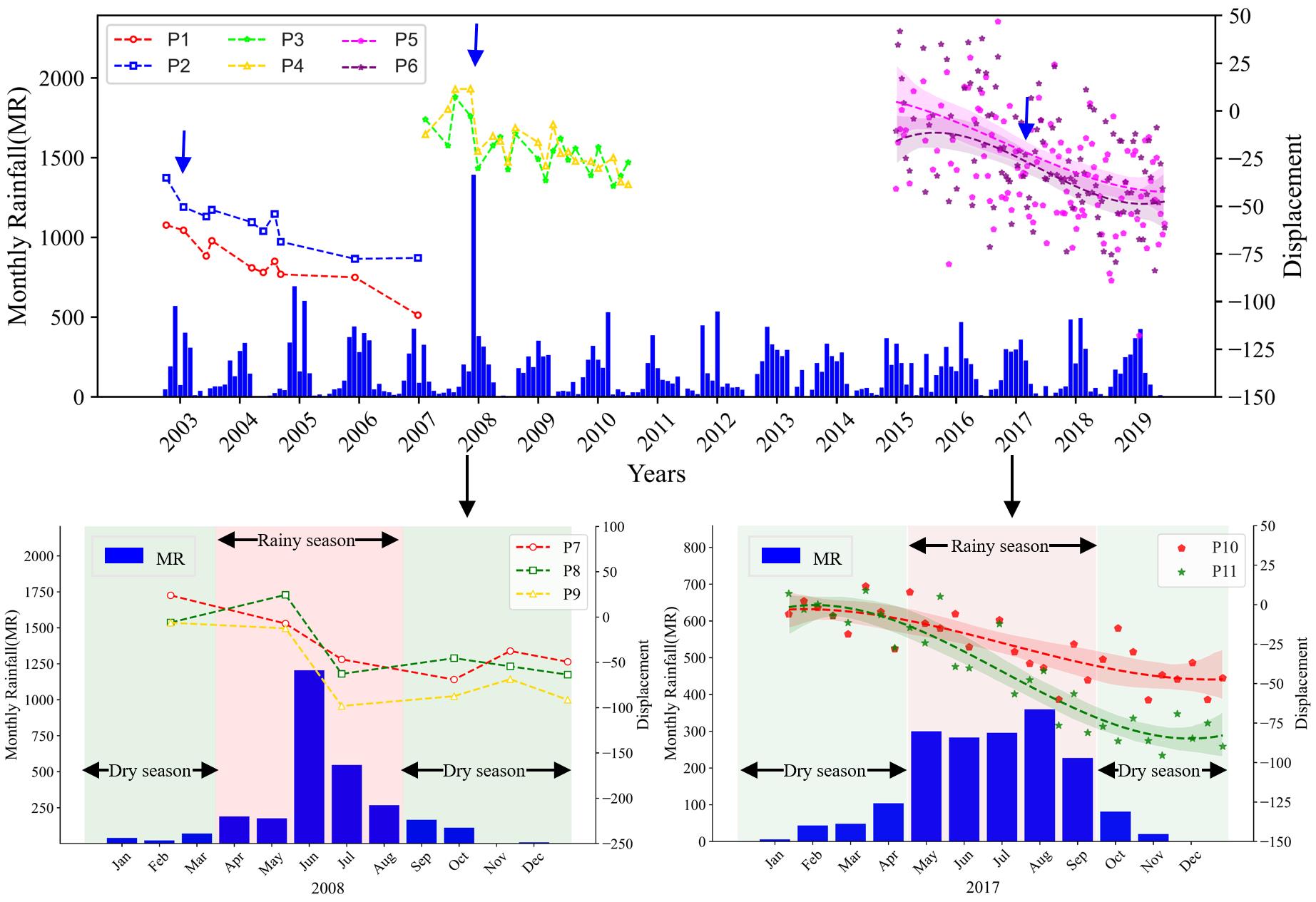}
	\caption{Relationship between monthly rainfall (mm) and TS point displacement (mm). The upper subfigure shows the relationships for the years with InSAR data (see in Table \ref{tab:data source}). The blue arrows indicate the positions with rapid deformation trend. The subfigures below detail the relationships for the years 2008 and 2017. The red and green shaded areas indicate the rainy and dry season, respectively.}
	\label{fig:rainfall and deformation}
\end{figure}

We conducted a study to examine the influence of rainfall on slope movement processes, as shown in Fig. \ref{fig:rainfall and deformation}. The upper subfigure shows the correlation between monthly rainfall conditions and the displacement of TS points (P1-6) in the years for which we have InSAR data. It can be seen that the deformation trend of TS points is significantly influenced by heavy rainfall in 2003, 2004, 2008, 2017 and 2018. Additionally, to comprehensively understand this correlation, we selected additional TS points (P7-11) from deformation regions between 2008 and 2017 and examined their relationship with monthly rainfall. The TS points (P7-9), obtained from ALOS data, had a temporal interval of either 46 or 92 days, while the TS points (P10-11) derived from Sentinel-1A data had a temporal interval of 12 days. To better present the deformation trend, we performed analysis on the TS points from 2015 to 2017 using \textit{seaborn} regression. During the rainy season, there is a noticeable increase in the deformation velocity of these TS points. In June 2008, when the rainfall reached approximately 1250 mm, TS points P1, P2, and P3 exhibited a rapid displacement. Similarly, from May to September 2017 (during the rainy season), both P4 and P5 experienced an accelerating trend in displacement. The above observations clearly demonstrate the direct impact of rainfall factors, such as AR and AERD, on slope dynamics. However, it is noteworthy that the occurrence of landslides is more influenced by monthly or even daily rainfall, rather than annual rainfall. This phenomenon can be also supported by the results of the feature permutation that AERD ranks higher than AR.

\subsection{Landslide causes variations by the effect of climate change and LPMitP}
\label{subs:5.4}
\begin{figure}[tbhp]
	\centering
	\includegraphics[width=\textwidth]{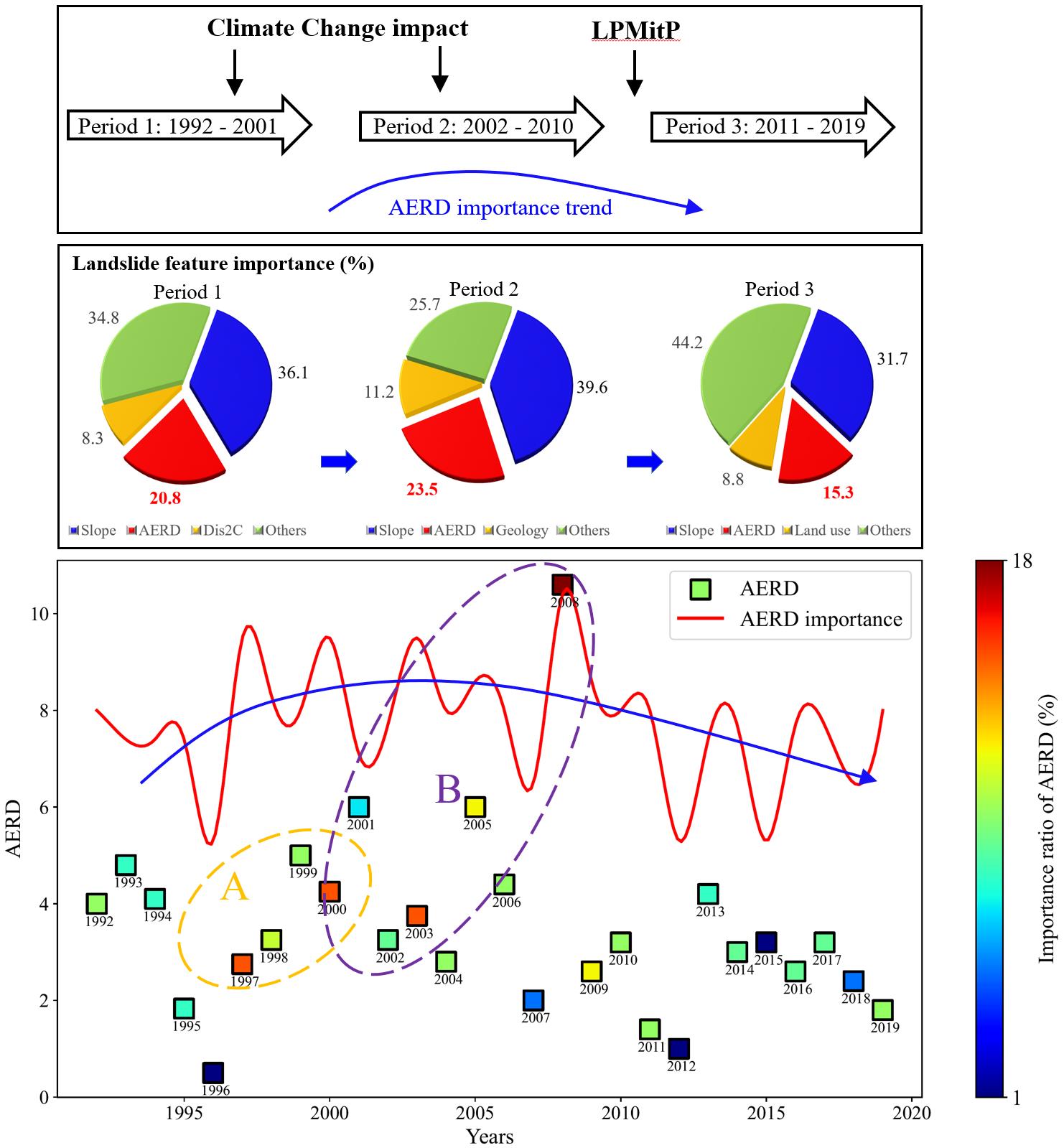}
	\caption{AERD importance variation by the effect of climate change and LPMitP.}
	\label{fig:AERD variation}
\end{figure}

The slope is the most significant factor contributing to landslide occurrence. On one hand, this is attributed to the hilly and steep terrain of Lantau Island. On the other hand, landslides are fundamentally characterized by the downward movement of a substantial quantity of rock or soil along a sloping surface, caused by the force of gravity. Annual extreme rainfall days (AERD) are distinct from other intrinsic LIFs such as topography, geology, and hydrology, as AERD varies over time and is dependent on local climate conditions and changes. Based on the feature permutation results in section \ref{subs:4.2}, it is evident that AERD holds significant importance as an LIF, ranking second only to slope overall. Two primary factors significantly influence the variation in the importance of AERD from 1992 to 2019. One evident factor is the increase in extreme weather events induced by global climate change in the early 21st century. Another factor is the implementation of the LPMitP scheme by the Hong Kong government starting in 2010, focusing on slope consolidation and potential landslide prevention. Slopes are more stable in extreme rainfall events, and the number of landslides is significantly reduced. Fig. \ref{fig:AERD variation} presents the variation of AERD importance by the effect of climate change and LPMitP scheme.

We have divided the study into three distinct temporal periods: 1992-2001 (period 1), 2002-2010 (period 2), and 2011-2019 (period 3). The trend of importance of AERD showed a slight increase during period 1, followed by a decrease in the subsequent periods. To illustrate this, Fig. \ref{fig:AERD variation} presents a pie chart where the AERD importance initially accounted for 20.8\%, then increased to 23.5\%, and eventually declined to 15.3\%. The bottom subfigure of Fig. \ref{fig:AERD variation} shows concretely the AERD and AERD importance with the time changing from 1992 to 2019. In 2008, the elevated position and dark red color of the rectangular shape signify an exceptional frequency of extreme rainfall events, alongside a significantly elevated fraction of AERD importance as elucidated by the data-driven model. Analysis of subfigures in regions A (within period 1) and B (within period 2) consistently demonstrate the paramount role of AERD in the onset of landslides.

\begin{figure}[tbhp]
	\centering
	\includegraphics[width=\textwidth]{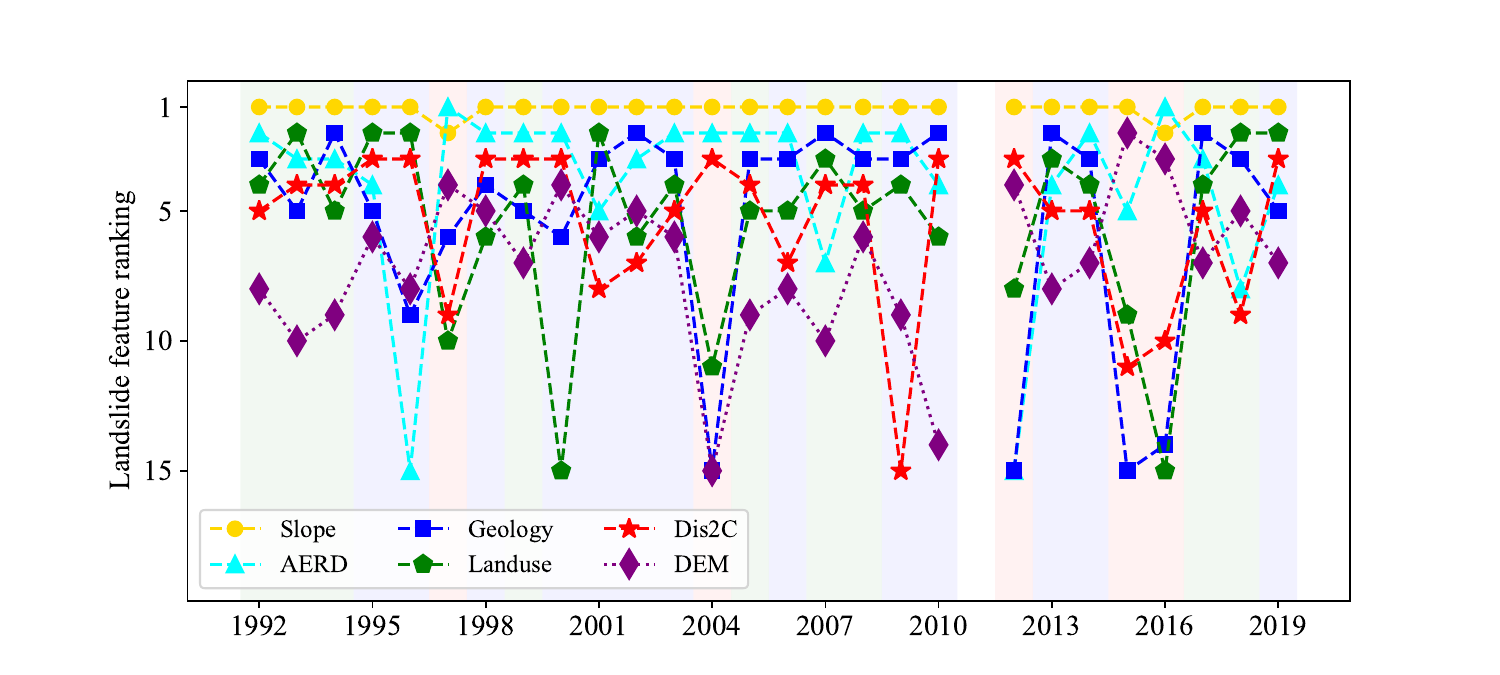}
	\caption{Ranking of top six LIFs from 1992 to 2019. The green shaded area represents the years with over 50 landslide samples. The blue shaded area indicates the years having over 10 but less than 50 landslide samples. The red shaded area indicates the years with less than 10 landslide samples. The white area indicates the absence of available data for that specific year.}
	\label{fig:feature ranking}
\end{figure}

To enhance the inference and analysis of landslide causes during the study period, Fig. \ref{fig:feature ranking} illustrates the ranking of the top six factors (in the order of slope, AERD, geology, land use, Dis2C, and DEM) that noticeably influence the landslide susceptibility prediction overall. The figure illustrates different shaded areas representing varying sample quantities. It is important to note that when there is a lower frequency of landslides, the ranking outcome for LIF assessment becomes less stable. Consequently, some of the top six LIFs may have a lower ranking in years with limited samples.

\section{Conclusion}
In summary, the main objective of this study is to conduct dynamic landslide susceptibility mapping for the inference of variation in landslide causes. Upon the objective, it is inevitable to face the following problems: (1) Dynamic Landslide susceptibility mapping with high time resolution leads to the small sample problem when data are scarce in some years; (2) Most data-driven LSA models are good at fitting samples but is not interpretable. Accordingly, this article presents meta-learning representations with strong fast generalization ability; and explains the impact of input landslide features on the prediction of LSA models. Besides, this study presents the application of InSAR strength for LSM enhancement and validation.

By employing a meta-learning strategy, landslide susceptibility mapping in years with limited samples reached high performance even based on a fast adaptation strategy using a few numbers of samples and gradient descent updates. The interpretation of each model facilitated the identification of the factors contributing to landslides and could provide guidance for effective landslide prevention measures. Slope and AERD have been identified as the most influential LIFs over the recent decades in Lantau Island. The cross-validation between the MT-InSAR-derived deformation velocity map and the data-driven model-derived LSM enhances the effectiveness of the proposed method. In terms of statistical performance, the overall accuracy of the proposed method improved by 3\%-7\% compared to SVM, MLP, and RF-based approaches. This improvement is quite apparent because most machine learning algorithms perform similarly when fitting a large size of samples. The results reveal that climate change and LPMitP are the primary contributors to the variation patterns of landslide causes in Lantau Island, Hong Kong over the past 30 years. The slope maintenance facilities constructed by the Hong Kong government have significantly reduced landslide risks catalyzed by various internal and external factors, including topography, hydrology, geology, land use, and particularly the extreme rainfall conditions. 

\section*{Declaration of Competing Interest}
The authors declare that they have no known competing financial interests or personal relationships that could have appeared to influence the work reported in this paper.

\section*{Acknowledgments}
This work was supported by National Natural Science Foundation of China (41971278), the Research Grants Council (RGC) of Hong Kong (CUHK14223422 and 14201923), and The Chinese University of Hong Kong Faculty of Social Science Direct Grant for Research (4052294).

\bibliographystyle{elsarticle-harv}
\bibliography{reference}
\end{document}